\documentclass[Afour,sageh,times]{aux/sagej}

\usepackage{moreverb,url}

\usepackage[colorlinks,bookmarksopen,bookmarksnumbered,citecolor=red,urlcolor=red]{hyperref}
\usepackage{graphicx}
\usepackage{subcaption}
\usepackage{amsmath,amssymb,amsfonts}
\usepackage[usenames,dvipsnames,svgnames,table]{xcolor}
\usepackage{soul}
\usepackage{ulem}
\usepackage{physics}
\usepackage{cases}
\usepackage{textcomp}
\usepackage{multirow}
\usepackage{booktabs}
\usepackage{tabularx}
\usepackage{threeparttable}
\usepackage{comment}
\usepackage{bbding}
\usepackage{pifont}
\usepackage[nolist]{acronym}
\usepackage[acronym, nomain, nonumberlist]{glossaries}
\usepackage{tikz}
\usetikzlibrary{arrows.meta, positioning, shapes.geometric, calc}
\glsenablehyper
\makenoidxglossaries

\let\origgls\gls
\let\origglspl\glspl
\let\origGls\Gls
\let\origGlspl\Glspl
\renewcommand{\gls}[1]{\ifglsentryexists{#1}{\origgls{#1}}{#1}}
\renewcommand{\glspl}[1]{\ifglsentryexists{#1}{\origglspl{#1}}{#1s}}
\renewcommand{\Gls}[1]{\ifglsentryexists{#1}{\origGls{#1}}{#1}}
\renewcommand{\Glspl}[1]{\ifglsentryexists{#1}{\origGlspl{#1}}{#1s}}
\newacronym{nn}{NN}{Neural Network}
\newacronym{ai}{AI}{Artificial Intelligence}
\newacronym{ml}{ML}{Machine Learning}
\newacronym{mlp}{MLP}{Multi-Layer Perceptron}
\newacronym{rnn}{RNN}{Recurrent Neural Network}
\newacronym{mpc}{MPC}{Model Predictive Control}
\newacronym{pinn}{PINN}{Physics-Informed Neural Network}
\newacronym{msnn}{MSNN}{Model-Structured Neural Network}
\newacronym{pde}{PDE}{Partial Differential Equation}
\newacronym{anfis}{ANFIS}{Adaptive Neuro-Fuzzy Inference System}
\newacronym{cnn}{CNN}{Convolutional Neural Network}
\newacronym{kf}{KF}{Kalman Filter}
\newacronym{ukf}{UKF}{Unscented Kalman Filter}
\newacronym{lstm}{LSTM}{Long Short-Term Memory Network}
\newacronym{rbfn}{RBFN}{Radial Basis Function Network}
\newacronym{gru}{GRU}{Gated Recurrent Unit}
\newacronym{fir}{FIR}{Finite Impulse Response}
\newacronym{node}{NODE}{Neural Ordinary Differential Equation}
\newacronym{vi}{VI}{Variational Integrator}
\newacronym{vin}{VIN}{Variational Integrator Network}
\newacronym{fvin}{FVIN}{Forced Variational Integrator Network}
\newacronym{delan}{DeLaN}{Deep Lagrangian Network}
\newacronym{lnn}{LNN}{Lagrangian Neural Network}
\newacronym{hnn}{HNN}{Hamiltonian Neural Network}
\newacronym{rk4}{RK4}{4$^{\mathrm{th}}$ order Runge Kutta}
\newacronym{tcn}{TCN}{Temporal Convolutional Network}
\newacronym{sindy}{SINDy}{Sparse Identification of Nonlinear Dynam. Systems}
\newacronym{eqln}{EQLN}{Equation Learner Network}
\newacronym{gpr}{GPR}{Gaussian Process Regression}
\newacronym{umv}{UMV}{Unmanned Marine Vehicle}
\newacronym{sysid}{SysID}{System Identification}
\newacronym{felan}{FeLaN}{Floating-Base Deep Lagrangian Network}
\newacronym{rl}{RL}{Reinforcement Learning}
\newacronym{il}{IL}{Imitation Learning}
\newacronym{dof}{DoF}{Degrees of Freedom}
\newacronym{deeponet}{DeepONet}{Deep Operator Network}
\newacronym{no}{NO}{Neural Operator}
\newacronym{fno}{FNO}{Fourier Neural Operator}
\newacronym{gno}{GNO}{Graph Neural Operator}
\newacronym{kan}{KAN}{Kolmogorov-Arnold Network}
\newacronym{gnn}{GNN}{Graph Neural Network}
\newacronym{pino}{PINO}{Physics-Informed Neural Operator}
\newacronym{dct}{DCT}{Discrete Cosine Transform}
\newacronym{ode}{ODE}{Ordinary Differential Equation}
\newacronym{sde}{SDE}{Stochastic Differential Equation}
\newacronym{svm}{SVM}{Support Vector Machine}
\newacronym{vae}{VAE}{Variational Autoencoder}
\newacronym{vlm}{VLM}{Vision-Language Model}
\newacronym{ekf}{EKF}{Extended Kalman Filter}
\newacronym{vla}{VLA}{Vision-Language-Action Model}
\newacronym{vwm}{VWM}{Video World Model}
\newacronym{gan}{GAN}{Generative Adversarial Network}
\newacronym{bnn}{BNN}{Bayesian Neural Network}
\newacronym{rkhs}{RKHS}{Reproducing Kernel Hilbert Space}
\newacronym{gpu}{GPU}{Graphics Processing Unit}

\newcommand{\ie}{\textit{i.e.},}
\newcommand{\eg}{\textit{e.g.},}

\newcommand{\genPos}{\mathbf{q}}
\newcommand{\genAcc}{\ddot{\mathbf{q}}}
\newcommand{\genVel}{\dot{\mathbf{q}}}
\newcommand{\genMom}{\mathbf{p}}
\newcommand{\genMomDot}{\dot{\mathbf{p}}}
\newcommand{\genForce}{\boldsymbol{\tau}}
\newcommand{\inertiaMat}{\mathbf{M}}
\newcommand{\inertiaChol}{\mathbf{L}}
\newcommand{\Lagrangian}{\mathcal{L}}
\newcommand{\Hamiltonian}{\mathcal{H}}
\newcommand{\EKin}{\mathcal{T}}
\newcommand{\EPot}{\mathcal{V}}
\newcommand{\ESystem}{\mathcal{E}}
\newcommand{\EoMC}{\mathbf{c}}
\newcommand{\EoMG}{\mathbf{g}}
\newcommand{\Transp}{^\mathrm{\scriptscriptstyle T}}
\newcommand{\SEThree}{\mathrm{SE}(3)}

\newcommand{\loss}{\mathbf{l}}

\newcommand\BibTeX{{\rmfamily B\kern-.05em \textsc{i\kern-.025em b}\kern-.08em
T\kern-.1667em\lower.7ex\hbox{E}\kern-.125emX}}

\def\volumeyear{2026}

\begin{document}

\runninghead{The Authors}

\title{Embedding Physics Priors\\in Robot Learning: A Survey}


\author{Mattia Piccinini\affilnum{1}, Lucas Schulze\affilnum{2}, Alice Plebe\affilnum{3}, Matteo Saveriano\affilnum{3}, Thomas Beckers\affilnum{4},\\Yuan Gao\affilnum{1}, Oleg Arenz\affilnum{2}, Baha Zarrouki\affilnum{1}, Dingrui Wang\affilnum{1}, Finn Rasmus Schäfer\affilnum{1},\\Jan Peters\affilnum{2}, Johannes Betz\affilnum{1} and Gastone Pietro Rosati Papini\affilnum{3}}

\affiliation{\affilnum{1}Professorship of Autonomous Vehicle Systems, Technical University of Munich, 85748 Garching, Germany; Munich Institute of Robotics
and Machine Intelligence (MIRMI).\\
\affilnum{2}Department of Computer Science, Technical University of Darmstadt,
Germany.\\
\affilnum{3}Department of Industrial Engineering, University of Trento, 38123 Trento, Italy.\\
\affilnum{4}Institute of Measurement, Control and Microtechnology, Ulm University, 89081 Ulm, Germany.}

\corrauth{Mattia Piccinini, Professorship of Autonomous Vehicle Systems, Technical University of Munich, 85748 Garching, Germany; Munich Institute of Robotics
and Machine Intelligence (MIRMI).}

\email{mattia.piccinini@tum.de}

\begin{abstract}
The rapid progress of artificial intelligence is reshaping robotics and accelerating the adoption of learning-based approaches. While purely data-driven methods have achieved remarkable success in computer vision and natural language processing, robotics remains constrained by limited data, complex real-world interactions, and the need for reliable operation. These challenges have motivated the exploration of physics-embedded robot learning, which embeds physics priors into learning algorithms. By encoding the underlying physical laws and constraints, physics priors can complement limited data with robotics-specific inductive biases, potentially improving generalization, interpretability, and sample efficiency. However, the literature on physics-embedded robot learning remains fragmented across terminology, methodologies, and application domains, making it difficult to assess this growing body of work. This survey reviews physics-embedded robot learning across a broad range of physics priors, robotics applications, and machine learning models, from single-layer perceptrons to generative foundation models. We adopt a unified taxonomy that classifies existing approaches according to their physics embedding: physics-guided inputs, data, and representations; physics-encoded model architectures; and physics-informed training loss functions. Building on this taxonomy, we review methods for robot dynamics learning, trajectory planning, prediction, control, and estimation, together with the corresponding open-source software ecosystem. We identify key open challenges, and outline promising future research directions. Overall, we argue that physics priors provide a particularly relevant robotics-specific inductive bias, complementing rather than replacing data-driven learning, and paving the way toward more generalizable, data-efficient, and trustworthy robotic systems. All reviewed papers, classification and search methods, tables, and software are available in a continuously updated public repository: \href{https://github.com/TUM-AVS/survey-physics-embedded-robot-learning}{https://github.com/TUM-AVS/survey-physics-embedded-robot-learning}.
\end{abstract}


\maketitle

\section{Introduction}
\label{sec:intro}

\begin{figure*}[t!]
    \centering
    \includegraphics[width=1\linewidth]{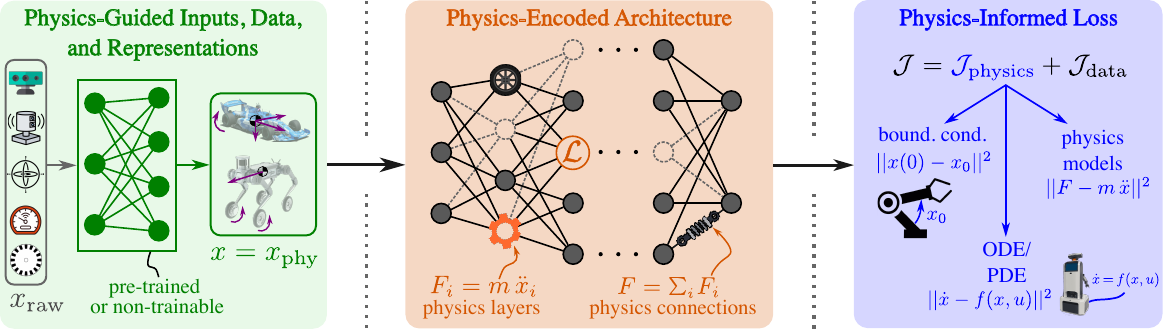}
    \caption{Different levels of embedding physics priors in robot learning: (a) physics-guided inputs, data, and representations; (b) physics-encoded architectures; (c) physics-informed loss functions. While most existing works focus on only one of these levels, we argue that the joint use of complementary routes may enable richer exploitation of prior physical knowledge, but systematic comparisons remain limited.}
    \label{fig:overview}
\end{figure*}

The remarkable success of \gls{ml}, from deep learning to foundation models, has shown that scaling data and computation can unlock unprecedented performance across diverse domains, including computer vision, natural language processing, and strategic games \citep{He2016-yr,Brown2020_GPT3,Schrittwieser2020,Vinyals2019,Jumper2021}. In robotics, however, collecting training data is expensive due to the cost of real-world experiments, specialized hardware, human supervision, and safety constraints. Furthermore, robots are often high-dimensional systems: for example, a dexterous robot hand may have more than 20 \gls{dof}, resulting in complex dynamics with nonlinearities, underactuation, and hybrid contact modes. Combined with the scarcity of large-scale open datasets and the need for real-time operation in safety-critical environments, these challenges make it difficult to rely solely on robot data and computation while ensuring safe generalization to unseen scenarios \citep{Billard2025,Goldberg2025DataGap}.

To address these challenges, the robotics community is increasingly exploring the embedding of \textbf{\textit{physics priors}} in \gls{ml} \citep{Amato2025,Sivtsov2025}, hereafter referred to as \textbf{\textit{physics-embedded learning}}. By encoding knowledge of the underlying physical laws, physics priors can complement limited data to potentially improve sample efficiency, generalization, interpretability, safety, and physical consistency.
Rather than replacing data-driven learning, we argue that physics priors should serve as scalable \textit{inductive biases} that complement and enhance purely data-driven methods. This reflects a broader result of modern \gls{ml}, where the success of foundation models stems not only from scaling data and computation, but also from tailored architectural and training biases, like the attention mechanism \citep{Vaswani2017Attention}. Because robots are embodied agents governed by physical laws, physics provides a natural domain-specific inductive bias. Nevertheless, a trade-off remains: although stronger physics priors can improve interpretability and generalization, they may also constrain model expressiveness, limiting the representation of behaviors that deviate from the assumed physical laws \citep{Amato2025}.

In this survey, we review existing approaches for physics-embedded robot learning, discussing their benefits, limitations, and open challenges. To unify the fragmented terminology in the literature, we adopt and extend the taxonomy of \cite{Faroughi2024}, identifying three complementary levels of physics embedding in \gls{ml}: \textbf{\textit{physics-guided}} inputs, data, and representations; \textbf{\textit{physics-encoded}} model architectures; and \textbf{\textit{physics-informed}} loss functions (Fig.~\ref{fig:overview}).

\paragraph{Survey Organization:}
This survey is organized as follows. We first define the scope of the reviewed literature (Sec.~\ref{sec:scope_literature}), compare our work with existing surveys (Sec.~\ref{sec:related_surveys}), summarize our contributions (Sec.~\ref{sec:contributions}), and provide a historical overview of physics-embedded robot learning (Sec.~\ref{sec:history}). Sec.~\ref{sec:overview} then introduces the three main physics-embedding routes for robot learning that underpin the taxonomy of our survey. We subsequently review physics-encoded architectures (Sec.~\ref{sec:encoded_architecture}), which constitute the largest body of work in the field, followed by physics-informed methods (Sec.~\ref{sec:physics_informed}) and physics-guided approaches (Sec.~\ref{sec:guided_inputs}). Next, we overview the available open-source software tools (Sec.~\ref{sec:software_tools}), before discussing open research questions (Sec.~\ref{sec:open_questions}), outlining proposed future research directions (Sec.~\ref{sec:future_directions}), and concluding the survey (Sec.~\ref{sec:conclusions}).

\subsection{Scope of the Considered Literature} \label{sec:scope_literature}

Our survey focuses on \gls{ml} approaches that embed physics priors and are applied to robotics.

\begin{figure*}[]
    \centering
    \includegraphics[width=1\linewidth]{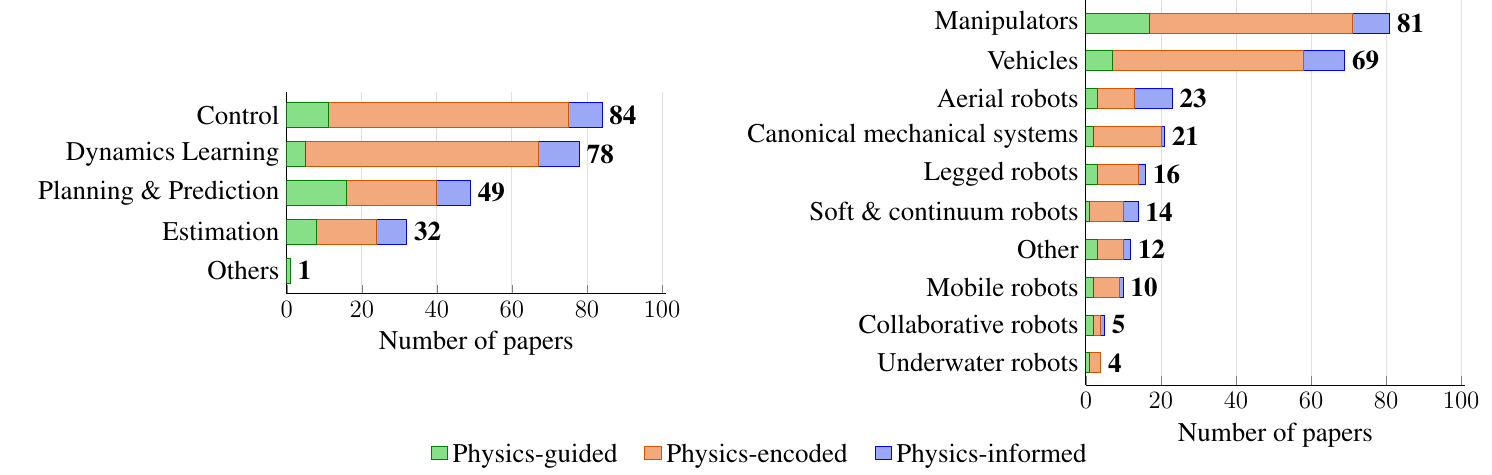}
    \caption{Distribution of the reviewed papers by application (left) and robot class (right), grouped by the three levels of physics embedding. In the robot class plot, ``canonical mechanical systems'' include pendulums, cart-poles, and other standard mechanical systems used for benchmarking. 
    Note that some papers are counted in more than one application or robot class, as they address multiple tasks or robot platforms. Some papers also employ two physics embedding routes.}
    \label{fig:paper_distribution}
\end{figure*}

\textit{Types of Physics Priors:} We consider a broad range of physics priors, grouped into five non-mutually exclusive families. (i) \textit{Governing equations}: Newton-Euler, Euler-Lagrange, Hamiltonian, and port-Hamiltonian formulations, together with the \glspl{ode} and \glspl{pde} describing rigid-body and continuum systems. (ii) \textit{Conservation laws, symmetries, and invariances}: conservation of energy, momentum, and power, together with symmetry, invariance, and equivariance principles (\eg{} $\SEThree$, $\mathrm{SO}(3)$, and morphological symmetries). (iii) \textit{Geometric and kinematic structure}: manifolds and Lie groups, kinematic trees, subsystem decompositions, connectivity and modularity of multi-body systems, and holonomic and nonholonomic constraints. (iv) \textit{Constitutive and interaction models}: friction, contact and impact laws, stiffness, damping, material behavior, and actuator and drivetrain dynamics. (v) \textit{Physical consistency and admissibility}: positive definiteness of inertia matrices, positive semi-definiteness of damping matrices, physically meaningful parameter bounds and sign constraints, passivity, dissipativity, stability properties,  actuator limits, and boundary conditions. Generic mathematical representations alone are not considered physics priors unless they explicitly encode physical knowledge.

We review how these priors are embedded into \gls{ml} approaches, with a
classification of physics-guided inputs, data, and representations, physics-encoded model architectures, and physics-informed loss functions.

\textit{Machine Learning Models and Methods:} While most reviewed works employ \glspl{nn}, our survey also covers \gls{gpr}, kernel methods, sparse identification and symbolic regression, equation learning, Koopman models, \glspl{no}, variational integrator networks,  generative models (including diffusion models, \glspl{vla}, and \glspl{vwm}), and other learning paradigms, whenever they employ mechanisms to embed physics priors.
We exclude \gls{rl} approaches in which physics is incorporated exclusively through \gls{rl}-specific mechanisms, such as state or action space design, exploration strategies, safety constraints, and simulator or environment augmentation, as these were reviewed by \cite{Banerjee2025}. \gls{rl} methods are instead included whenever physics is embedded through one of our taxonomy mechanisms, namely physics-guided inputs, data, or representations, physics-encoded model architectures, or physics-informed training objectives.

\textit{Robotics Applications \& Domains:} We cover a broad range of robotics problems, grouped into four application categories. (i) \textit{Dynamics learning}: forward and inverse dynamics, rigid-, soft-, and multi-body system identification, friction and contact modeling, continuum-robot shape learning, and equation or governing-law discovery. (ii) \textit{Trajectory planning and prediction}:  path and motion planning, motion and video prediction, trajectory imitation, planning-oriented policy generation, geometric planning on manifolds, and generative action prediction. (iii) \textit{Control}:  trajectory and path tracking, inverse-dynamics control, energy-shaping and passivity-based control. (iv) \textit{Estimation}: state and parameter estimation, localization, disturbance and force estimation, fault detection, and condition monitoring. These applications are used in the summary tables of our survey to classify the reviewed papers.

Robot types include manipulators, mobile robots, vehicles, legged robots, soft robots, collaborative robots, underwater and aerial robots. 

Fig.~\ref{fig:paper_distribution} shows the distribution of the reviewed papers across the considered applications (left) and robot classes (right), grouped by our three levels of physics embedding: physics-guided inputs, data, and representations; physics-encoded architectures; and physics-informed loss functions. 

\textit{Literature Sources:} We review papers published until August 2026, including peer-reviewed journal and conference contributions. Due to the rapidly evolving nature of the field, we also include a few relevant arXiv preprints that have not yet appeared in peer-reviewed venues but contribute significantly to the state of the art. 
Because the literature does not follow a unified taxonomy, we searched Google Scholar across multiple categories of physics embedding, using a set of keywords reported in our online repository. 


\textit{Open-Source Repository:} To facilitate reproducibility and continuous updates by the community, we maintain an online open-source repository that collects all reviewed papers, including their classification according to the taxonomy of physics embedding. The repository also provides a decision flow for classifying new papers, enabling the community to contribute with new publications, beyond the scope of this survey. The repository is available at \href{https://github.com/TUM-AVS/survey-physics-embedded-robot-learning}{https://github.com/TUM-AVS/survey-physics-embedded-robot-learning}.

\subsection{Related Surveys} \label{sec:related_surveys}
\begin{table*}[h!]
  \small
  \setlength{\tabcolsep}{2pt}
  \renewcommand{\arraystretch}{1.12}
  \begin{threeparttable}
  \centering
  \caption{Analysis of the most relevant surveys on embedding physics or model-based priors into learning methods. Our survey is the first to cover physics-guided inputs/data, physics-encoded architectures, and physics-informed losses in the context of robot learning, including a broad range of \gls{ml} models, physics priors, and robotics applications.} 
  \renewcommand{\tabularxcolumn}[1]{m{#1}}
  \begin{tabularx}{\textwidth}{>{\centering\arraybackslash}m{3.2cm}|>{\centering\arraybackslash}m{2.1cm}|>{\centering\arraybackslash}m{2.5cm}|>{\centering\arraybackslash}m{2.5cm}|>{\centering\arraybackslash}X|>{\centering\arraybackslash}m{1.9cm}}
    \toprule
    \textbf{Survey Paper} & \textbf{\begin{tabular}{@{}c@{}}Physics-Guided\\Inputs / Data\end{tabular}} & \textbf{\begin{tabular}{@{}c@{}}Physics-Encoded\\Architectures\end{tabular}} & \textbf{\begin{tabular}{@{}c@{}}Physics-Informed\\Losses\end{tabular}} & \textbf{\begin{tabular}{@{}c@{}}Machine Learning\\Models\tnote{1}\end{tabular}} & \textbf{\begin{tabular}{@{}c@{}}Physics Priors\\in Robotics\tnote{2}\end{tabular}} \\
    \hline
    \cite{shlezinger2023model} &  & \checkmark &  & \gls{mlp}, \gls{rnn}, \gls{vae}, \gls{cnn}, \gls{gan} &  \\
    \hline
    \cite{Cuomo2022} &  &  & \checkmark & \gls{mlp}, \gls{rnn}, \gls{lstm}, \gls{bnn}, \gls{cnn}, \gls{gan}, \gls{no} &  \\
    \hline
    \cite{Goswami2023} &  &  & \checkmark & \gls{no} &  \\
    \hline
    \cite{Faroughi2024} & \checkmark & \checkmark & \checkmark & \gls{mlp}, \gls{rnn}, \gls{lstm}, \gls{gan}, \gls{cnn}, \gls{vae}, \gls{no} &  \\
    \hline
    \cite{Karniadakis2021} & \checkmark & \checkmark & \checkmark & \gls{mlp}, \gls{rnn}, \gls{bnn}, \gls{cnn}, \gls{gan}, \gls{gpr}, \gls{no} &  \\
    \hline
    \cite{Geist2021structured} &  & \checkmark &  & \gls{mlp}, \gls{gpr}, analytical parametric \glspl{nn}, analytical residual models & \checkmark (16/108)\tnote{2} \\
    \hline
    \cite{watson2025machine} &  &  & \checkmark & \gls{mlp}, \gls{rnn}, \gls{delan}, \gls{lnn}, \gls{hnn}, \gls{node}, \gls{no}, Diffusion & \checkmark(21/359)\tnote{2} \\
    \hline
    \cite{Sivtsov2025} &  &  & \checkmark & \gls{mlp}, \gls{rnn}, \gls{lstm}, \gls{gan}, \gls{cnn}, Liquid \gls{nn}, Transformers & \checkmark (16/112)\tnote{2}\\ \hline
    \cite{Li2026worldmodels} & \checkmark &  &  & \gls{vwm} & \checkmark (6/262)\tnote{2} \\
    \hline
    \textbf{Ours} & \checkmark & \checkmark & \checkmark & \gls{delan}, \gls{lnn}, \gls{hnn}, \gls{msnn}, \gls{node}, \gls{vin}, \gls{eqln}, \gls{sindy}, \gls{mlp}, \gls{rnn}, \gls{lstm}, \gls{no}, \gls{gpr}, \gls{rkhs}, \gls{anfis}, \gls{vwm}, Diffusion, \gls{vla} & \checkmark (232/329)\tnote{2} \\
    \bottomrule
  \end{tabularx}
  \label{tab:related_surveys}
  \begin{tablenotes}
    \item[1] Acronyms for \gls{ml} model names are defined in Sec.~\ref{sec:list_of_acronyms}. 
    \item[2] Number of references embedding physics priors in robot learning methods over the total number of references in each survey.
  \end{tablenotes}
  \end{threeparttable}
\end{table*}
Table~\ref{tab:related_surveys} summarizes the most relevant surveys on embedding physics and model-based priors into machine learning. Below, we briefly review their scope, strengths, and limitations relative to our survey.
\paragraph{Surveys in Other Domains:}
\cite{shlezinger2023model} surveyed neural architectures combining deep learning and model-based methods for signal processing and communication networks, yet without considering physics priors or robotics. \cite{Cuomo2022} reviewed \glspl{pinn} and \glspl{no} for solving \glspl{pde}, including soft and hard constraints for boundary conditions, but did not cover structured \gls{ml} architectures or robotics. 
\cite{Goswami2023} reviewed three neural operator architectures and their physics-informed extensions, showing how structured loss functions enable accurate modeling of computational mechanics problems without labeled data. However, the survey focuses on porous media, fluid, and solid mechanics, without covering robotics.
\cite{Faroughi2024} reviewed physics-informed loss functions, physics-encoded architectures, \glspl{no}, and physics-guided data generation, but focused on fluid and solid mechanics rather than robotics. 
Similarly, \cite{Karniadakis2021} surveyed physics priors embedded through loss functions, architectures, operators, and datasets for scientific computing applications governed by \glspl{pde}, without addressing robotics.

\paragraph{Surveys Related to Robotics:}
\cite{Geist2021structured} reviewed structured learning for rigid-body dynamics modeling, including certain robotic systems. However, the survey is limited to learning inertial and force terms in Newton--Euler and Lagrangian formulations, with a limited scope of \gls{ml} models, and without covering robot control, trajectory planning, or state estimation.
\cite{watson2025machine} surveyed physics-informed \gls{ml} for prediction, forecasting, and system identification, but included very few robotics applications.
\cite{Sivtsov2025} reviewed \glspl{pinn} across multiple application domains, including robotics. However, robotics accounts for only 16 of its 112 references. Moreover, although the survey mentions physics-aware \gls{nn} architectures, it only discusses conventional recurrent, convolutional, and transformer architectures, rather than models explicitly designed to encode physical knowledge. It also omits physics-guided and physics-encoded learning, Lagrangian and Hamiltonian \glspl{nn}, \glspl{no}, generative models, and other physics-embedded learning paradigms.
\cite{Li2026worldmodels} surveyed \glspl{vwm} for embodied \gls{ai}. While emphasizing physical consistency as an evaluation criterion and discussing a few representative physics-aware world models, it does not review physics-embedded learning or systematically analyze how physics priors are incorporated through model architectures, loss functions, training procedures, and physics-guided data generation.

Adjacent to our scope, \cite{Tsuji2026} surveyed \gls{il} for contact-rich robotic tasks, organizing the literature by demonstration collection, sensing modality, and learning approach. While  learning under complex contact dynamics is one of the applications covered in our review, the perspectives of the two surveys are complementary: \cite{Tsuji2026} reviews methods that compensate for unavailable physical models through human demonstrations, whereas we review methods that embed available physics priors through inputs and data, model architectures, or training objectives. Accordingly, \gls{il} in contact-rich manipulation falls within our scope only when physics is embedded through one of these three mechanisms.

\subsection{Contributions} \label{sec:contributions}

To the best of our knowledge, none of the previously discussed surveys provides a structured review of the broad spectrum of approaches embedding physics priors into robot learning methods. Our survey aims to fill this gap with the following contributions:
\begin{itemize}
  \item A structured review on embedding physics priors into robot learning methods and applications. We cover a wide range of \gls{ml} models, from single-layer perceptrons to generative models, and a broad spectrum of physics priors, from Newton-Euler and Lagrangian dynamics to conservation laws, symmetries, and invariances.
  \item A classification of the existing literature into three categories, based on where physics priors are embedded: inputs, data, and representations (physics-guided), internal architectures (physics-encoded), and loss functions (physics-informed). 
  \item A categorization of the corresponding applications into robot dynamics learning, trajectory planning and prediction, control, and estimation, and an overview of the open-source software tools to support physics-embedded robot learning.
  \item An identification of the main challenges and open research questions. Based on our analysis, we propose future research directions to advance the field.
\end{itemize}
%
%
%
\subsection{Historical Perspective: From Analytical Models to Physics-Embedded Learning} \label{sec:history}

\paragraph{Analytical Models:}
For decades, robotics has relied primarily on analytical models derived from kinematics, rigid-body dynamics, and optimal control to describe and control physical systems \citep{SicilianoKhatib2016}. These models encode strong physics priors through kinematic and dynamic equations, conservation laws, and geometric constraints, providing interpretable representations and, in some cases, analytical guarantees for planning and control. However, their accuracy is limited by simplifying assumptions, unmodeled dynamics, and the complexity of real-world environments.

\paragraph{Data-Driven Robot Learning:}
Learning-based methods later emerged as a complementary paradigm, using data to model phenomena that are difficult to capture analytically. Early applications employed \glspl{nn} for robot control and dynamics modeling, including Cerebellar Model Arithmetic Computer (CMAC)-based controllers for manipulation and bipedal locomotion \citep{cmac_1979,cmac_manipulator_1987,MILLER199717_dynamic_walk}, and \glspl{rbfn} for adaptive robot control \citep{Broomhead1988MultivariableFI,sanner_slotine_stable_1995}. Learning from demonstration further reduced the need for manually designed task models, while ALVINN \citep{Pomerleau_alvinn} was an early example of end-to-end autonomous driving from sensor observations. The scalability of deep learning later marked a major advance in robot learning. Following the success of deep \glspl{cnn} \citep{NIPS2012_c399862d}, robotics adopted end-to-end visuomotor policies \citep{jmlr_e2e_deep_levine} and \gls{rl} for complex manipulation \citep{openai2019solvingrubikscuberobot}. More recently, transformers, diffusion models, and foundation models have further expanded the capabilities and generality of robot learning \citep{Vaswani2017Attention,diffusion_policy_2023}, paving the way for \glspl{vla}, \glspl{vwm}, and other multimodal approaches. 

\paragraph{Incorporating Domain Priors:}
The evolution of \gls{ml} has repeatedly shown that scalable, tailored inductive biases embedded in \gls{ml} architectures can improve learning efficiency without sacrificing expressive power. For example, \glspl{cnn} encode translation invariance for image recognition, while \gls{lstm} recurrent architectures have specialized gating mechanisms to encode sequential dependencies. 
Since robots are embodied systems governed by physical laws, physics provides a natural robotics-specific inductive bias. This has motivated methods that embed physics priors directly into \gls{ml} algorithms. Early examples include \cite{Nguyen2010}, who incorporated a rigid-body model as the mean function and kernel of a Gaussian process for manipulator inverse dynamics learning, while \cite{Cheng2016_RKHS} encoded the Lagrangian structure of the dynamics in a reproducing kernel Hilbert space. Shortly thereafter, \cite{Ledezma2017,Ledezma2018} constructed network topologies from Newton--Euler operators, anticipating the physics-encoded architectures later formalized by \gls{delan} \citep{delan_lutter2019}. These early ideas have since evolved into the physics-guided, physics-encoded, and physics-informed paradigms reviewed in this survey.

\section{Physics-Guided, Physics-Encoded, and Physics-Informed Robot Learning: Overview and Organization} \label{sec:overview}
\begin{figure}[t]
    \centering
    \includegraphics[width=0.95\linewidth]{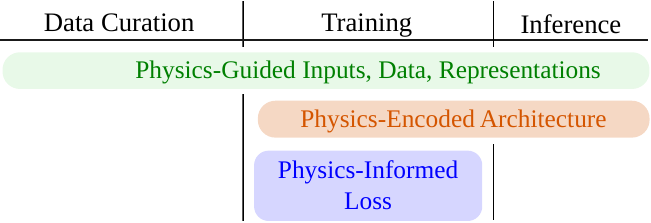}
    \caption{Lifecycle of physics priors in robot learning. Physics-guided components transform or enrich input features and data, either during data curation or as pre-processing modules that remain active at training and inference time. Physics-encoded priors are embedded in the model architecture, and remain active during training and inference. Physics-informed components embed physical knowledge via the training loss, and are active only during training.}
    \label{fig:lifecycle}
\end{figure}
\definecolor{guided}{HTML}{87e087}
\definecolor{guidedDark}{HTML}{008000}
\definecolor{encoded}{HTML}{d45500}
\definecolor{encodedDark}{HTML}{d45500}
\definecolor{informed}{HTML}{0000ff}
\definecolor{informedDark}{HTML}{0000ff}
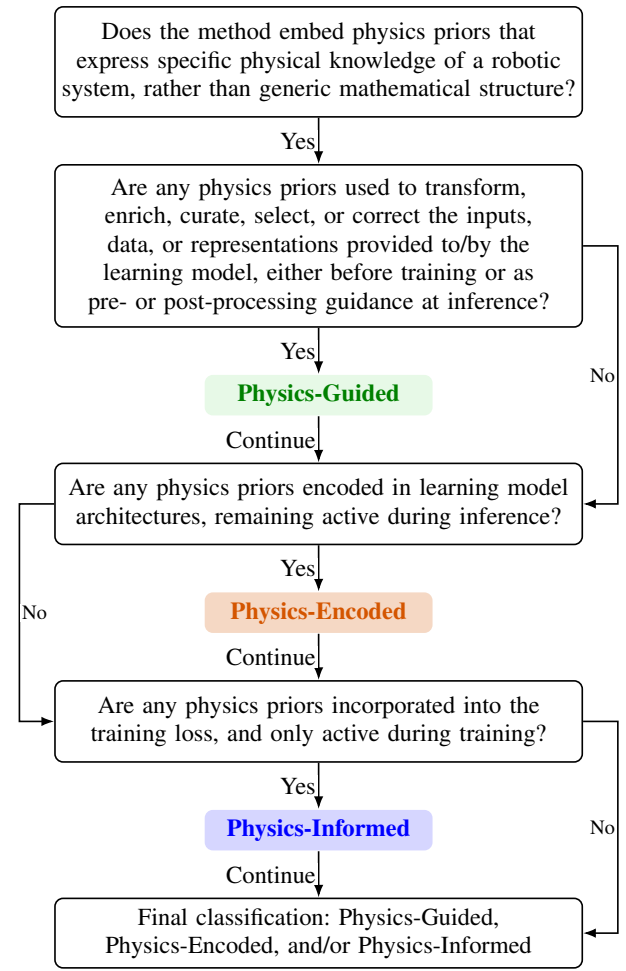
\begin{figure}[t]
\centering
\resizebox{0.95\linewidth}{!}{
\begin{tikzpicture}[
    node distance=0.55cm and 2.4cm,
    font=\large,
    decision/.style={
        rectangle,
        rounded corners,
        draw,
        thick,
        align=center,
        minimum width=9cm,
        minimum height=1.8cm,
        inner sep=1.5pt,
        text width=9.0cm,
    },
    process/.style={
        rectangle,
        rounded corners,
        draw,
        thick,
        align=center,
        minimum width=9cm,
        minimum height=1.8cm,
        inner sep=1.5pt,
        text width=9.0cm
    },
    class/.style={
        rectangle,
        rounded corners,
        thick,
        align=center,
        minimum height=0.7cm,
        text width=3.7cm
    },
    arrow/.style={
        -{Latex[length=2.2mm]},
        thick
    },
    label/.style={
        midway,
        fill=white,
        inner sep=1pt,
    }
]



\node[decision, minimum height=1.9cm] (scope)
{Does the method embed physics priors that express specific physical knowledge of a robotic system, rather than generic mathematical structure?};

\node[decision, below=0.8cm of scope, minimum height=2.8cm] (guided)
{Are any physics priors used to transform, enrich, curate, select, or correct the inputs, data, or representations provided to/by the learning model, either before training or as pre- or post-processing guidance at inference?
};

\node[class, below=0.8cm of guided, fill=guided!20] (pg) 
{\textcolor{guidedDark}{\large\textbf{Physics-Guided}}};

\node[decision, below=0.8cm of pg, minimum height=1.4cm] (encoded)
{Are any physics priors encoded in learning model architectures, remaining active during inference?};

\node[class, below=0.8cm of encoded, fill=encoded!23] (pe) 
{\textcolor{encodedDark}{\large\textbf{Physics-Encoded}}};

\node[decision, below=0.8cm of pe, minimum height=1.4cm] (informed)
{Are any physics priors incorporated into the training loss, and only active during training?};

\node[class, below=0.8cm of informed, fill=informed!16] (pi) 
{\textcolor{informedDark}{\large\textbf{Physics-Informed}}};

\node[process, below=0.8cm of pi, minimum height=1.2cm] (datadriven)
{Final classification: Physics-Guided,   Physics-Encoded, and/or Physics-Informed};

\coordinate (m1) at ($(guided.south)!0.5!(encoded.north)$);
\coordinate (m2) at ($(encoded.south)!0.5!(informed.north)$);
\coordinate (m3) at ($(informed.south)!0.5!(datadriven.north)$);


\draw[arrow] (scope.south) -- node[label, left] {Yes} (guided.north);

\draw[arrow] (guided.south) -- node[label, left] {Yes} (pg.north);
\draw[arrow] (pg.south) -- node[label, left] {Continue} (encoded.north);

\draw[arrow] (encoded.south) -- node[label, left] {Yes} (pe.north);
\draw[arrow] (pe.south) -- node[label, left] {Continue} (informed.north);

\draw[arrow] (informed.south) -- node[label, left] {Yes} (pi.north);
\draw[arrow] (pi.south) -- node[label, left] {Continue} (datadriven.north);

\draw[arrow] (guided.east) -- ++(0.6,0)
    |- node[pos=0.25, left, fill=white, inner sep=1pt, font=\normalsize] {No}
    (encoded.east);

\draw[arrow] (encoded.west) -- ++(-0.6,0)
    |- node[pos=0.25, right, fill=white, inner sep=1pt, font=\normalsize] {No}
    (informed.west);

\draw[arrow] (informed.east) -- ++(0.6,0)
    |- node[pos=0.25, left, fill=white, inner sep=1pt, font=\normalsize] {No}
    (datadriven.east);




\end{tikzpicture}
}
\caption{Decision flow to classify physics-embedded robot learning approaches.
Methods that embed only generic mathematical structure, or that are not applied
to robotic systems, do not satisfy the first criterion and fall outside the scope
of this survey. The three categories (physics-guided, physics-encoded,
physics-informed) are not mutually exclusive: multiple labels may be assigned by
sequentially evaluating all criteria.}
\label{fig:physics_prior_classification_flow}
\end{figure}
\paragraph{Taxonomy:}
As anticipated in the introduction, we adapt the taxonomy of \cite{Faroughi2024} to classify physics-embedded robot learning into three categories, illustrated in Fig.~\ref{fig:overview}. In our survey, \textit{Physics-guided learning} exploits physics priors to transform, enrich, curate, select, or correct the inputs, data, or representations of learning models, either before training or as pre- or post-processing guidance at inference. 
\textit{Physics-encoded learning} embeds physics directly into model
architectures, via tailored structures, layers, and topologies, governing
equations and structure-preserving integrators, geometric and kinematic
structure, energy and conservation principles, symmetries and invariances,
architectural constraints, or the composition of learnable components with
analytical physics-based models.
Finally, \textit{Physics-informed learning} incorporates physics into the training objective, typically through regularization terms or residual losses derived from governing physical equations. Together, these three categories define \textit{physics-embedded learning}, whose goal is to complement data with the available physics priors.

Depending on the application, our taxonomy may be applied to the entire robot learning pipeline, an individual learning model, or a specific submodule within a larger architecture.

\paragraph{Lifecycle of Physics Priors:}
Physics priors can be incorporated at different stages of the learning lifecycle, as illustrated in Fig.~\ref{fig:lifecycle}. We split the lifecycle into three stages: (i) data curation, (ii) model training, and (iii) inference.

Physics-guided components transform, curate, or enrich input features and data through physically meaningful representations. They may be applied during data curation, or implemented as pre-processing modules that remain part of the training and inference pipelines. In the latter case, any learnable quantities of the pre-processing modules are pre-trained or frozen when training the main model (Fig.~\ref{fig:overview}).

Physics-encoded components embed physical knowledge directly into the model architecture, and therefore remain active during both training and inference.

Finally, physics-informed residual loss terms are evaluated only during training.  Their influence persists at inference only through the learned model parameters, but they do not affect the model architecture or input features (Fig.~\ref{fig:lifecycle}).

\paragraph{Classification Flow:}
To classify existing methods, we adopt the decision flow in Fig.~\ref{fig:physics_prior_classification_flow}. The three categories (physics-guided, physics-encoded, physics-informed) are not mutually exclusive, and a method may belong to multiple categories. The same decision flow is implemented in our open-source repository to continuously classify newly published methods.

\section{Physics-Encoded Architectures} \label{sec:encoded_architecture}

A first approach to embedding physics priors in robot learning is to design model architectures that explicitly encode physical principles and constraints. This introduces inductive biases aligned with the underlying dynamics, with the potential to improve interpretability, generalization, and sample efficiency.

Following \cite{Faroughi2024}, we refer to \textit{physics-encoded \glspl{nn}} as models that ``forcibly encode known physics into their core architecture''. We extend this concept beyond \glspl{nn}, to encompass \gls{ml} models that embed physical principles directly into their architecture. In our survey, physics-encoded architectures incorporate physical insights through tailored internal structures, layers, topologies, equations, architectural constraints, energy, invariance and symmetry principles, extending or augmenting physics-based models with robotics domain knowledge. 

Physics-encoded architectures account for the majority of the surveyed methods (71\%), with most applications in manipulators and vehicles (Fig.~\ref{fig:paper_distribution}). Their development spans the last decade, accelerating markedly since 2020. 
%
%

\begin{figure}[t!]
    \centering
    \includegraphics[width=1\linewidth]{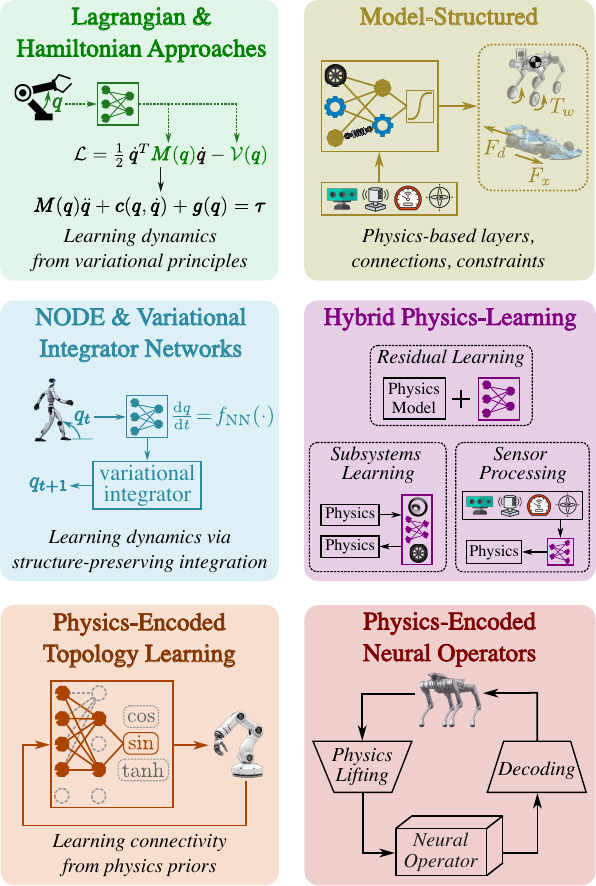}
    \caption{Sub-categories of physics-encoded robot learning architectures (Sec. \ref{sec:encoded_architecture}).} 
    \label{fig:physics_encoded}
\end{figure}

Our review covers six non-mutually exclusive classes, (Fig.~\ref{fig:physics_encoded}): Lagrangian and Hamiltonian learning models (Sec. \ref{sec:lagrangian} and \ref{sec:hamiltonian}), model-structured architectures (Sec.~\ref{sec:msnn}), neural \glspl{ode} and variational integrator networks (Sec.~\ref{sec:node_vi}), hybrid physics-neural architectures (Sec.~\ref{sec:hybrid_physics_nn}), physics-encoded topology learning (Sec.~\ref{sec:topology}), and physics-encoded neural operators (Sec.~\ref{sec:neural_operators_encoded}). 
Additional approaches are discussed in Sec.~\ref{sec:other_encoded}. 

\subsection{Lagrangian Learning Models} \label{sec:lagrangian}

The motion laws of mechanical systems can be formulated using Lagrangian mechanics. 
This formalism defines the Lagrangian $\Lagrangian$ as a function of the system's generalized coordinates $\genPos$ and their time derivatives $\genVel$. For mechanical systems such as robots, $\Lagrangian$ is typically defined as
\begin{equation}
    \label{eq:lag_definition}
    \Lagrangian(\genPos,\genVel) = \EKin(\genPos,\genVel) - \EPot(\genPos) =  \frac{1}{2}\,\genVel\Transp\,\inertiaMat(\genPos)\,\genVel - \EPot(\genPos),
\end{equation}
where $\genVel$ are the generalized velocities; $\EKin$ and $\EPot$ are the kinetic and potential energies, respectively; and $\inertiaMat(\genPos)$ is the inertia matrix. The kinetic energy is quadratic in the velocity $\genVel$ for the class of robotic systems considered in this survey, which include rigid-body, soft, and continuum robots. $\inertiaMat(\genPos)$ is a symmetric and positive definite matrix ($\inertiaMat(\genPos) \succ 0$), as any nonzero velocity results in a positive $\EKin$.
Non-conservative forces $\genForce$ are related to the Lagrangian via the Euler-Lagrange equation \citep{Wit1996}
\begin{equation}
    \label{eq:eom}
    \frac{d}{dt}\frac{\partial \Lagrangian}{\partial \genVel}
    - \frac{\partial \Lagrangian}{\partial{\genPos}} = \genForce.
\end{equation}
Substituting \eqref{eq:lag_definition} into \eqref{eq:eom} and using $\frac{\partial}{\partial \genPos}\EPot(\genPos)=\EoMG(\genPos)$, where $\EoMG(\genPos)$ denotes the generalized gravity forces, yields
\begin{equation}
    \label{eq:eom_compact}
    \inertiaMat(\genPos)\, \genAcc + \EoMC(\genPos,\genVel) + \EoMG(\genPos) = \genForce,
\end{equation}
where $\EoMC(\genPos,\genVel)$ comprises Coriolis and centripetal forces. 

Next, we review the main approaches that embed Lagrangian mechanics into learning architectures, summarized in Table~\ref{tab:lnn_hnn_summary}.

\begin{table*}[t]
\scriptsize
\centering
\setlength{\tabcolsep}{2pt}
\renewcommand{\arraystretch}{1.12}
\begin{threeparttable}
\caption{Summary of the reviewed papers on Lagrangian and Hamiltonian learning (Sec.~\ref{sec:lagrangian}-\ref{sec:hamiltonian}), grouped by application category.}
\label{tab:lnn_hnn_summary}
\renewcommand{\tabularxcolumn}[1]{m{#1}}
\begin{tabularx}{\textwidth}{>{\centering\arraybackslash}m{3.0cm}|>{\centering\arraybackslash}m{2.7cm}|>{\centering\arraybackslash}m{2.2cm}|>{\centering\arraybackslash}X|>{\centering\arraybackslash}m{1.9cm}}
\toprule
\textbf{Paper} & \textbf{Application} & \textbf{Robot type} & \textbf{Additional physical principles} & \textbf{Method} \\
\hline
\multicolumn{5}{c}{\rule{0pt}{2.5ex}\textbf{\textit{Dynamics Learning}}}\\
\hline
\cite{lnn_2020, modlanet_se3_cartesian_map} & \multirow{9}{*}{\gls{sysid}} & Pendulum &  & \gls{lnn} \\
\cline{1-1}\cline{3-5}
\cite{delan_induced_sparsity_Lahoud_2025} &  & Parallel robot & Sparsity from decoupled translational and rotational inertia & \multirow{6}{*}{\gls{delan}}\\
\cline{1-1}\cline{3-4}
\cite{felan} &  & Quadrupeds / humanoids & Branch-induced sparsity, full physical consistency of the spatial inertia & \\
\cline{1-1}\cline{3-4}
\cite{delan_plus_ffnn} &  & \multirow{3}{*}{Manipulator} & & \\
\cline{1-1}\cline{4-4}
\cite{delan_non_sym_friction} &  &  & Viscous and non-symmetrical Coulomb friction, motor rotor inertia & \\
\cline{1-1}\cline{4-4}
\cite{delan_friction_iterative_2025} &  &  & Stribeck friction & \\
\cline{1-1}\cline{3-5}
\cite{hnn} &  & Pendulum &  & \multirow{2}{*}{\gls{hnn}} \\
\cline{1-1}\cline{3-4}
\cite{Jin2025_port_hnn_underwater} &  & Underwater vehicle & Actuator input matrix, dissipation & \\
\cline{1-1}\cline{3-5}
\cite{hnn_lnn_exp_const_se3} &  & Pendulum & & \gls{hnn}, \gls{lnn} \\
\hline
\cite{lnn_mbrl_pmlr_ramesh23a} & \multirow{3}{*}{Simulator for \gls{rl}} & Pendulum / cart-pole / acrobot &  & \multirow{3}{*}{\gls{delan}} \\
\cline{1-1}\cline{3-4}
\cite{lnn_inf_horizon_quadruped} &  & Quadruped & External force & \\
\cline{1-1}\cline{3-4}
\cite{mbrl_baxter} &  & Manipulator & Stribeck friction, kinematic model & \\
\hline
\cite{Beckers2023} & \multirow{5}{*}{Dynamics learning} & Hopping robot & Embedding energy-based structure of robot model & \multirow{3}{*}{Port-Hamiltonian \gls{gpr}}\\
\cline{1-1}\cline{3-4}
\cite{ali2026learningbasedmodelingsoftrobots} &  & Soft robotic manipulator & Embedding energy-based structure of \gls{pde} robot model & \\
\cline{1-1}\cline{3-5}
\cite{Smith2024} &  & Manipulator & Enforcing odd symmetry in the Hamiltonian vector field & Hamiltonian \gls{rkhs} \\
\hline
\cite{Libera2020,Giacomuzzo2022_MED,Giacomuzzo2024} & \gls{sysid} & Manipulator & Lagrangian polynomial structure embedded in \gls{gpr} kernel & Lagrangian \gls{gpr} \\
\hline
\multicolumn{5}{c}{\rule{0pt}{2.5ex}\textbf{\textit{Trajectory Planning and Prediction}}}\\
\hline
\cite{delan_exoskeleton, pilan_assit_ctc_exoskeleton} & Human-Exoskeleton cooperative control & Exoskeleton &  & \multirow{2}{*}{\gls{delan}} \\
\cline{1-4}
\cite{time_optimal_planning_delan} & Path Planning & Parallel robot / Manipulator & Coulomb friction & \\
\hline
\multicolumn{5}{c}{\rule{0pt}{2.5ex}\textbf{\textit{Control}}}\\
\hline
\cite{delan_lutter2019, delan_2021} & Trajectory tracking & \multirow{2}{*}{Manipulator} & & \multirow{4}{*}{\gls{delan}} \\
\cline{1-2}\cline{4-4}
\cite{cadelac_iros25} & Adaptive \gls{mpc} &  & & \\
\cline{1-4}
\cite{Gupta2019,Gupta2020} & Trajectory tracking & Cart-pole / pendulum / acrobot & Actuator input matrix, dissipative terms & \\
\cline{1-4}
\cite{delan_4ec} & Energy shaping & Cart-pole / pendulum & Friction & \\
\hline
\cite{lnn_hnn_soft_robots} & Trajectory tracking & Soft robot / manipulator & Actuator input matrix, non-collocated coordinates, dissipative terms & \gls{delan}, \gls{hnn} \\
\hline
\cite{Zhong2020Symplectic} & \multirow{4}{*}{Energy shaping} & Cart-pole / pendulum & Actuator input matrix & \multirow{4}{*}{\gls{hnn}} \\
\cline{1-1}\cline{3-4}
\cite{duong21_hnn_dyn_ctrl_se3} &  & \multirow{2}{*}{Quadrotor} & Actuator input matrix & \\
\cline{1-1}\cline{4-4}
\cite{port_hnn_lie_control} &  &  & Actuator input matrix, dissipative terms, air drag & \\
\cline{1-1}\cline{3-4}
\cite{Altawaitan2023HamiltonianDL} &  & Wheeled robot & Actuator input matrix, dissipative terms & \\
\hline
\cite{Evangelisti2022} & \multirow{3}{*}{Trajectory tracking} & Manipulator & Energy conservation, passivity, positive definiteness of inertia matrix & \multirow{3}{*}{Lagrangian \gls{gpr}} \\
\cline{1-1}\cline{3-4}
\cite{Evangelisti2026} &  & Manipulator, soft robotic manipulator & Energy conservation, passivity, Lagrangian structure preservation & \\
\hline
\cite{Cheng2016_RKHS} & Inverse dynamics learning & Holonomic robot & Polynomial \gls{rkhs} kernels to embed the Lagrangian structure & Lagrangian \gls{rkhs} \\
\hline
\multicolumn{5}{c}{\rule{0pt}{2.5ex}\textbf{\textit{Estimation}}}\\
\hline
\cite{delan_excavator} & Disturbance observer, trajectory tracking & Excavator & Hydraulic actuators modeling and stick-slip friction & \multirow{4}{*}{\gls{delan}} \\
\cline{1-4}
\cite{pinn_delan_contact_force} & \multirow{2}{*}{Contact force estimation} & Manipulator & Stribeck friction and joint Backlash & \\
\cline{1-1}\cline{3-4}
\cite{delan_h_net_grf_estimation} &  & Quadruped & Friction, hysteresis & \\

\bottomrule
\end{tabularx}
\end{threeparttable}
\end{table*}
 

\subsubsection{Deep Lagrangian Networks (DeLaN):} \label{sec:delan}
By embedding this formalism into a deep learning framework, \cite{delan_lutter2019,delan_2021} proposed \gls{delan}, which parametrize the inertial matrix $\inertiaMat(\genPos)$ and potential energy\footnote{The Coriolis and centripetal forces $\EoMC(\genPos,\genVel)$ are then computed from $\inertiaMat(\genPos)$ and $\genVel$ \citep{delan_lutter2019}, thus they do not need to be learned independently.} $\EPot(\genPos)$ through \glspl{nn} (typically \glspl{mlp}). 
To ensure physical consistency, \ie{} $\inertiaMat(\genPos) \succ 0$, \gls{delan} represent $\inertiaMat(\genPos)$ through its Cholesky factorization,
\begin{equation}
    \inertiaMat(\genPos) = \inertiaChol(\genPos)\;\inertiaChol(\genPos)\Transp\,,
\end{equation}
where $\inertiaChol(\genPos)$ is a learnable lower triangular matrix, whose diagonal elements are enforced to be positive. 
As the \glspl{nn} that parameterize $\inertiaMat(\genPos)$ and $\EPot(\genPos)$ take only $\genPos$ as input, the dependence of the resulting forward and inverse dynamics models on velocities and accelerations is analytically determined by the Lagrangian formulation \eqref{eq:eom_compact}, rather than learned from data, improving generalization to unseen velocities and accelerations.
By learning the Lagrangian terms, \gls{delan} improve energy conservation of unforced systems and long-term stability of learned dynamic models, while providing interpretable physical quantities such as $\inertiaMat(\genPos)$ and $\EPot(\genPos)$.

\paragraph{Training DeLaN for Forward and Inverse Dynamics:}
\gls{delan} do not only learn an energy model, but also provide forward and inverse dynamics models obtained via \eqref{eq:eom}. As a result, they can be trained using loss functions defined on any of these formulations, or on a combination of them.
This further enables their use within model-based frameworks based on \eqref{eq:eom_compact}. For example, \cite{delan_lutter2019,delan_2021} used \gls{delan} in an inverse-dynamics torque tracking controller for a robotic arm. Specifically,  they trained $\inertiaMat(\genPos)$ and $\EPot(\genPos)$ with supervised learning, by minimizing the deviations between the predicted and applied torques $\genForce$ in \eqref{eq:eom_compact}. Exploiting the flexibility of neural representations, \cite{cadelac_iros25} proposed the Context-Aware \gls{delan}, which conditions the system Lagrangian on a context latent variable inferred at runtime, thereby enabling online \gls{sysid} with a model trained only on simulated data. On top of this architecture, the learned model was used in an \gls{mpc} framework for zero-shot adaptive control of manipulation tasks. 
Based on \gls{delan}, \cite{pilan_assit_ctc_exoskeleton} performed lower-limb motion-intention perception on a hip-knee powered exoskeleton, predicting future joint states that serve as references for computed-torque control. Experimental results demonstrated smooth tracking and generalization across motion sequences. 

\paragraph{Encoding Sparsity in DeLaN:}
Parameterizing the dynamics via the inertia matrix also enables incorporating additional prior knowledge about input independence and sparsity. For example, \cite{delan_induced_sparsity_Lahoud_2025} directly enforced zeros in $\inertiaChol(\genPos)$ by exploiting the decoupling between the rotational and translational degrees of freedom of a parallel robot. To encode more general sparsity and input-independence structures, \cite{felan} adopted the reordered Cholesky factorization of \cite{efficient_factorization_featherstone2005}, \ie{} $\inertiaMat(\genPos) = \inertiaChol(\genPos)\Transp\inertiaChol(\genPos)$. This parametrization enables encoding sparsity patterns induced by branched kinematic chains, namely branch-induced sparsity.
Moreover, \cite{felan} used this structure to impose full physical consistency on the spatial inertia terms appearing in floating-base dynamics. Building on this parametrization, the authors introduced \gls{felan}, extending \gls{delan} to floating-base systems.

\paragraph{Modeling Non-Conservative Effects in DeLaN:}
The original \gls{delan} formulation does not impose any prior knowledge or structure on the non-conservative generalized forces $\genForce$, assuming that these forces are fully measured or that they are only the control inputs.
In practice, however, additional non-conservative effects may arise from several sources, e.g., actuator dynamics, friction, damping, or unmodeled contacts. The Euler-Lagrange equations yield the decomposed equations of motion \eqref{eq:eom_compact}, allowing additional non-conservative effects to be incorporated as learned torque contributions in the training loss. For example, \cite{delan_plus_ffnn} jointly learned \gls{delan} with a black-box \gls{mlp} that models residual torque contributions for manipulator \gls{sysid}.
Several extensions employ structured models for non-conservative effects.
\cite{Gupta2019,Gupta2020} extended the \gls{delan} framework by learning the dissipative and input forces in a structured way. Since most mechanical systems are control-affine, \ie{} the actuator inputs map linearly with the applied forces, they learned the control input Jacobian and bias terms with \glspl{mlp}, as well as the mass matrix and potential energy terms. Their methods were used to control a simulated double pendulum \citep{Gupta2019}, double Cartpole and Acrobot systems \citep{Gupta2020}. \cite{Gupta2020} showed improved generalization to unseen data compared to general-purpose \glspl{mlp}.

\paragraph{Friction Modeling in DeLaN:}
\cite{delan_4ec} learned \gls{delan} jointly with a white-box actuator-friction model, while promoting a smoother energy model through temporal coherence loss. The resulting model was used in an energy controller and evaluated on simulated and real cart-pole and inverted-pendulum systems.
\cite{delan_non_sym_friction} addressed the identification of non-symmetrical Coulomb friction using linear models, while also estimating viscous friction coefficients and rotor inertia. The authors validated the proposed architecture on a real manipulator and compared it against classical \gls{sysid} methods.
Similarly, \cite{time_optimal_planning_delan} extended \gls{delan} with Coulomb friction modeling and used the learned dynamics within a time-optimal path trajectory generator. The method is validated experimentally on real robotic platforms, including a 6-\gls{dof} parallel robot and a manipulator.
\cite{delan_friction_iterative_2025} combined \gls{delan} with a Stribeck friction model. Their method uses an iterative learning strategy to separate dynamics and friction torque components without requiring component-wise torque labels, followed by a temporal \gls{cnn} for residual torque compensation. The framework was validated for \gls{sysid} on robot manipulators.
To capture nonlinear phenomena due to hydraulic actuators and stick-slip friction, \cite{delan_excavator} combined a \gls{delan} with a \gls{cnn} and an \gls{lstm}. The learned model was integrated into an inverse dynamics controller for trajectory tracking. They further combined the controller with a disturbance \gls{kf}, demonstrating the synergy between the learned dynamics model and disturbance observer. The approach was evaluated on a scaled-down excavator platform, and using data from a real excavator.
In addition to learning a \gls{delan} with a Stribeck friction model, \cite{pinn_delan_contact_force} also estimated residual dynamic errors associated with backlash using Gaussian basis functions. Their model also provides online uncertainty estimates, which were incorporated into an adaptive \gls{kf} for contact force estimation on a real manipulator. Similarly, \cite{delan_h_net_grf_estimation} combined \gls{delan} with a hysteresis-aware \gls{tcn} to estimate ground reaction forces for a simulated quadruped.

\subsubsection{Lagrangian Neural Networks (LNNs):}\label{sec:lnn}
Introduced by \cite{lnn_2020}, \glspl{lnn} learn the Lagrangian $\Lagrangian$ directly as an unstructured \gls{nn} function of $\genPos$ and $\genVel$, rather than imposing the energy decomposition in \eqref{eq:lag_definition}. Unlike \gls{delan}, which assume the specific form \eqref{eq:lag_definition} of the Lagrangian and its kinetic energy (which works for certain classes of systems like rigid bodies), \glspl{lnn} can represent more general Lagrangians, enabling dynamics learning of charged particle systems and others. The generalized accelerations $\genAcc$ are recovered from the learned Lagrangian via \eqref{eq:eom}, and the model is trained by minimizing the error between predicted and measured accelerations. \cite{lnn_2020} showed that \glspl{lnn} preserve energy over longer time horizons than unconstrained \glspl{nn}, and applied \glspl{lnn} to simple mechanical systems like a double pendulum.

\cite{modlanet_se3_cartesian_map} proposed a modular version of \gls{lnn} by learning a coordinate transformation map based on the known kinematic tree.




Some authors used \gls{lnn} and \gls{delan} for model-based \gls{rl}, typically using the learned dynamics as a differentiable simulator to generate trajectories for policy optimization. An early example is \cite{lnn_mbrl_pmlr_ramesh23a}, which compared dynamics learning with a black-box \gls{mlp} and a \gls{delan}. They evaluated the method on pendulum, cart-pole, and acrobot systems, where \gls{delan} achieve better performance and sample efficiency. \cite{lnn_inf_horizon_quadruped} evaluated multiple inertial parameterizations using \gls{delan} for locomotion. The learned dynamics were used within \gls{mpc} and evaluated on a simulated quadruped, showing improved prediction accuracy and sample efficiency.
\cite{mbrl_baxter} learned a \gls{delan}-based transition model, with a Stribeck friction model and a kinematic manipulator model. The transition model was used as a simulator for \gls{rl}, where the policy optimizes computed-torque controller gains instead of torques directly, with validation on a real manipulator.

\subsubsection{Other Lagrangian Learning Methods:} \label{sec:other_lagrangian}
Some works used Lagrangian approaches with learning models other than \gls{delan} and \gls{lnn}. For example, \cite{Evangelisti2022} introduced a Lagrangian \gls{gpr} by embedding the Euler-Lagrange equations into matrix-valued kernels and Cholesky factors, preserving energy, passivity, and positive definiteness of the inertia matrix. Their method was used for trajectory tracking with a two-link manipulator model. \cite{Evangelisti2026} extended the same Lagrangian-\gls{gpr} idea to projector-based tracking control, combining uncertainty-adaptive feedforward-feedback control with exponential stability guarantees on a two-link system and a soft robotic manipulator. 
\cite{Libera2020,Giacomuzzo2022_MED} leveraged the fact that Lagrangian inverse dynamics can be expressed as a polynomial function in a suitable feature space, and designed a polynomial kernel in a \gls{gpr} model to learn the inertial, Coriolis, and gravity components of a robot manipulator. \cite{Giacomuzzo2024} extended further a Lagrangian-inspired polynomial kernel to learn the kinetic and potential energies of a manipulator, achieving data-efficient inverse-dynamics identification. 
\cite{delan_exoskeleton} trained \gls{delan} for lower-limb exoskeleton dynamics and used \gls{gpr} to compensate for the remaining residual errors while quantifying uncertainty. The resulting model was combined with a nonlinear extended state observer, and it was used within a backstepping controller for human-exoskeleton cooperative motion. 
\cite{Cheng2016_RKHS} devised structured \gls{rkhs} to learn the inverse dynamics and perform trajectory tracking with holonomic rigid-body robots. Specifically, they designed hybrid polynomial kernels to embed the Lagrangian structure of the robot dynamics, and proved that their method converges to the true dynamics with few samples.

\paragraph{Discussion:}
In suitable settings, Lagrangian learning methods offer a favorable trade-off between physical consistency, data efficiency, and generality. \Gls{delan} enforce positive definiteness of the inertia matrix by construction and expose interpretable quantities, including $\inertiaMat(\genPos)$, $\EPot(\genPos)$, and the system energy, which can be directly reused in computed-torque, energy-shaping, and \gls{mpc} controllers. However, their explicit energy decomposition \eqref{eq:lag_definition} is well suited to rigid-body systems but limits applicability to broader system classes. \Glspl{lnn} recover greater generality by learning the Lagrangian directly, at the cost of losing structure. Several works report higher accuracy and sample efficiency than black-box \glspl{mlp} models on the same systems \citep{lnn_mbrl_pmlr_ramesh23a}. However, validation remains concentrated on low-\gls{dof} benchmarks, with applications to floating-base and high-\gls{dof} robots only beginning to emerge \citep{felan,lnn_inf_horizon_quadruped}.

\subsection{Hamiltonian Learning Models} \label{sec:hamiltonian}


An alternative formalism to Lagrangian mechanics is the Hamiltonian approach \citep{Sakurai2020}.
The Hamiltonian describes a mechanical system in terms of its total energy, expressed as a function of the generalized coordinates $\genPos$ and momenta~$\genMom$.
The Hamiltonian is related to the Lagrangian through the Legendre transformation
\begin{equation}
    \Hamiltonian(\genPos, \genMom) = \genVel\Transp\frac{\partial \Lagrangian(\genPos, \genVel)}{\partial \genVel} - \Lagrangian(\genPos, \genVel).
    \label{eq:legendre}
\end{equation}
Using \eqref{eq:legendre}, the Euler-Lagrange equations \eqref{eq:eom} can be rewritten in terms of $\Hamiltonian$, resulting in the Hamiltonian equations
\begin{equation}
    \label{eq:hamiltonian_eq}
    \genVel = \frac{\partial \Hamiltonian(\genPos, \genMom)}{\partial \genMom}, \quad 
    \genMomDot = - \frac{\partial \Hamiltonian(\genPos, \genMom)}{\partial \genPos} + \genForce\,.
\end{equation}
where $\genVel$ and $\genMomDot$ are the time derivatives of the generalized coordinates and momenta, respectively, and $\genForce$ are the generalized forces.
%
%
%
\subsubsection{Hamiltonian Neural Networks (HNNs):} \label{sec:hnn}
Considering the autonomous conservative case, \ie{} $\genForce = 0$ in \eqref{eq:hamiltonian_eq}, \cite{hnn} introduced \glspl{hnn} by learning the Hamiltonian $\Hamiltonian(\genPos,\genMom)$ with an \gls{mlp} $\Hamiltonian_{\theta}(\genPos,\genMom)$, parameterized by $\theta$. Their training loss function aims to satisfy the Hamiltonian equations \eqref{eq:hamiltonian_eq}, which makes the learned dynamics energy-conserving by construction:
\begin{equation}
    \label{eq:hamiltonian_loss}
    \bigg|\bigg|\,\genVel - \frac{\partial \Hamiltonian_{\theta}(\genPos, \genMom)}{\partial \genMom}\,\bigg|\bigg|_2 + \bigg|\bigg|\,\genMomDot + \frac{\partial \Hamiltonian_{\theta}(\genPos, \genMom)}{\partial \genPos}\,\bigg|\bigg|_2\,.
\end{equation} 
In \eqref{eq:hamiltonian_loss}, $\frac{\partial \Hamiltonian_{\theta}}{\partial \genMom}$ and $\frac{\partial \Hamiltonian_{\theta}}{\partial \genPos}$ are computed via automatic differentiation of the \gls{mlp} $\Hamiltonian_{\theta}$, while the time derivatives $\genVel$ and $\genMomDot$ are available from the training data.
By minimizing the residuals of the Hamiltonian equations \eqref{eq:hamiltonian_eq}, the loss function \eqref{eq:hamiltonian_loss} is \textit{physics-informed}. Hence, the \gls{hnn} formulation of \cite{hnn} also belongs to the class of physics-informed learning methods (Sec.~\ref{sec:physics_informed}).
%
%

\paragraph{Applications of HNNs:}
\cite{hnn} applied \glspl{hnn} to learn the dynamics of pendulum and mass-spring systems, including settings where the state is inferred from pixel observations.
\cite{Zhong2020Symplectic} extended \glspl{hnn} to controlled Hamiltonian systems by introducing generalized forces through a learned input map. They also adopted structured Hamiltonian parameterizations based on kinetic and potential energy, similarly to \gls{delan}, and demonstrated model-based energy control on simulated pendulum and cart-pole systems.
\cite{Altawaitan2023HamiltonianDL} introduced an observation-space error function to train a \gls{hnn} directly from point cloud observations, employing an energy-based tracking controller for a real nonholonomic wheeled robot.
In soft robotics, where continuum mechanics and non-collocated actuation complicate minimal-coordinate modeling, \cite{lnn_hnn_soft_robots} jointly learned an input transformation matrix with \glspl{hnn} and \gls{delan}, while incorporating a dissipation term to capture non-conservative forces. The approach was applied to model-based tracking control.
In the setting of generative models (Table \ref{tab:generative_models_summary}), \cite{cui2026physical} developed Hamiltonian \glspl{vwm} with a structured and interpretable latent space, learning an energy function that satisfies Hamiltonian dynamics. Their generated trajectories were decoded into videos and used for long-horizon planning in manipulation.

Using a \gls{rkhs}, \cite{Smith2024} designed a new kernel function to enforce odd symmetry in the learned Hamiltonian vector field, and used it to learn the dynamics of a simulated two-link manipulator. Their method could generalize to unseen scenarios, and was shown to preserve total energy and odd symmetry of the learned dynamics. 

\cite{hnn_lnn_exp_const_se3} proposed learning \glspl{hnn} and \glspl{lnn} in $\SEThree$ body coordinates instead of minimal coordinates, enforcing the kinematic constraints through Lagrange multipliers. While $\SEThree$ representations simplify the dynamics by expressing the inertia matrix $\inertiaMat$ as a spatial rather than joint-space inertia, they introduce redundant states and require prior knowledge of the kinematic constraints \citep{felan}. Consequently, their application has so far been limited to low-dimensional systems such as pendulums, quadrotors, and wheeled robots.

\subsubsection{Port-Hamiltonian Learning Methods:}
Port-Hamiltonian approaches extend the Hamiltonian formulation to explicitly model energy storage, dissipation, and external energy exchange via energy ports, providing a unified framework for multi-domain physical systems, including mechanical, electrical, and thermal systems~\citep{port_hamiltonian_vanderSchaft2014}.
For instance, \cite{duong21_hnn_dyn_ctrl_se3} formulated port-Hamiltonian dynamics on the Lie group $\SEThree$ and combined them with passivity-based control for trajectory tracking, demonstrating the method on a simulated quadrotor. \cite{port_hnn_lie_control} extended this work by incorporating general Lie group constraints and an explicit dissipation term to capture friction and air drag, and validated the method on pendulum systems, ground vehicles, and quadrotors. 
%
%
\cite{Jin2025_port_hnn_underwater} used the port-Hamiltonian framework for \gls{sysid} of an underwater vehicle, by also learning a dissipative and input matrix.
\cite{Beckers2023} learned switching port-Hamiltonian dynamics with uncertainty quantification, embedding the energy-based structure of a hopping robot model and the switching behavior in a \gls{gpr} prior. A similar approach has been used to learn distributed port-Hamiltonian models of soft robotic manipulators in~\cite{ali2026learningbasedmodelingsoftrobots}.

\paragraph{Discussion:}
Hamiltonian methods share the interpretability and energy-consistency advantages of Lagrangian formulations. Moreover, the port-Hamiltonian extension models dissipation, actuation, and energy exchange explicitly rather than as residual terms, providing a principled treatment of non-conservative effects and a natural interface to passivity-based and energy-shaping control. However, because the Hamiltonian is defined in terms of the generalized momenta $\genMom$, which are rarely measured directly, recovering it from velocities requires either the inertia matrix or a learned transformation. Learning $\Hamiltonian$ with an unstructured \gls{mlp} also loses the explicit decomposition into inertia and potential energy, unless it is enforced by design \citep{Zhong2020Symplectic}, while energy consistency over long rollouts still requires structure-preserving integration. Finally, as with Lagrangian methods, $\SEThree$ formulations introduce redundant states, require known kinematic constraints, and have so far been validated primarily on low-\gls{dof} robotic platforms.

\subsection{Model-Structured Learning Architectures} \label{sec:msnn}

We use the term \textit{model-structured learning} to denote methods that embed physics priors into learning model architectures by: (i) designing specialized layers (\eg{} \gls{nn} layers) or internal architectural connections derived from physical principles, and/or (ii) enforcing physical consistency through architectural constraints. In the context of \glspl{nn}, these methods are referred to as \textit{\glspl{msnn}}, a term recently introduced in \cite{piccinini2025road,piccinini2025model}. We adopt the broader term \textit{model-structured learning} to encompass analogous approaches beyond \glspl{nn}, to unify the fragmented taxonomy used in the literature.

Table \ref{tab:msnn_summary} groups the papers discussed in this section by application category, summarizing the robot type, target application, specialized architectural elements, and physical-consistency constraints. 

\begin{table*}[t]
\scriptsize
\centering
\setlength{\tabcolsep}{2pt}
\renewcommand{\arraystretch}{1.12}
\begin{threeparttable}
\caption{Summary of the reviewed papers on model-structured learning approaches (Sec.~\ref{sec:msnn}), grouped by application category.}
\label{tab:msnn_summary}
\renewcommand{\tabularxcolumn}[1]{m{#1}}
\begin{tabularx}{\textwidth}{>{\centering\arraybackslash}m{2.0cm}|>{\centering\arraybackslash}m{3.1cm}|>{\centering\arraybackslash}m{2.2cm}|>{\centering\arraybackslash}X|>{\centering\arraybackslash}m{3.0cm}}
\toprule
\textbf{Paper} & \textbf{Application} & \textbf{Robot} & \textbf{Physical Principles in Archit. Layers / Connections / Design} & \textbf{Physical Constraints} \\
\hline
\multicolumn{5}{c}{\rule{0pt}{2.5ex}\textbf{\textit{Dynamics Learning}}}\\
\hline
\cite{DaLio2020modelling,james2020longitudinal} & Longitudinal dynamics learning & Vehicle & Force superposition; drag law; discrete gear-dependent dynamics & \ding{55} \\
\hline
\cite{piccinini2025road} & Longitudinal dynamics learning / friction estimation & Vehicle & Newtonian longitudinal dynamic laws; dependency of the dynamics on past input history & \ding{55} \\
\hline
\cite{chrosniak2024deep,Kolluri2025,fang2025fine} & Vehicle dynamics learning / estimation & Vehicle / race car & Single-track vehicle parameterization & Parameters constrained to remain physically meaningful \\
\hline
\cite{Kamp2023} & Suspension dynamics learning & Vehicle & Spring-damper dynamics & Regularization penalizing non-zero damper force at zero velocity \\
\hline
\cite{mungiello2026_roboracer} & Coupled longitudinal-lateral dynamics learning & Race car & Steady-state, transient, and coupling decomposition & \ding{55} \\
\hline
\cite{Yang2023} & Dynamics learning with physical human-robot interaction & Manipulator / collaborative robot & \gls{rk4} integration; inertial-stiffness-damping parameterization & Non-negativity constraints on selected parameters \\
\hline
\cite{delan_friction_iterative_2025} & Friction and dynamics learning & Manipulator & Friction decomposition: viscous, Coulomb, Stribeck & \ding{55} \\
\hline
\cite{Pfrommer2020} & Contact dynamics learning & Manipulator & Polytopic architecture with signed-distance and contact-Jacobian parameterization & \ding{55} \\
\hline
\cite{Beckers2026_Nonholonomic} & Nonholonomic dynamics learning & Wheeled system & Geometric structure-preserving for nonholonomic systems & \ding{55} \\
\hline
\cite{djeumou2022_pinn} & Dynamics learning & Multi-body robotic systems & Physics laws for multi-body systems & Constraints on symmetries and contact-force relations \\

\hline
\multicolumn{5}{c}{\rule{0pt}{2.5ex}\textbf{\textit{Trajectory Planning and Prediction}}}\\
\hline
\cite{piccinini2023physics,piccinini2025model,Piccinini2026_roboracer} & Trajectory planning / steering control & Race car & Double-track vehicle model; steady-state and transient dynamics; handling diagram & \ding{55} \\
\hline
\cite{antonucci2021efficient} & Human motion prediction for robot planning & Human motion in robotics & Social-force modeling & \ding{55} \\

\hline
\multicolumn{5}{c}{\rule{0pt}{2.5ex}\textbf{\textit{Control}}}\\
\hline
\cite{DaLio2020mental} & Steering control / lateral dynamics learning & Vehicle & Single-track steering relations & \ding{55} \\
\hline
\cite{Pagot2023Parking} & Steering control & Vehicle & Nonlinear steering characteristics; steady-state and transient steering behavior & \ding{55} \\
\hline
\cite{Bolderman2021} & Feedforward motor control & Linear motor & Linear friction structure; positive-mass prior & Rotor mass constrained positive \\
\hline
\cite{Sutanto2020} & Inverse dynamics learning & Robot arm manipulator & Differentiable Newton-Euler parametrization & Positive definite mass/inertia matrices; triangular inequality on principal inertias \\
\hline
\cite{Deng2024} & Inverse dynamics learning & Manipulator & Newton--Euler equation-embedded layers with composition operators and adaptive inter-layer connections & \ding{55} \\
\hline
\cite{Rezaei2019} & Inverse dynamics learning & Robot arm & Recursive Newton-Euler formulation to inform a cascaded \gls{gpr} & \ding{55} \\
\hline

\multicolumn{5}{c}{\rule{0pt}{2.5ex}\textbf{\textit{Estimation}}}\\
\hline
\cite{DaLio_SideSlip_2023} & Lateral velocity estimation & Vehicle & Kinematic equations and observers & \ding{55} \\
\hline
\cite{Candeo2026} & Brake pad emission estimation & Vehicle braking system & Physical dependence on temperature, deceleration, and velocity & \ding{55} \\
\hline
\cite{delan_non_sym_friction} & Friction and inertia identification & Manipulator & 
Linearity of viscous forces and inertia in the joint velocities and accelerations; direction-dependent Coulomb friction & \ding{55} \\
\hline
\cite{Dai2024} & Aerodynamic friction estimation & Aerial robot & \gls{gpr} structure enforcing passivity & Positive semi-definiteness of the damping matrix \\
\bottomrule
\end{tabularx}
\end{threeparttable}
\end{table*}

\paragraph{MSNNs with Physical Layers and Internal Connections:}
\cite{DaLio2020modelling} and \cite{james2020longitudinal} developed \glspl{msnn} to learn the longitudinal dynamics of vehicle. Their architectures embed physical principles including force superposition, the quadratic dependence of aerodynamic drag on vehicle speed, and the effect of past control inputs on the current longitudinal acceleration. Dedicated layers model the contributions of aerodynamic, traction, and braking forces, while locally-activated neural models capture gear-dependent dynamics. \cite{piccinini2025road} extended these architectures to varying road surfaces by introducing neuro-fuzzy \gls{fir} layers that model the dependence of longitudinal acceleration on the history of brake and throttle inputs, augmenting the Newtonian dynamics equations.
\cite{DaLio2020mental} proposed \glspl{msnn} to learn and control the lateral vehicle dynamics. Their architecture embeds steering characteristics derived from single-track vehicle models \citep{Abe2009}. After learning the forward dynamics, they trained an inverse-model steering controller end-to-end, by backpropagating control losses through the learned dynamics, without additional training data. Subsequent work \citep{piccinini2023physics,piccinini2025model,Piccinini2026_roboracer} extended this approach to driving at the handling limits by embedding local learnable approximations of double-track vehicle models.
\cite{Pagot2023Parking} introduced a \gls{msnn} for steering control in automated parking, with a neural polynomial layer to learn the nonlinear steering characteristics, and a \gls{fir} layer to learn the transient dynamics. \cite{DaLio_SideSlip_2023} proposed an \gls{msnn} for lateral velocity estimation. They extended the equations of a kinematic observer with interpretable neural terms, to estimate the lateral velocity from the measured accelerations, yaw rate, and longitudinal speed.
\cite{mungiello2026_roboracer} introduced a unified autoregressive \gls{msnn} to model the coupled longitudinal-lateral vehicle dynamics near the limits. Their architecture uses specialized layers to capture steady-state and transient effects, as well as their coupling.

Beyond vehicle dynamics, \cite{antonucci2021efficient} designed an \gls{msnn} for real-time human motion prediction in robotics, by neuralizing a social force model and estimating the interaction forces from past pose sequences. \cite{delan_non_sym_friction} introduced a \gls{msnn} to complement a \gls{delan} and learn friction and inertial effects of a robotic manipulator. They designed linear layers to learn the viscous forces and motor inertia (reflecting the linearity of these effects in the joint velocities and accelerations), and two tailored layers to learn the non-symmetrical Coulomb friction forces considering the joint rotation direction. Similarly, \cite{delan_friction_iterative_2025} introduced a structured layer to learn the joint friction components (viscous, Coulomb, and nonlinear Stribeck effects) within a \gls{delan} architecture for manipulator dynamics learning.

\cite{Candeo2026} developed a recurrent \gls{msnn} for brake pad emission prediction, embedding physical dependencies on brake temperature, vehicle deceleration, and velocity. 
\cite{Pfrommer2020} devised a polytopic \gls{nn} architecture to learn discontinuous contact dynamics for manipulation systems with frictional impacts and stiction. They parameterized inter-body signed distance functions and contact-frame Jacobians, thus embedding rigid contact mechanics and geometric priors into the model architecture.
\cite{Deng2024} proposed an Equation-Embedded \gls{nn} for inverse dynamics learning in robotic manipulators. The Newton--Euler equations of motion are encoded directly in the \gls{nn} architecture as specialized layers with composition-operator activations, while a liquid mechanism dynamically adapts inter-layer connections to the input data. Their model achieved competitive torque prediction and parameter identification with limited data, outperforming black-box baselines under varying friction conditions.

\paragraph{MSNNs with Physical Constraints:}
In \cite{chrosniak2024deep,Kolluri2025,fang2025fine}, \glspl{nn} learned the parameters of a single-track vehicle model, while sigmoid activations embedded in the \gls{nn} architecture constrained the parameters to physically meaningful ranges. 
Similarly, \cite{Yang2023} used \glspl{rnn} with \gls{rk4} integration layers to learn the inertial, stiffness and damping parameters of a manipulator's joints, while enforcing non-negativity constraints on certain parameters through custom \gls{rnn} cells.
\cite{Kamp2023} introduced neural layers to model mechanical springs and dampers, augmenting a vehicle's suspension dynamics model with learnable components. They also enforced physical consistency through loss function regularization, \eg{} by penalizing non-zero damper forces at zero velocity.
In the field of linear motor control, \cite{Bolderman2021} designed a structured neural layer to encode the linear effect of friction components, the positivity of the rotor mass, and physics-based weights initialization. \cite{Sutanto2020} learned the inverse dynamics of a robot arm with tailored model parametrizations to enforce physical consistency of the learned parameters. Specifically, their training process constrains the positive definiteness of mass and inertia matrices, and the triangular inequality of the principal moments of inertia. 
For multi-body robotic systems, \cite{djeumou2022_pinn} modeled the dynamics as the sum of known physics-based and learnable neural terms, while enforcing symmetries and contact force relations through architectural constraints on the learnable parts.

\paragraph{Model-Structured Learning with Gaussian Processes:}
Model-structured learning models have also been developed with \gls{gpr} frameworks.
In particular, \cite{Beckers2026_Nonholonomic} incorporated the geometric structure of nonholonomic dynamics into a matrix-valued kernel of a \gls{gpr} model, enabling dynamics learning of nonholonomic systems while rigorously certifying constraint satisfaction. Their method was validated on a nonholonomic rolling disc, with applications to wheeled robots. 
\cite{Dai2024} modeled nonlinear friction with a dissipative \gls{gpr}, whose matrix-vector structure enforces passivity and positive semi-definiteness of the damping matrix. Their model was used for aerodynamic friction estimation of an aerial robot. 
\cite{Rezaei2019} controlled a robot arm using a cascaded \gls{gpr} model arranged along the kinematic chain, inspired by recursive Newton-Euler algorithms, and predicting joint torques recursively from the end-effector to the base or vice versa.

\paragraph{Discussion:}
In many of the previously cited works, \glspl{msnn} were compared against general-purpose \glspl{nn} and learning models (\eg{} \glspl{mlp} and \glspl{rnn}), demonstrating improved sample efficiency, generalization from limited training data, and better interpretability due to the explicit embedding of physical principles \citep{DaLio2020modelling,djeumou2022_pinn,piccinini2023physics,DaLio_SideSlip_2023,piccinini2025road,mungiello2026_roboracer}. However, the design of model-structured architectures often requires significant engineering effort and domain expertise, and their performance is strongly influenced by the choice of the embedded physical principles and the structure of the architectural layers, connection, and constraints. In Sec.~\ref{sec:software_model_structured}, we review open-source libraries that facilitate the design of model-structured learning architectures.

\subsection{Neural ODEs and Variational Integrator Networks} \label{sec:node_vi}

Physical systems, including robots, operate in a continuous-time domain. However, computational algorithms are inherently discrete. To bridge this gap, \cite{Haber2018,Weinan_resnet_as_dyn_sys_2017} reinterpreted ResNets~\citep{He2016-yr} as the Euler discretization of the continuous-time transformation:
\begin{equation}
\label{eq:ode}
\underbrace{\frac{d\mathbf{x}(t)}{dt} = \mathbf{f}_{\boldsymbol{\theta}}\big(\mathbf{x}(t),t\big)}_{\text{NODE}} \rightarrow \mathbf{x}(t+\Delta t) = \mathbf{x}(t) + \underbrace{\mathbf{f}_{\boldsymbol{\theta}}(\mathbf{x}(t),t) \; \Delta t}_{\text{ResNet block}},
\end{equation}
where $\mathbf{x}(t)$ is the state at time $t$, $\Delta t$ is the time step, and $\mathbf{f}_{\boldsymbol{\theta}}$ is a \gls{nn} parameterized by $\boldsymbol{\theta}$. \cite{chen2018neural} formalized the continuous-time \gls{ode} in \eqref{eq:ode} as a Neural ODE (\gls{node}), embedding standard \gls{ode} solvers directly into the \gls{nn}'s computation graph. Using the adjoint sensitivity method for automatic differentiation, \glspl{node} can be trained with backpropagation through any \gls{ode} solver, enabling end-to-end learning of continuous-time dynamics. 

\paragraph{Neural ODEs in Robotics:}
\glsplural{node} have been applied to learning system dynamics and continuous motion plans. \cite{rhode2024} combined analytical kinematics with a \gls{node} that models complex velocity dynamics, improving the prediction of a vehicle's motion. \cite{knode_mpc} embedded a physics-prior residual \gls{node} into \gls{mpc} for quadrotor tracking control. To handle dynamic environments, \cite{context_aware_node_mpc} conditioned a \gls{node}-\gls{mpc} scheme on a latent context representation. In manipulation, \glspl{node} were used to learn skills from demonstrations, leveraging hypernetworks for continual adaptation~\citep{continual_robot_skills}, or integrating Control Lyapunov and Barrier Functions to guarantee stability and safety in multi-attractor periodic tasks~\citep{learn_complex_node_2024}. \glspl{node} also provide the theoretical foundation for flow matching~\citep{lipman2023flow}, a state-of-the-art approach for imitation learning.

\paragraph{Variational Integrator Networks:}
While \glspl{node} successfully integrate deep learning into continuous-time modeling, they rely on general-purpose numerical integrators (\eg{} Runge-Kutta methods) that do not preserve the geometric and energy properties of physical systems. In contrast, \glspl{vi} \citep{MARSDEN2001253} leverage discrete variational principles to construct symplectic integrators, naturally preserving the symplectic structure and, under appropriate discrete symmetries, momentum and favorable long-term energy conservation. 
By leveraging \glspl{vi}, \cite{Saemundsson2020} proposed Variational Integrator Networks (\glspl{vin}) that explicitly match discrete-time equations of motion. Their approach encodes symplecticity, energy and momentum conservation laws as hard architectural constraints, promoting accurate long-term prediction, interpretability, and data-efficient learning. \glspl{vin} were validated on mechanical systems like pendulums and mass-spring systems, learning from both noisy observations and directly from image pixels.

To extend \glspl{vin} to non-conservative and actuated systems, \cite{pmlr-v144-havens21a} introduced \glspl{fvin}. By leveraging discrete d'Alembert's principles, \glspl{fvin} capture external control inputs and damping while preserving underlying passive energy dynamics. The approach was applied on a real cartpole system using \gls{mpc}.
Similarly, \cite{duruisseaux2023lie} proposed Lie-group \glspl{fvin} to learn discrete-time flow maps, which were validated in simulation to control a pendulum and a quadrotor.
Based on the same formulation, \cite{PhysORD_2024} learned a symplectic map for a real off-road vehicle by parameterizing the potential energy and non-conservative forces via a \gls{nn}, and demonstrated improved motion prediction performance.

Collectively, \glspl{vin} and \glspl{fvin} provide a promising approach to learn continuous-time dynamics while preserving the underlying physical structure and properties of the system. However, their application to complex robotic systems and contact-rich dynamics remains an open challenge.

\subsection{Hybrid Physics-Learning Architectures} \label{sec:hybrid_physics_nn}

This body of research combines physics-based models with \gls{ml} in a modular way, where learning-based methods learn specific components of the dynamics or correct the output of physics-based models. Unlike \glspl{msnn}, these techniques use learning-based methods as separate modules that interact with physics-based models. This modularity allows for a flexible design of the \gls{ml} components, while still leveraging unchanged physics-based models to capture parts of the system dynamics. 

\subsubsection{Learning Complex Subsystems:}
Some authors used \gls{ml} to learn complex subsystems of physics-based models when accurate parameterization is costly, time-consuming, or requires extensive domain expertise. In such cases, the subsystem dynamics can be learned from data, while the remaining system components are described by physics-based equations.
For example, in the field of vehicle dynamics, the parameterization of classical tire models (\eg{} Pacejka Magic Formula \citep{pacejka2012tire}) is expensive and time-consuming, as it requires extensive experimental data with costly test machines. To address these issues, \gls{nn} tire models were developed by \cite{Djeumou2025,wkegrzynowski2024learning,zhou2024vehicle,Kim2022,Viehweger2021,Acosta2019,Acosta2018,Acosta2018drift}, to learn the tire forces and moments as a function of a vehicle model's states and controls. These \gls{nn} tire models were integrated into physics-based vehicle models \citep{Guiggiani2018}, and used for velocity estimation \citep{wkegrzynowski2024learning}, vehicle dynamics simulation \citep{zhou2024vehicle} and estimation \citep{Viehweger2021,Acosta2018,Acosta2019}, \gls{mpc}-based vehicle control \citep{Kim2022,Acosta2018drift}, and autonomous drifting \citep{Djeumou2025}. \cite{sun2024trajectory} learned the tire cornering stiffness and other physical parameters of a vehicle model, and used them for \gls{mpc} trajectory tracking.
\cite{Zhou2025} adopted a \gls{nn} to learn a vehicle's traction forces, and fed them to a physics-based estimator, for mass and acceleration estimation. 
In these works, the \gls{nn} architectures were general-purpose \glspl{mlp}, \glspl{anfis} \citep{Acosta2019}, and neural-ExpTanh networks \citep{djeumou2023}.

In \cite{yang2023physics}, an \gls{rnn} with dedicated nonlinear terms was used to learn the dynamics of a manipulator's gearbox (harmonic drive), offering an interpretation of the trained weights in terms of physical parameters. \cite{djeumou2023how} proposed physics-constrained neural \glspl{sde} for long-horizon prediction and control of an hexacopter using only three minutes of training data. The drift term of their \gls{sde} combines physics-based rigid-body dynamics with learned neural components for the complex thrust and aerodynamic effects, while the diffusion term provides distance-aware uncertainty estimates outside the training distribution.

Taken together, these works demonstrate that \glspl{nn} can effectively learn complex subsystems of physics-based models, enabling accurate modeling, control and estimation of robotic systems with limited data and domain knowledge.

\subsubsection{Residual Learning:}
Several works learn residual dynamics on top of physics-based models using different \gls{ml} techniques. Residual learning captures discrepancies caused by unmodeled dynamics, parameter uncertainty, or simplifying modeling assumptions. \cite{dikici2025learning} trained an \gls{mlp} to correct the output of a physics-based single-track vehicle model for autonomous racing, using limited track data. For a similar purpose, \cite{miao2025residual}, \cite{Cai2024_hybrid} and \cite{Xue2024} used transformers, \glspl{gru} and local affine models, respectively. \cite{Kabzan2019,hewing2018cautious,Costa2023} employed \gls{gpr} and \gls{rbfn} correctors for single-track vehicle models, and integrated them into \gls{mpc} controllers for autonomous racing. \cite{DiDino2011} used a small \gls{mlp} to learn the residual dynamics of a vehicle's shock absorber. In the context of dynamics learning for robotic manipulators, \cite{delan_friction_iterative_2025} used a \gls{tcn} to learn the residual torque not captured by a \gls{delan} model, taking as input a past window of joint positions, velocities and accelerations. \cite{Fan2023} used an \gls{mlp} to learn the residuals of a physics-based feedforward controller for a robotic stepper motor, training first the physics-based model, and then the \gls{nn} on top of it. \cite{Nguyen2010} proposed a semi-parametric approach to learn the discrepancies of a physics-based model for a robotic manipulator, using \gls{gpr} with priors in the mean and kernel functions derived from physics-based knowledge.
\cite{beckers2017stable} leveraged \gls{gpr} to learn the residual dynamics of a two-link manipulator for feedforward control compensation. This work was extended in \cite{Beckers2019} by incorporating model-confidence information to safely adapt the feedback gains during trajectory-tracking tasks.
\cite{Camoriano2016} learned the residuals of a rigid-body dynamics model with an \gls{rkhs} model, updating both parts online with a recursive least-squares algorithm, to learn the inverse dynamics of a humanoid arm.
\cite{Lai2021} used a \gls{node} to learn the residual dynamics of a physics-based model for mass-spring-damper systems. Their additive \gls{mlp} corrector was trained with the \gls{sindy} method \citep{Brunton2016}. 
Following a similar principle, \cite{donati_2025a} combined a physics-based model with a sparse residual for vehicle dynamics learning over multi-step prediction horizons. Their residual is expressed as an $\ell_1$-regularized linear combination of candidate basis functions, allowing it to capture only unmodeled dynamics while preserving the interpretability of the  physical parameters. They also derived parameter error bounds and conditions for exact recovery of the sparse residual, and applied the method to vehicle lateral dynamics learning.
With an \gls{mlp}, \cite{Bolderman2024,Bolderman2022_feedforward} learned the residuals of a physics-based model for feedforward control of a robotic linear motor. They used regularization terms in the loss function to reduce the deviation of the learned parameters from their physical values. In the context of long-horizon manipulation planning, \cite{Zhong2026} employed an analytical physics-based model to predict the motion of pushed objects, and fed the model's predictions to an \gls{mlp} that learns the residuals. Residual learning can also be applied to the control parameters, rather than to the predicted dynamics. In TossingBot \citep{zeng2019tossingbot}, a manipulator learns to grasp objects from a bin and throw them beyond its kinematic reach. An analytical ballistic model predicts the release velocity, while a learned residual compensates for unmodeled effects such as mass distribution, friction, aerodynamics, and grasp offset. The jointly trained grasping and throwing policies achieved higher throwing accuracy than both purely analytical and purely data-driven baselines.

Overall, residual learning can complement physics-based models by improving accuracy in the presence of unmodeled dynamics, while avoiding the design of specialized architectures or explicit physics constraints. However, its performance ultimately depends on the fidelity of the underlying physics model and the expressive power of the \gls{ml} technique.

\subsubsection{Learning-Based Sensor Pre-Processing for Physics-Based Models:}
This category uses learning-based pre-processing of raw sensor measurements, whose outputs are then fed to physics-based models. In the context of velocity estimation, \cite{Bertipaglia2024,Kim2021_integrated,Kim2020,Novi2019_integrated,Boada2016} used \glspl{kf} with physics-based vehicle models to process the outputs of \glspl{nn} reading on-board sensor measurements. They adopted different \glspl{nn} architectures, from \glspl{cnn} \citep{Bertipaglia2024} to \glspl{mlp} \citep{Novi2019_integrated}, \glspl{lstm} \citep{Kim2021_integrated,Kim2020} and \glspl{anfis} \citep{Boada2016}. Similarly, \cite{Kim2024_hybrid} used a transformer to pre-process onboard sensor measurements, and fed them to a physics-based estimator for yaw rate and sideslip angle estimation. For vehicle steering control, \cite{Zia2025} used neural layers to learn an adaptive look-ahead distance and target heading, which were fed to a physics-based pure-pursuit controller, together with a \gls{nn} residual to compensate for unmodeled dynamics.

In these works, \gls{ml} models yield virtual measurements or reference signals used by physics-based estimators or controllers, improving the overall performance.

\subsection{Physics-Encoded Topology Learning} \label{sec:topology}
Another line of research focuses on learning \gls{ml} model topologies by assembling basic operators and connections while embedding physics priors.
Table \ref{tab:topology_summary} summarizes the reviewed papers by application, robot type, and underlying topology-learning or physics-encoding principle.

\begin{table*}[t]
\scriptsize
\centering
\setlength{\tabcolsep}{2pt}
\renewcommand{\arraystretch}{1.12}
\begin{threeparttable}
\caption{Summary of the reviewed papers on topology learning with physics-encoded architectures (Sec.~\ref{sec:topology}).}
\label{tab:topology_summary}
\renewcommand{\tabularxcolumn}[1]{m{#1}}
\begin{tabularx}{\textwidth}{>{\centering\arraybackslash}m{2.6cm}|>{\centering\arraybackslash}m{3.0cm}|>{\centering\arraybackslash}m{2.3cm}|>{\centering\arraybackslash}X}
\toprule
\textbf{Paper} & \textbf{Application} & \textbf{Robot} & \textbf{Topology-Learning / Physics-Encoding Principles} \\
\hline
\multicolumn{4}{c}{\rule{0pt}{2.5ex}\textbf{\textit{Dynamics Learning}}}\\
\hline
\cite{Papageorgiou2024} & Dynamics learning & Soft robot & \gls{sindy} library from kinematic/dynamic principles, including trigonometric terms for rotational effects \\
\hline
\cite{Liu2024} & Dynamics learning & \gls{umv} & Physics-inspired \gls{sindy} candidate functions derived from Newtonian and hydrodynamic terms \\
\hline
\cite{Buriani2026} & Dynamics learning & Jumping quadruped & Reduced-order \gls{sindy} in a physically interpretable latent space \\
\hline
\cite{Omar2024} & Dynamics learning with friction and payload effects & Manipulator & Euler-Lagrange-informed \gls{sindy} library with inequality-constrained sparse identification \\
\hline
\cite{Bhattacharya2020_sparse} & Dynamics equation discovery & Soft robotic esophagus & Reduced-order modeling combined with \gls{sindy} for sparse differential equation discovery \\
\hline
\cite{Wei2025} & Dynamics learning (including collisions) & Rigid multi-body systems & Neural-mesh topology with connectivity induced by Newtonian interaction priors \\

\hline
\multicolumn{4}{c}{\rule{0pt}{2.5ex}\textbf{\textit{Trajectory Planning}}}\\
\hline
\cite{Villeda2023} & Trajectory imitation / planning-oriented policy generation & Manipulator & Equation Learner Network topology from candidate operators with task-physics-aware parametrization \\
\hline
\cite{Chen2021} & Trajectory planning with obstacle avoidance & Manipulator & Sparse-regression model with candidate library seeded by simpler pendulum dynamics priors \\
\hline

\multicolumn{4}{c}{\rule{0pt}{2.5ex}\textbf{\textit{Control}}}\\
\hline
\cite{Ledezma2017,Ledezma2018} & Inverse dynamics learning & Manipulator; humanoid & First-Order Principle operators with Newton-Euler priors; automatic topology construction from physics-motivated building blocks \\
\hline
\cite{Cazenille2012} & Neural controller design and control performance analysis & Modular robotic control benchmark (inverted pendulum) & Topology search over micro/macro \gls{nn} structures (regularity, modularity, neuron interaction models) \\
\hline
\cite{Weihmayr2025} & Path-tracking control (\gls{mpc}) & Vehicle & \gls{sindy}-identified vehicle dynamics integrated into model-based control \\
\hline
\cite{Paparazzo2025} & Trajectory-tracking control (\gls{mpc}) & Vehicle & Online \gls{sindy} for lateral tire-force modeling with symmetry-aware odd polynomial libraries \\
\hline
\cite{Bhattacharya2022} & Predictive control (\gls{mpc}) & Soft robotic esophagus & \gls{sindy} with control for interpretable predictive modeling from pressure and time-of-flight sensing \\
\hline

\multicolumn{4}{c}{\rule{0pt}{2.5ex}\textbf{\textit{Estimation}}}\\
\hline
\cite{Shen2022} & Online dynamics estimation / identification of time-varying behavior & SCARA robot & Sparse Bayesian regression with a physics-informed dictionary and recursive updates \\
\bottomrule
\end{tabularx}
\end{threeparttable}
\end{table*}

\subsubsection{NN-based Topology Learning:}
\cite{Ledezma2017} introduced First-Order Principle (FOP) \glspl{nn}, which learn to combine a set of basic operators designed with prior knowledge of a system's physics. Their FOP-\glspl{nn} were applied to inverse dynamic problems with manipulators, where the kinematic structure and inertial parameters were learned using the Newton-Euler laws as FOP operators. Their work was extended by \cite{Ledezma2018}, who learned the robot geometrical and inertial properties with two decoupled problems, and constructed the \gls{nn} topology automatically from FOPs. Their method was used to learn the inverse dynamics of a simulated humanoid and a real-world manipulator. 
\cite{Cazenille2012} analyzed how the micro- and macro-structures of \glspl{nn} affect neural controllers for modular robots, comparing summing versus multiplicative neuron models and different \gls{nn} regularity and modularity properties. They validated their approach on the inverted pendulum benchmarks for distributed modular robotic control. Their study shows how topology and modularity choices improve control performance and generalization.
\cite{Wei2025} proposed topological \glspl{nn} to learn rigid body dynamics, including multi-body interactions and collisions. They built neural meshes on body surfaces with physics-encoded connectivity among mesh nodes, using inductive biases derived from Newton laws.   
\cite{Villeda2023} used Equation Learner Networks (\glspl{eqln}) \citep{Sahoo2018} to imitate expert trajectories for a robotic manipulator, by learning to combine a set of candidate activation functions in an \gls{mlp} to best fit the trajectory data. They encoded physical properties of the robotic task with dedicated parameters, and showed improved generalization to unseen scenarios.

Overall, \gls{nn}-based topology learning with physics priors enables the discovery of interpretable and generalizable models, while reducing the need for extensive domain knowledge and manual architecture design.

\subsubsection{Sparse Identification of Nonlinear Dynamical Systems (SINDy):} \label{sec:sindy}
\cite{Brunton2016} introduced \gls{sindy} to discover governing equations from data by selecting sparse terms from a library of candidate nonlinear functions. Starting from this seminal work, some authors proposed methodological extensions of \gls{sindy} to embed physics priors into the learning process. In particular, \cite{Chu2020} proposed Lagrangian-\gls{sindy} to discover compact expressions of the total energy and Lagrangian of multi-\gls{dof} systems, using prior knowledge from simpler subsystems. \cite{Lathourakis2024} proposed physics-encoded \gls{sindy}, which injects known friction laws and domain biases to capture discontinuous stick-slip dynamics. \cite{Chen2021_physics} combined \glspl{pinn} with sparse regression to learn governing equations from scarce and noisy data by embedding residual equations in the loss. \cite{Zolman2025} combined \gls{sindy} with \gls{rl} to learn sparse, interpretable models of dynamics, reward, and policy. \cite{Purnomo2023} proposed xL-\gls{sindy} to identify sparse Lagrangian expressions via the Euler-Lagrange equations and a proximal gradient method, with the option to leverage prior Lagrangian knowledge from simpler subsystems.

Other works \textit{applied} \gls{sindy} variants to learn or control the dynamics of robotic systems, by embedding physics insights into the design of the candidate library and the learning process. \cite{Weihmayr2025} used \gls{sindy} to learn the dynamics of a vehicle, and integrated the model into an \gls{mpc} controller for path tracking, showing improved results in comparison with a physics-based single-track model. \cite{Paparazzo2025} employed \gls{sindy} for online learning of a vehicle's lateral tire forces, using odd polynomial functions to reflect the symmetry of the lateral forces with respect to the slip angle. They constrained the learnable parameters to be physically meaningful, and integrated the learned model into an \gls{mpc} controller for trajectory tracking. In \cite{Papageorgiou2024}, \gls{sindy} was adopted to learn the dynamics of a soft robot, using candidate functions derived from kinematic and dynamic principles, including trigonometric terms to capture rotational dynamics.
\cite{Liu2024} applied \gls{sindy} to learn the dynamics of an \gls{umv}, using a library of physics-inspired candidate functions, derived from Newton's laws and hydrodynamics. In \cite{Buriani2026}, the dynamics of a jumping quadruped was learned using \gls{sindy} in a reduced-order, physically-interpretable latent space.
\cite{Omar2024} used \gls{sindy} to learn the dynamics of a robot manipulator, including friction and payload effects. Their candidate functions were designed from Euler-Lagrange principles, integrating prior knowledge of the robot kinematics and dynamics. They incorporated inequality constraints into \gls{sindy} to handle the ill-conditioning that arises from simple excitation trajectories.
\cite{Chen2021} proposed a sparse-regression framework for manipulator dynamics learning, where prior knowledge from simpler pendulum systems was used to construct the candidate library for a 3-\gls{dof} manipulator. They validated the identified model in simulated obstacle-avoidance trajectory planning for a manipulator. \cite{Bhattacharya2022} used \gls{sindy} with control to discover an interpretable predictive model of a robotic soft esophagus directly from pressure and time-of-flight sensor data, without relying on a first-principles soft-body model. Their model was integrated in an \gls{mpc} to control a robotic esophagus. \cite{Bhattacharya2020_sparse} combined reduced-order modeling with \gls{sindy} to discover the differential equations of an esophageal swallowing robot from pressure-driven deformation data. Their method was validated on a soft robotic esophagus, using measurements from a quarter-scale platform and validation on the full system. \cite{Shen2022} cast industrial robot dynamics identification as a sparse Bayesian regression problem, using a dictionary of candidate robot dynamics terms and recursive online updates to track time-varying behavior. They validated the method on simulated and real SCARA robots.

\paragraph{Discussion:}
Collectively, \gls{sindy}-based methods enable the discovery of compact, interpretable robot dynamics by embedding physics priors into the candidate function library and sparse regression process. However, their effectiveness heavily depends on the candidate library and on the coordinates in which the dynamics are expressed: designing both still requires substantial human expertise.
Most formulations also regress on state derivatives estimated from noisy measurements, and become ill-conditioned when the excitation trajectories are not sufficiently informative.
Finally, phenomena such as contact, hysteresis, and backlash are difficult to capture with compact libraries of candidate functions, so scaling \gls{sindy} to high-dimensional, interacting, and contact-rich robotic systems remains an open research challenge.

\subsection{Physics-Encoded Neural Operators} \label{sec:neural_operators_encoded}

Neural operators (\glspl{no}) extend \glspl{nn} from learning fixed-dimensional input-output mappings to learning operators between infinite-dimensional function spaces \citep{Li_2024,Faroughi2024}. Physics priors can be incorporated into these architectures by embedding governing equations or physical principles into the operator structure (physics-encoded \glspl{no}). 

Table \ref{tab:neural_operator_summary} summarizes the papers discussed in this section, and in the sections on physics-informed and physics-guided \glspl{no}, indicating the type of operator, the way physics is embedded, and the robot class.

\begin{table*}[t]
\scriptsize
\centering
\setlength{\tabcolsep}{2pt}
\renewcommand{\arraystretch}{1.1}
\begin{threeparttable}
\caption{Summary of the reviewed neural-operator papers for robotics and related robotic systems. Checkmarks indicate whether a paper embeds physics through operator architecture design (Physics-Encoded, P.E., Sec.~\ref{sec:neural_operators_encoded}), physics-based training loss functions (Physics-Informed, P.I., Sec.~\ref{sec:neural_operators_loss}), and/or physically structured inputs, states or data (Physics-Guided, P.G., Sec.~\ref{sec:neural_operators_guided}).}
\label{tab:neural_operator_summary}
\renewcommand{\tabularxcolumn}[1]{m{#1}}
\begin{tabularx}{\textwidth}{>{\centering\arraybackslash}m{3.3cm}|>{\centering\arraybackslash}m{1.6cm}|>{\centering\arraybackslash}m{3.2cm}|>{\centering\arraybackslash}m{0.6cm}|>{\centering\arraybackslash}m{0.6cm}|>{\centering\arraybackslash}m{0.6cm}|>{\centering\arraybackslash}X}
\toprule
\textbf{Paper} & \textbf{Neural Operator} & \textbf{Robot} & \textbf{P.E.} & \textbf{P.I.} & \textbf{P.G.} & \textbf{How Physics is Incorporated} \\
\hline
\multicolumn{7}{c}{\rule{0pt}{2.4ex}\textbf{\textit{Dynamics Learning}}} \\
\hline
\cite{Zhang2026_physics} & \multirow{4}{*}{Koopman} & Vehicle &  & \checkmark &  & Acceleration-consistency physics loss \\
\cline{1-1}\cline{3-7}
\cite{Ristich2025} &  & Soft robotic arm &  & \checkmark &  & PDE-residual split-operator training \\
\cline{1-1}\cline{3-7}
\cite{Shi2022} &  & Wheeled robot / 2-link arm / soft robotic leg & \checkmark &  &  & Topology-derived analytical lifting dictionary \\
\hline
\cite{Wang2025_friction} & \gls{kan} & Manipulator & \checkmark &  &  & Interpretable friction-law parameterization \\
\hline
\cite{Tan2025_vehicle} & \gls{deeponet} & Vehicle &  & \checkmark & \checkmark & Model-based latent state mapping + physics loss \\
\hline
\multicolumn{7}{c}{\rule{0pt}{2.4ex}\textbf{\textit{Trajectory Planning}}} \\
\hline
\cite{Chen2024} & \gls{gno} & Vehicles on traffic graph &  & \checkmark &  & Fokker-Planck equation residuals in loss function \\
\hline
\multicolumn{7}{c}{\rule{0pt}{2.4ex}\textbf{\textit{Control}}} \\
\hline
\cite{Joglekar2023} & \multirow{13}{*}{Koopman} & Vehicle & \checkmark &  &  & Lie-bracket-informed lifting functions design \\
\cline{1-1}\cline{3-7}
\cite{Abraham2017} &  & Cart-pendulum / spherical mobile robot & \checkmark &  &  & System-structure-based lifting functions design \\
\cline{1-1}\cline{3-7}
\cite{Shi2022_grippers} &  & Soft multi-finger gripper & \checkmark &  &  & Geometry/actuator-informed lifting functions design \\
\cline{1-1}\cline{3-7}
\cite{Castano2020} &  & Robotic fish / soft swimmer & \checkmark &  &  & Kinematics-informed lifting functions design \\
\cline{1-1}\cline{3-7}
\cite{Wang2024} &  & Cart-pole & \checkmark &  &  & Energy-preserving Koopman constraints \\
\cline{1-1}\cline{3-7}
\cite{Zhang2025} &  & Nonholonomic mobile robot & \checkmark &  &  & Kinematic-coordinate observables with yaw embeddings \\
\cline{1-1}\cline{3-7}
\cite{Kamath2026} &  & Quadrotor & & \checkmark &  & Quadrotor dynamics residuals in physics loss \\
\cline{1-1}\cline{3-7}
\cite{Mamakoukas2021} &  & Inverted pendulum / robotic fish &  &  & \checkmark & Derivative-augmented observables \\
\cline{1-1}\cline{3-7}
\cite{Rosenfelder2025} &  & Differential-drive robot &  &  & \checkmark & Rolling-constraint geometric state parameterization \\
\cline{1-1}\cline{3-7}
\cite{Zinage2022}; \cite{Sriram2023} &  & Quadrotor &  &  & \checkmark & $SE(3)$-structured Koopman coordinates \\
\cline{1-1}\cline{3-7}
\cite{Zhang2025_Koopman} &  & Flexible robotic needle &  &  & \checkmark & Bicycle/needle-kinematics state lifting \\
\hline
\multicolumn{7}{c}{\rule{0pt}{2.4ex}\textbf{\textit{Estimation}}} \\
\hline
\cite{Sivakumar2026} & \gls{fno} & Quadrotor &  & \checkmark &  & Latent-conditioned FNO with physics penalties \\
\hline
\cite{lim2026_vega} & \gls{pino} & Vehicle &  & \checkmark &  & Hybrid residual loss for parameter inference \\
\bottomrule
\end{tabularx}
\end{threeparttable}
\end{table*}

Most works in this category focused on Koopman operators for learning the dynamics of robotic systems.
Koopman operator theory represents nonlinear dynamics via a linear, typically infinite-dimensional operator acting on observable functions of the state \citep{Koopman1931}. In practice, the system is lifted to a higher-dimensional space (Fig.~\ref{fig:physics_encoded}) where its evolution is approximately linear, making Koopman models attractive for learning and control of nonlinear robotic systems. In this setting, the dictionary is the set of lifting functions mapping the state to the lifted space. Its choice critically affects accuracy, interpretability, and control performance.
Some works encoded physics through lifted architecture/dictionary design (Fig.~\ref{fig:physics_encoded}) and consistency constraints. \cite{Shi2022} exploited the topological structure of the robot's configuration space (for rigid robots) and workspace (for soft robots) to analytically construct polynomial-based Koopman lifting functions via Kronecker products. Their approach was applied for dynamics learning with a wheeled robot, a two-link robotic arm, and a soft robotic leg. Similarly, \cite{Joglekar2023} designed Koopman candidate functions for Ackermann-steered vehicles from Lie-bracket constructions informed by holonomic/nonholonomic constraints, and used the learned model for trajectory tracking. To learn the dynamics and control cart-pendulum and spherical mobile robots, \cite{Abraham2017} chose polynomial and Fourier lifting functions inspired by the systems' properties, such as pendulum angle periodicity on rotational manifolds. For nonholonomic mobile-robot navigation, \cite{Zhang2025} used a bilinear Koopman-\gls{mpc} formulation with observables built from physically meaningful kinematic coordinates, including yaw-angle trigonometric embeddings (\eg{} $\cos(\cdot)$, $\sin(\cdot)$), so the lifted model preserves heading geometry and control-affine structure under uncertainty.
In soft robotics, \cite{Shi2022_grippers} analytically designed Koopman lifting functions based on the workspace geometry and pneumatic-actuator structure of a soft multi-finger gripper, enabling physics-based dictionary design for online dynamics learning and control. \cite{Castano2020} exploited the dynamics structure and tail-actuation kinematics of a rigid-body robotic fish to construct Koopman lifting functions, with a physics-encoded basis-function design for learning and predictive control of soft robotic swimmer dynamics. \cite{Wang2024} further enforced physical consistency with an energy-preserving Koopman formulation that imposes hard energy-difference matching constraints for Lagrangian systems to improve dynamics learning and trajectory tracking with a cart-pole system.  
The reader can refer to \cite{Shi2026} for other examples of Koopman operators in robot learning.

Beyond Koopman methods, \cite{Wang2025_friction} used a \gls{kan} for static friction identification in industrial manipulators, retaining physically meaningful values of the trained parameters. 

\paragraph{Discussion:}
By lifting nonlinear robot dynamics to an approximately linear latent space, physics-encoded \glspl{no} enable linear and bilinear control methods, such as Koopman-\gls{mpc}, to be applied to nonlinear systems. Constructing the lifting functions from configuration-space topology, kinematic constraints, or actuator structure replaces generic dictionary learning with interpretable observables, and enables physical consistency to be enforced in the lifted space \citep{Shi2022,Abraham2017,Shi2022_grippers,Wang2024}. This is particularly valuable for soft and continuum robots, where minimal-coordinate models are difficult to derive.
However, designing the lifting functions remains highly application-specific, and no transferable methodology has yet emerged. Moreover, the lifted dimension grows rapidly with the state dimension, limiting current applications to low-\gls{dof} systems. Finally, physics-encoded operators beyond the Koopman framework remain largely unexplored.

\subsection{Other Types of Physics-Encoded Architectures} \label{sec:other_encoded}
This section reviews learning approaches that embed other forms of physics or domain knowledge into their architectures, and do not fit the previous categories.
\subsubsection{Architectures for Robot Motion Planning:}
This line of research focuses on tailoring \gls{nn} architectures to perform motion planning or prediction tasks in robotics.
\cite{Karle2023} proposed MixNet for motion prediction in autonomous racing, combining a \gls{nn} encoder with a physics-constrained output layer that predicts the weights of dynamically feasible base curves. Their design enforces smooth, physically consistent, in-track trajectories, and was deployed on an autonomous race car.
\cite{Piccinini2024_primitives} devised a polynomial \gls{nn} to learn the time-optimal maneuvers generated by optimal control problems for mobile robots and vehicles. Their \gls{nn} architecture is tailored to exactly satisfy the desired boundary conditions on the systems' physical states and derivatives. 
\cite{Hu2023} proposed a modular \gls{nn} architecture for in-flight object catching with a robot hand-arm system, combining five neural modules: trajectory prediction, catching-pose scoring, reaching policy, grasping policy, and a gating network that blends reaching and grasping actions. Their task-informed modular \gls{nn} was validated in dynamic catching problems.
\cite{Viereck2018} proposed a structured policy \gls{nn} for dynamic hopping that predicts physically meaningful control-law terms, including reference motion, feedforward torques, and feedback gains. Their control-informed architecture leverages learned contact-rich dynamics and was successfully transferred from simulation to a single-legged hopping robot.

Some authors designed \glspl{nn} to solve the Eikonal equation, a first-order nonlinear \gls{pde} for minimum-time motion planning and collision avoidance \citep{Clawson2014}. In particular, \cite{ni2023ntfields} devised a \gls{nn} architecture to solve the Eikonal equation with robot manipulators. Their architecture, derived from a ResNet, is conceived to respect the symmetry of the time field (model output), meaning the travel time is the same in both directions between any start and goal states. Their work was extended by \cite{ni2023progressive}, who introduced a progressive strategy to guide the \gls{nn} training in complex obstacle avoidance scenarios. 
Building on this line, \cite{Ren2024} extended neural time fields from manipulator path planning to prehensile manipulation in cluttered environments. They introduced an architecture and training pipeline for multimodal manipulation, including Dirichlet-energy minimization for faster convergence, an object-generalizable neural representation, and reactive replanning integrated with the learned time field.
\cite{Liu2025physics} further developed this approach to generate online travel time fields in unknown environments. Their \gls{nn} architecture is inspired by the spectral distance formulation to approximate the minimum travel time from a start to a goal state, integrates Fourier positional encoding, and has a sinusoidal activation function to learn high-frequency behaviors of the time field. 

Beyond planning-focused architectures, \cite{stolze2024} proposed input-to-state stable Coupled Oscillator Networks (CONs) to learn control-oriented latent dynamics from pixel observations. Their architecture embeds damped coupled-oscillator dynamics with a Lagrangian interpretation, provides global input-to-state stability guarantees via Lyapunov analysis, and was validated on continuum soft robots for trajectory tracking.

\paragraph{Discussion:}
Collectively, these architectures replace computationally expensive planning and optimal-control optimization with a single forward pass, in some cases without requiring expert demonstrations. They have also been deployed on diverse robotic platforms, including race cars \citep{Karle2023}, hopping robots \citep{Viereck2018}, and continuum soft robots \citep{stolze2024}. However, the embedded priors remain highly application-specific, with no transferable design methodology, motivating future research on  automated discovery of motion planning architectures. Moreover, the provided guarantees typically apply only to the assumed model class and output parameterization, rather than the full robot dynamics. Finally, time-field methods largely focus on static collision avoidance, with limited treatment of dynamic obstacles and actuation constraints.

\subsubsection{Bio-inspired Architectures:} \label{sec:bio_inspired}
A complementary line of research embeds priors derived from biological principles of perception, action selection, and neural representation into learning architectures. These methods encode physical insights in the form of structured computational mechanisms inspired by natural intelligence.

A first class of approaches adopts bio-inspired control architectures, where perception and action are coupled through structured competition mechanisms. For example, layered architectures based on the affordance competition hypothesis \citep{cisek2007cortical} were used by \cite{dalio2023complex} for evaluation and selection of multiple candidate actions via a dynamical competition process. The resulting architecture embeds action selection as an intrinsic computational mechanism, rather than an explicit optimization step, and was applied to autonomous driving and human-vehicle interaction \citep{dalio2021biasing}.

A second line of work embeds bio-inspired priors into the internal representations of learning models. Inspired by the circular neural structures underlying head-direction encoding in animals, \cite{plebe2023bio} proposed circular latent spaces, where rotations are represented as cyclic shifts, establishing a direct correspondence between latent transformations and physical rotations. Similarly, \cite{plebe2021occupancy} encoded the cortical magnification principle into occupancy-grid representations through a non-uniform spatial warping that allocates higher resolution to nearby regions. Together, these approaches demonstrate how biologically inspired latent spaces and representations can embed domain knowledge directly into the architecture, improving generalization and perception accuracy.

Finally, cognitive principles can also guide the learning of latent representations. Inspired by predictive coding and the free-energy principle \citep{friston2010free}, \cite{plebe2020road} structured a variational autoencoder around task-relevant latent variables and future-state prediction. By explicitly organizing the latent space according to semantically meaningful concepts (\eg{} obstacles and lanes) and predictive objectives, the architecture embeds cognitive priors that improve representation learning, interpretability, and scene prediction.

Overall, bio-inspired architectures embed inductive biases through structured control mechanisms, latent representations, and predictive objectives. They capture principles of natural intelligence to promote adaptive behavior and structured representations. Combining them with physics-based priors and model-structured architectures offers a promising direction toward robot learning systems that are both physically grounded and biologically inspired.

\subsubsection{Symmetry-Aware Architectures:} \label{sec:symmetry_aware}
Besides their role as geometric priors in data-driven methods, symmetries are fundamentally linked to conservation laws through Noether's theorem~\citep{watson2025machine}, motivating their use as inductive biases in robotics.
\cite{OrdonezApraez2023,morphologial_sym_2025} developed a framework to exploit morphological symmetries as physics-encoded geometric priors in robotics. They showed how these symmetries can be incorporated into learning-based methods for dynamics modeling, control, estimation, and design, either through model architectures or data augmentation. They showed improved sample efficiency and generalization over black-box baselines in centroidal momentum estimation and contact detection for legged robots. 
In the same experimental setting, \cite{pmlr-v283-xie25a} proposed a \gls{gnn} that encodes the robot kinematic tree and morphological symmetries for quadruped dynamics learning, obtaining improved generalization and data efficiency over black-box approaches.
\cite{symmetry_rl_legged_2024} incorporated morphological symmetries into \gls{rl} for legged locomotion and loco-manipulation tasks, using data augmentation and symmetry-constrained policy architectures. They achieved improved sample efficiency and task performance over unconstrained baselines.
Similarly, \cite{rl_ambidextrous_bimanual} used equivariant policies to enable ambidextrous bimanual manipulation, demonstrating task transferability across arms in both simulation and real-robot experiments.
\cite{Sharma2026} designed a physics-inspired \gls{gnn} for multi-body dynamical systems that enforces pairwise conservation of linear and angular momentum through edge-local reference frames equivariant to $SE(3)$ symmetries. Their architecture embeds momentum conservation as a hard inductive bias in the message-passing design, enabling interpretable and stable long-horizon predictions. Their method was validated on 3D granular systems with inelastic collisions, and was presented in the context of robotics and engineered mechanical systems.

\begin{table*}[t]
\scriptsize
\centering
\setlength{\tabcolsep}{2pt}
\renewcommand{\arraystretch}{1.1}
\begin{threeparttable}
\caption{Summary of generative-model approaches, reviewed throughout this survey, grouped by how physics is incorporated. Checkmarks indicate whether a paper embeds physics through architecture design (Physics-Encoded, P.E.), physics-based training loss functions (Physics-Informed, P.I.), and/or physically structured inputs, data or representations (Physics-Guided, P.G.).}
\label{tab:generative_models_summary}
\renewcommand{\tabularxcolumn}[1]{m{#1}}
\begin{tabularx}{\textwidth}{>{\centering\arraybackslash}m{2.2cm}|>{\centering\arraybackslash}m{1.95cm}|>{\centering\arraybackslash}m{1.9cm}|>{\centering\arraybackslash}m{0.55cm}|>{\centering\arraybackslash}m{0.55cm}|>{\centering\arraybackslash}m{0.55cm}|>{\centering\arraybackslash}m{1.55cm}|>{\centering\arraybackslash}X}
\toprule
\textbf{Paper} & \textbf{Application} & \textbf{Generative Model} & \textbf{P.E.} & \textbf{P.I.} & \textbf{P.G.} & \textbf{Robot} & \textbf{How Physics is Incorporated} \\
\hline
\multicolumn{8}{c}{\rule{0pt}{2.4ex}\textbf{\textit{Symmetry-Aware Architectures}}} \\
\hline
\cite{yang2024equibot} & \multirow{3}{=}{\centering Control} & \multirow{3}{=}{\centering Diffusion} & \checkmark &  &  & Mobile manipulator & SIM(3)-equivariant architecture for scaling/translation robustness \\\cline{1-1}\cline{4-8}
\cite{wang2024equivariant} &  &  & \checkmark &  &  & \multirow{2}{=}{\centering Manipulator} & SO(2)/T(3)-equivariant denoising policy for sample-efficient learning \\\cline{1-1}\cline{4-6}\cline{8-8}
\cite{Seo2025EquiContact} &  &  & \checkmark &  &  &  & SE(3)-equivariant pipeline from perception to action generation \\
\hline
\cite{tie2025etseed} & \multirow{2}{=}{\centering Trajectory planning and control} & \multirow{2}{=}{\centering Diffusion} & \checkmark &  &  & \multirow{2}{=}{\centering Manipulator} & Trajectory-level SE(3) equivariance with Markov-kernel denoising \\\cline{1-1}\cline{4-6}\cline{8-8}
\cite{Zhu2025SE3} &  &  & \checkmark &  &  &  & SE(3)-equivariant spherical-Fourier conditioning and denoising \\
\hline
\multicolumn{8}{c}{\rule{0pt}{2.4ex}\textbf{\textit{Hamiltonian Approaches}}} \\
\hline
\cite{cui2026physical} & Dynamics learning & \gls{vwm} & \checkmark &  &  & Manipulator & Latent energy-based dynamics with Hamiltonian structure \\
\hline
\multicolumn{8}{c}{\rule{0pt}{2.4ex}\textbf{\textit{\gls{pinn}-Style Losses and Reward Functions}}} \\
\hline
\cite{stylevla2026} & Trajectory planning and prediction & \gls{vla} &  & \checkmark &  & Vehicle & Kinematic-consistency loss on predicted trajectories \\
\hline
\cite{bouvier2025ddat} & \multirow{3}{=}{\centering Trajectory planning and control} & \multirow{2}{=}{\centering Diffusion} &  & \checkmark &  & Quadrotor / quadruped & Projection to reachable sets with dynamics-consistency penalties \\\cline{1-1}\cline{4-8}
\cite{Serifi2024Robot} &  &  &  & \checkmark &  & Humanoid & Differentiable physics-based tracking reward for policy alignment \\\cline{1-1}\cline{3-8}
\cite{Zhang2025MINDV} &  & \multirow{3}{=}{\centering \gls{vwm}} &  & \checkmark &  & \multirow{3}{=}{\centering Manipulator} & Physical-foresight coherence reward from a world-model referee \\\cline{1-2}\cline{4-6}\cline{8-8}
\cite{shang2026roboscape} & Trajectory planning and prediction &  &  & \checkmark &  &  & Loss terms for geometric temporal consistency and dynamics-consistent keypoint motion \\
\cline{1-8}
\cite{zeng2025physicsinformed} & Vehicle (state estimation) & Diffusion &  & \checkmark &  & Vehicle & Physics-conditioned embedding and joint denoising-physics loss for robust state estimation \\
\hline
\multicolumn{8}{c}{\rule{0pt}{2.4ex}\textbf{\textit{Physically Consistent World Representations}}} \\
\hline
\cite{Yang2025GenieDrive} & \multirow{9}{=}{\centering Trajectory planning and prediction} & \multirow{9}{=}{\centering \gls{vwm}} &  &  & \checkmark & \multirow{5}{=}{\centering Vehicle} & 4D occupancy representation, control-conditioned occupancy evolution, explicit space-time physical state structure \\\cline{1-1}\cline{4-6}\cline{8-8}
\cite{Zhou2026TowardPC} &  &  &  &  & \checkmark &  & Physical-condition generator repairs infeasible trajectories before video generation \\\cline{1-1}\cline{4-6}\cline{8-8}
\cite{wang2026ego} &  &  &  &  & \checkmark &  & Ego-dynamics context conditions latent transitions and imagination for cross-chassis adaptation \\\cline{1-1}\cline{4-8}
\cite{wang2026contactgaussianwmlearningphysicsgroundedworld} &  &  &  &  & \checkmark & \multirow{3}{=}{\centering Manipulator} & Joint rendering/contact representation and differentiable contact dynamics for physical-parameter learning \\\cline{1-1}\cline{4-6}\cline{8-8}
\cite{Romero2025Learning} &  &  &  &  & \checkmark &  & Object-level kinematic masks and interaction-focused conditioning/training data \\\hline
\cite{Buschoff2026} & Trajectory planning & \gls{vlm} &  &  & \checkmark & Stacking tasks & \gls{rl} interaction with physics-engine rewards based on stability \\
\hline
\cite{Mao2025RobotLF} & \multirow{3}{=}{\centering Control} & \multirow{3}{=}{\centering\gls{vwm}} &  &  & \checkmark & \multirow{3}{=}{\centering Manipulator} & Reconstructed physical world (mass/friction/gravity parameters)\\\cline{1-1}\cline{4-6}\cline{8-8}
\cite{pin_wm_2025} & & & & & \checkmark & & Differentiable rigid-body dynamics and Gaussian-Splatting visual loss identify inertia, friction, and restitution for Sim2Real transfer \\
\hline
\multicolumn{8}{c}{\rule{0pt}{2.4ex}\textbf{\textit{Action Generation with Physics Guidance}}} \\
\hline
\cite{Lv2026KinematicsAware} & \multirow{3}{=}{\centering Control} & \multirow{2}{=}{\centering Diffusion} &  &  & \checkmark & \multirow{2}{=}{\centering Manipulator} & Kinematic constraints between node- and joint-space actions \\\cline{1-1}\cline{4-6}\cline{8-8}
\cite{Du2025DynaGuide} &  &  &  &  & \checkmark &  & Gradient guidance from external dynamics model during denoising \\\cline{1-1}\cline{3-8}
\cite{tucker2026pi} &  & \gls{vla} &  &  & \checkmark & Aerial manipulator & Payload-aware physics guidance terms injected at inference \\\cline{1-8}
\cite{Xu2024Dynamics} & Manipulator design & \multirow{3}{=}{\centering Diffusion} &  &  & \checkmark & \multirow{3}{=}{\centering Manipulator} & Differentiable dynamics model guides diffusion sampling \\\cline{1-2}\cline{4-6}\cline{8-8}
\cite{Ma2024Hierarchical} & Trajectory planning and control &  &  &  & \checkmark &  & Differentiable kinematics distills end-effector diffusion trajectories into joint-space trajectories \\
\hline
\multicolumn{8}{c}{\rule{0pt}{2.4ex}\textbf{\textit{Physics-Guided Features and Training Data}}} \\
\hline
\cite{Jin2025DiffGen} & \multirow{3}{=}{\centering Trajectory planning and control} & \multirow{3}{=}{\centering Diffusion} &  &  & \checkmark & Manipulator & Differentiable simulation + differentiable rendering for physically plausible demonstrations \\\cline{1-1}\cline{4-8}
\cite{Xu2025PARC} &  &  &  &  & \checkmark & Legged robot & Controller-corrected generated motions recycled as physics-guided augmented data \\
\hline
\multicolumn{8}{c}{\rule{0pt}{2.4ex}\textbf{\textit{Physical Models as Structured Inputs to Learning Algorithms}}} \\
\hline
\cite{yang2026physical} & Estimation & \gls{vwm} &  &  & \checkmark & Vehicle & Physics modules in conditioning pathway (frame alignment, 3D flow guidance, box-coordinate guidance) \\
\bottomrule
\end{tabularx}
\end{threeparttable}
\end{table*}

\paragraph{Symmetry-Aware Generative Models:}
Symmetry-aware architectures have also been proposed in the context of generative models for robotics (Table \ref{tab:generative_models_summary}).
\cite{yang2024equibot} modified a diffusion model architecture to encode equivariance to scaling and translation (SIM(3)), yielding improved generalization to spatial transformations in mobile manipulation tasks. \cite{wang2024equivariant} proposed an equivariant diffusion policy for visuomotor manipulation, enforcing SO(2) rotational and T(3) translation symmetries in denoising to improve sample efficiency and generalization. \cite{tie2025etseed} enforced trajectory-level SE(3) equivariance on 3D rigid transformations, with an equivariant Markov-kernel condition in the denoising process. This enhanced spatial generalization and data efficiency for manipulation tasks. \cite{Zhu2025SE3} built SE(3)-equivariance in spherical Fourier space, including equivariant conditioning and denoising diffusion modules, and applied the method to trajectory generation and control with single-arm and bi-manual manipulation. \cite{Seo2025EquiContact} extended this line to contact-rich tasks by designing a whole SE(3)-equivariant pipeline from visual perception to action, explicitly coupling geometric equivariance and force-level action generation.

Overall, symmetry-aware \gls{ml} architectures leverage geometric priors to boost generalization and sample efficiency in robot learning tasks, showing improved performance over non-equivariant baselines.


\begin{table*}[t]
\scriptsize
\centering
\setlength{\tabcolsep}{2pt}
\renewcommand{\arraystretch}{1.12}
\begin{threeparttable}
\caption{Comparison of the main physics-encoded architecture families reviewed in Sec.~\ref{sec:encoded_architecture}. The table highlights the design trade-offs associated with different ways of embedding physics into learning architectures, including the encoded physics prior, the scenarios in which each family is most suitable, and the corresponding strengths and limitations. The comparison refers to architectural properties rather than measured performance.}
\label{tab:encoded_tradeoffs}

\renewcommand{\tabularxcolumn}[1]{m{#1}}

\begin{tabularx}{\textwidth}{
>{\centering\arraybackslash}m{1.6cm}|
>{\centering\arraybackslash}m{3.0cm}|
>{\centering\arraybackslash}m{3.8cm}|
>{\centering\arraybackslash}m{3.8cm}|
>{\centering\arraybackslash}X}

\toprule

\textbf{Family}
&
\textbf{Main Embedded Priors}
&
\textbf{When to Use}
&
\textbf{Main Strengths}
&
\textbf{Main Limitations}
\\

\hline

\gls{delan}
&
Energy conservation; energy decomposition; positive-definite inertia matrix
&
Non-conservative forces are known or negligible
&
Interpretable physical quantities; energy, inverse, and forward dynamics from one learned model; reusable in model-based control
&
Assumes collocated coordinates; dissipation and contact terms require separate models
\\

\hline

\gls{hnn}/port-Hamiltonian
&
Hamiltonian structure, symplectic geometry, passivity
&
Systems governed by energy exchange / dissipation / passivity
&
Energy-based interpretation; explicit treatment of dissipation and interconnection
&
Requires generalized momenta or their reconstruction from observable states; limited validation on high-\gls{dof} robots
\\

\hline

\gls{lnn}
&
Euler--Lagrange equations
&
General Lagrangian systems
&
Greater flexibility than \gls{delan} while retaining energy-based interpretation
&
Weaker structural guarantees and reduced interpretability than \gls{delan}
\\

\hline

\gls{msnn}
&
Physical principles in architectural layers / connections / constraints
&
Known dependencies among physical quantities;  force decomposition; bounds on physical quantities
&
Application-specific architecture design; interpretable trained sub-modules
&
Substantial domain expertise required; limited generality across robot classes; no prescribed design rules
\\

\hline

Topology learning, \gls{sindy}-like
&
Physics-based candidate operators designed from first principles
&
Discovering governing equations from data
&
Compact and interpretable models
&
Basis candidate design remains manual; limited scalability to contact-rich systems
\\

\hline

Neural operators
&
Physics-aware lifting functions and observables
&
Complex nonlinear dynamics
&
Enables linear control methods for underlying nonlinear dynamics
&
Application-specific lifting design; finite lifting limits exactness
\\

\hline

\glspl{node}, \glspl{vin}
&
Continuous-time dynamics; variational or Lie-group structure for \glspl{vin}
&
Continuous-time robot dynamics and long-horizon prediction
&
Flexible continuous-time modeling; structure preservation for \glspl{vin}
&
Limited embedded physics unless additional structure is imposed
\\

\hline

Hybrid physics-learning
&
Analytical models combined with learned subsystems / residuals
&
Partially-known dynamics with hard-to-model effects
&
Balances physical priors (analytical physics-based part) with flexible learning (data-driven part)
&
Guarantees / interpretability can apply only to the analytical component; data-driven components are black-boxes
\\
\bottomrule
\end{tabularx}
\end{threeparttable}
\end{table*}

\subsection{Trade-offs Among Physics-Encoded Architectures}

Table~\ref{tab:encoded_tradeoffs} summarizes the main trade-offs among the physics-encoded families reviewed in this section. These families are not competing solutions to the same problem: they embed different forms of physical prior knowledge, and the appropriate choice depends primarily on the available priors and the desired balance between structural bias and modeling flexibility.

When the robot dynamics are well understood, analytical mechanics provides the strongest priors. \gls{delan} are suitable when the kinetic/potential energy decomposition \eqref{eq:lag_definition} holds, and interpretable inertial and energy terms are desired. This also includes soft robots when a suitable generalized coordinate representation supports the energy decomposition.
\glspl{hnn} and port-Hamiltonian models are preferable when energy exchange, dissipation, and passivity are central, whereas \glspl{lnn} relax the energy decomposition at the cost of weaker structural guarantees. When only parts of the dynamics are known, \glspl{msnn} can encode specific physical dependencies directly in the architecture, while hybrid physics-learning methods augment analytical models with learned modular components. If the governing equations are unknown but expected to admit a compact representation with a few basis functions, topology learning and \gls{sindy}-like methods can recover them from data. Finally, neural operators are attractive when nonlinear robot dynamics should be controlled with linear tools, whereas \glspl{node} and \glspl{vin} target continuous-time dynamics learning and long-horizon prediction.

Stronger priors generally improve interpretability and embed more domain knowledge, but require accurate physical modeling and often additional measurements (\eg{} torques, accelerations, or generalized momenta). Conversely, weaker priors offer greater flexibility but rely more heavily on data and provide fewer structural guarantees, which in hybrid models apply only to the analytical component. 
Many families also require manual design choices, including the architectural layout in \glspl{msnn}, the candidate basis in \gls{sindy}-like methods, and the lifting functions in neural operators, for which no general design methodology exists. Overall, no family dominates, and two gaps are common to all of them, namely contact-rich dynamics and validation on high-\gls{dof} robots (Sec.~\ref{sec:future_directions}).

\section{Physics-Informed Loss Functions} \label{sec:physics_informed}

Physics-informed approaches provide a flexible and data-efficient framework for solving forward and inverse problems governed by differential equations, by embedding physical laws as soft constraints in the training loss functions. The combination of data-driven and physics-informed loss terms can improve generalization in data-scarce regions, and hence improve the physical consistency of the learned model. 
However, since physics priors are enforced through soft constraints (loss terms), a small training residual does not necessarily guarantee exact satisfaction of the governing equations at inference time. 

Following the seminal work of \cite{Raissi2019}, physics-informed \gls{ml} has gained increasing attention in robotics since 2022, 
with a balanced distribution across applications and robot classes (Fig.~\ref{fig:paper_distribution}). Notably, aerial robotics has the highest proportion of physics-informed methods (43\%), largely due to the availability of low-dimensional dynamical models that can be embedded into the training loss.

\subsection{Physics-Informed Neural Networks} \label{sec:pinns}

Originally proposed for solving complex \glspl{pde} across diverse physical domains \citep{Raissi2019,Meng2025}, \glspl{pinn} have recently been employed in robotics problems governed by \glspl{ode}, where physics-informed losses complement data-driven objectives to improve generalization in data-scarce regions \citep{Bianchi2024_Drones}.  
We emphasize that, unlike physics-encoded learning, physics-informed learning does \textit{not} embed physics priors into model architectures, which typically remain  conventional feedforward \glspl{nn}. Instead, physical knowledge is incorporated exclusively through the training loss.

In the setting of \glspl{pinn} \citep{Raissi2019}, the solution $u(t,x)$ of a \gls{pde} is approximated by a \gls{nn}, while the governing \gls{pde} is imposed as a soft constraint in the loss function for \gls{nn} training. Specifically, many physical systems can be expressed in the general form
\begin{equation}
\partial_t u(t,x) + \mathcal{N}[u(t,x);\lambda] = 0,
\end{equation}
where $\mathcal{N}[\cdot;\lambda]$ is a (possibly nonlinear) differential operator parameterized by $\lambda$. Using automatic differentiation, a physics-informed residual function
\begin{equation}
f(t,x) := \partial_t \hat{u}(t,x) + \mathcal{N}[\hat{u}(t,x)]
\end{equation}
is constructed, whose learnable parameters are inside the \gls{nn} approximator $\hat{u}(t,x)$. The model is then trained by minimizing a composite loss function of the form
\begin{equation}
\mathcal{J} = \underbrace{\frac{1}{N_u}\sum_{i=1}^{N_u} \left|u(t_i,x_i) - \hat{u}(t_i,x_i)\right|^2}_{\text{data loss} \; \mathcal{J}_{\text{data}}} 
+ \underbrace{\frac{1}{N_f}\sum_{j=1}^{N_f} \left|f(t_j,x_j)\right|^2}_{\text{physics loss} \; \mathcal{J}_{\text{physics}}}.
\label{eq:pinn_loss}
\end{equation}
The data loss term $\mathcal{J}_{\text{data}}$ in \eqref{eq:pinn_loss} enforces agreement with $N_u$ observed samples, including initial and boundary conditions or real system measurements. Conversely, the physics loss $\mathcal{J}_{\text{physics}}$ penalizes violations of the governing \gls{pde} at $N_f$ collocation points. By embedding prior physical knowledge into the training objective, \glspl{pinn} act as universal function approximators guided to satisfy conservation laws and other physical constraints. In robotics, they have been applied to learn \gls{nn} surrogates for \glspl{pde} arising in soft robotics and continuum mechanics, replacing expensive numerical solvers with real-time models. They have also been used to learn rigid-body dynamics governed by \glspl{ode}, where the physics-informed loss improves generalization in data-scarce regions by constraining the \gls{nn} to satisfy the governing equations. This reduces overfitting and is particularly valuable in robotics, where acquiring large labeled datasets is often expensive or impractical.

Table \ref{tab:physics_informed_loss_summary} summarizes the works reviewed in this section by application class, including the robot type and the physical principles embedded in their loss functions.

\paragraph{Dynamics Learning and Estimation:}
In vehicle dynamics, \cite{sun2024trajectory} trained an \gls{mlp} with a hybrid loss combining data fitting and physics-based regularization, penalizing deviations from an analytical single-track model. Similarly, \cite{cortese2026,Cortese2026_TVT} regularized the training loss of \glspl{mlp} and \glspl{lstm} with multi-body vehicle models to estimate the lateral velocity, outperforming purely data-driven approaches. Additional vehicle applications of \glspl{pinn} include predicting the energy consumption of electric vehicles \citep{Lim2025} and learning tractor--trailer dynamics \citep{Zeipel2024}.
In soft and continuum robotics,
\cite{Beaber2024} proposed a \gls{pinn} surrogate framework to predict the deformation of pneumatic soft robots. They embedded the governing continuum-mechanics \glspl{pde} as a \gls{pinn} loss, alongside boundary and initial conditions. Once trained, the surrogate replaces expensive \glspl{pde} numerical solvers, enabling real-time deformation prediction of soft fingers. \cite{Bensch2024} used a \gls{pinn} to learn the static shape of a tendon-driven continuum robot. They integrated the corresponding \glspl{pde} into the training loss, with boundary values obtained from a shooting-method reference solver. The learned model predicts the full robot shape significantly faster than the reference, enabling use inside sampling-based path planners. 
\cite{Wang2024PINNRay} extended this line in soft grasping by injecting minimum-potential-energy principles from elastic mechanics into the loss of a \gls{pinn} surrogate for Fin-Ray soft robotic fingers. They reported improved deformation prediction over finite-element references after sim-to-real data assimilation.
In aerial robotics, \cite{Bianchi2024_Drones} applied \glspl{pinn} to estimate the thrust and drag parameters of a quadrotor, embedding the rigid-body equations of motion as \gls{ode} residuals in the training loss. With limited flight data, their \gls{pinn}-based estimator outperforms an \gls{ekf} baseline.
Some authors also showed how physics-regularized losses improve robustness under distribution shifts. For example, \cite{Serrano2024SlungLoadPINN} added a discretized first-principles consistency term (with slack variables) to a sequence predictor for multirotor slung-load systems, and outperformed both a pure physical model and a data-only baseline. \cite{Wang2025PIWAN} combined temporal convolutions with a wind-aware physics loss for quadrotor dynamics learning, improving prediction and \gls{mpc} tracking in unknown environments. \cite{Liu2025QuadStateOnline} used an online continual physics-informed objective for quadrotor state estimation under wind disturbances, showing improved adaptation during streaming updates. For ground robots, \cite{Sahoo2025MoRPIPINN} embedded inertial-navigation constraints into a lightweight \gls{pinn}, and reported drift reduction for pure-inertial mobile robot localization.

\paragraph{Control:}
\cite{Bolderman2022_feedforward} and \cite{Bolderman2024} combined a physics-based model with an \gls{mlp} to control the dynamics of a linear motor. Their loss function contains a regularization term that trusts more the physics-based model in regions of the state space with limited or no training data, thereby improving the model's extrapolation capabilities. \cite{Bolderman2024} also derived the conditions for the input-to-state stability of the resulting controller. \cite{nicodemus2022physics} employed \glspl{pinn} for nonlinear \gls{mpc} with a multi-link robotic manipulator. They incorporated control inputs and initial conditions into their \gls{nn}, allowing \glspl{pinn} to serve as surrogate models for dynamical systems. \cite{Saviolo2022} presented a physics-informed temporal convolutional network for quadrotor dynamics learning and \gls{mpc}-based trajectory tracking. They used physics constraints based on Newton-Euler equations as additional loss terms during training, fostering generalization beyond the training distribution. \cite{Antonelo2024} extended standard \glspl{pinn} to accommodate control inputs and variable-length time horizons via autoregressive chaining. They included the \gls{ode} residuals of a system's dynamics in the training loss without requiring labeled data. They performed dynamics learning and \gls{mpc}-based control of nonlinear systems, including a Van der Pol oscillator and an electric submersible pump. \cite{Li2025_vibration} proposed a \gls{pinn}-based input-shaping method to suppress vibrations in flexible single-link robotic arms. Their training loss combines the \gls{pde} constraints of a flexible beam's modal equations with vibration modal conditions, and a two-phase training procedure identifies the optimal impulse sequence.
\cite{Sanyal2023RAMPNet} further combined physics and data losses in robust adaptive \gls{mpc} for quadrotors: an ODE-based physics term regularizes the model under parametric uncertainty, while a data term captures residual disturbances, improving trajectory-tracking performance over data-driven \gls{mpc} baselines.

\paragraph{Trajectory Planning:}
Besides the architectural symmetry discussed in Sec.~\ref{sec:other_encoded}, the neural time fields of \cite{ni2023ntfields,ni2024physics} for time-optimal manipulator motion planning are trained with a physics-informed loss built on the residuals of the Eikonal equation (a first-order nonlinear \gls{pde}), which removes the need for expert demonstrations. \cite{Lee2023} modeled contactless handovers between collaborative robots using an \gls{lstm}-\gls{mlp} architecture. Their training relies on a physics-informed loss based on energy functions derived from the system dynamics, enabling prediction of the catching points of flying packages.

In the context of generative models (Table \ref{tab:generative_models_summary}), \cite{stylevla2026} trained a \gls{vla} with adaptive driving styles for vehicle trajectory planning, and used a loss term that penalizes violations of kinematic laws on predicted trajectories, improving generalization.

\paragraph{Physics-Informed Losses for Lagrangian \& Hamiltonian NNs:}


\gls{delan}, \glspl{lnn}, and \glspl{hnn} are commonly trained using forward or inverse dynamics errors. Since these approaches also provide an energy model, additional physics-informed losses can be considered. In particular, the rate of change of the total mechanical energy $\dot{\ESystem}$ equals the input power, \ie{} $\genVel\Transp\genForce$. Based on this relation, \cite{delan_4ec} adopted an auxiliary loss that matches the measured power input with the time derivative of the learned energy,
\begin{equation}
    \label{eq:loss_power_input}
    \loss_p = \Big|\Big|\genVel\Transp\genForce - \dot{\hat{\ESystem}}\Big|\Big|_2^2.
\end{equation}
In their formulation, generalized forces are decomposed as $\genForce = \genForce_m + \genForce_f$, where $\genForce_m$ denotes the measured input torques and $\genForce_f$ the estimated friction torques. This auxiliary objective is particularly useful when friction is explicitly modeled, since it promotes energy consistency and passivity of the learned system. 

Since the Hamiltonian $\Hamiltonian$ represents the total system energy, $\dot{\ESystem} = \dot{\Hamiltonian}$, the power consistency loss in \eqref{eq:loss_power_input} can also be combined with any \gls{hnn} method.


To promote temporal coherence and smoothness of the learned kinetic and potential energies $(\EKin,\EPot)$ in \eqref{eq:lag_definition}, \cite{delan_4ec} introduced a loss term that penalizes the deviation of the predicted energy at time $t+\Delta t$ and its first-order Taylor approximation from time $t$,
\begin{equation}
\label{eq:loss_energy_coherence}
\begin{aligned}
    \loss_e &= \Big|\Big|\EKin_{t+\Delta t} - \text{sg}\Big(\EKin_{t} + \dot{\EKin}_{t}{\Delta t}\Big)\Big|\Big|_2^2\\
    & + \Big|\Big|\EPot_{t+\Delta t} - \text{sg}\Big(\EPot_{t} + \dot{\EPot}_{t}{\Delta t}\Big)\Big|\Big|_2^2\,,
\end{aligned}
\end{equation}
where $\text{sg}(\cdot)$ denotes the stop-gradient operator. Since the true energy values at $t+\Delta t$ are not directly observed, the approximated terms serve as bootstrapped targets, with gradients propagated only through the directly predicted energies $\EKin_{t+\Delta t}$ and $\EPot_{t+\Delta t}$. This auxiliary loss regularizes the energy representation, yielding a smoother energy model that is validated on both simulated and real robots for energy-based control.

\begin{table*}[t]
\scriptsize
\centering
\setlength{\tabcolsep}{2pt}
\renewcommand{\arraystretch}{1.12}
\begin{threeparttable}
\caption{Summary of the reviewed papers using physics-informed loss functions (Sec.~\ref{sec:pinns}), grouped by application category.}
\label{tab:physics_informed_loss_summary}
\renewcommand{\tabularxcolumn}[1]{m{#1}}
\begin{tabularx}{\textwidth}{>{\centering\arraybackslash}m{2.5cm}|>{\centering\arraybackslash}m{3.2cm}|>{\centering\arraybackslash}m{2.3cm}|>{\centering\arraybackslash}X}
\toprule
\textbf{Paper} & \textbf{Application} & \textbf{Robot} & \textbf{Physical Principles Embedded in the Loss Function} \\
\hline
\multicolumn{4}{c}{\rule{0pt}{2.5ex}\textbf{\textit{Dynamics Learning}}}\\
\hline
\cite{sun2024trajectory,Lim2025,Zeipel2024} & Vehicle dynamics and energy-related dynamics prediction & Vehicle & Residual penalties w.r.t. analytical vehicle models (single-track / tractor-trailer) and physics-based regularization terms \\
\hline
\cite{Beaber2024,Wang2024PINNRay} & Soft-robot deformation surrogate modeling & Soft robot & Continuum-mechanics constraints and minimum-potential-energy principles in the training objective \\
\hline
\cite{Bensch2024} & Continuum robot shape learning & Continuum robot & Cosserat-rod-inspired \gls{pde} residual losses with boundary-condition consistency \\
\hline
\cite{Serrano2024SlungLoadPINN,Wang2025PIWAN} & Aerial system dynamics modeling under disturbances & Quadrotor / multirotor with slung load & Discretized first-principles dynamics consistency, Newton--Euler-informed residuals, and wind-aware physics penalties \\
\hline
\multicolumn{4}{c}{\rule{0pt}{2.5ex}\textbf{\textit{Trajectory Planning}}}\\
\hline
\cite{ni2023ntfields,ni2024physics} & Time-optimal motion planning & Manipulator & Eikonal-equation residual loss without expert labels \\
\hline
\cite{Lee2023} & Contactless handover prediction for planning & Collaborative robot & Energy-function-based physics-informed loss from handover dynamics \\
\hline
\cite{stylevla2026} & Vehicle trajectory planning with adaptive driving styles & Vehicle & Kinematic-law consistency penalties on predicted trajectories \\
\hline
\multicolumn{4}{c}{\rule{0pt}{2.5ex}\textbf{\textit{Control}}}\\
\hline
\cite{Bolderman2022_feedforward,Bolderman2024} & Feedforward motor control & Linear motor & Physics-trust regularization favoring physics-based predictions in data-scarce regions \\
\hline
\cite{nicodemus2022physics,Saviolo2022,Sanyal2023RAMPNet} & \gls{mpc}-based trajectory tracking & Manipulator / quadrotor / aerial robot & \gls{ode} or Newton--Euler residual losses combined with data terms \\
\hline
\cite{Antonelo2024,Li2025_vibration,delan_4ec} & Model learning for control with physical consistency \& passivity & Flexible arm / robot dynamics models & Dynamics-residual constraints, modal \gls{pde} losses, and power-consistency energy losses \\
\hline
\multicolumn{4}{c}{\rule{0pt}{2.5ex}\textbf{\textit{Estimation}}}\\
\hline
\cite{cortese2026,Cortese2026_TVT} & Vehicle lateral-velocity estimation & Vehicle & Multi-body vehicle-model residual regularization \\
\hline
\cite{Bianchi2024_Drones,Liu2025QuadStateOnline} & Aerial state/parameter estimation under disturbances & Quadrotor & Rigid-body dynamics residuals and continual physics-consistency terms \\
\hline
\cite{Sahoo2025MoRPIPINN} & Inertial localization and navigation-state estimation & Mobile robot & Inertial-navigation kinematic/dynamic equations \\
\bottomrule
\end{tabularx}
\end{threeparttable}
\end{table*}

\paragraph{Discussion:}
The main appeal of physics-informed losses is their architectural flexibility: since physics enters only through the training objective, they can be combined with diverse learning models, from \glspl{mlp} to \glspl{lstm} and \glspl{vla}, without modifying their architecture. They also have the potential to reduce reliance on labeled data by training directly from physics-based equation residuals, improve extrapolation in data-scarce regimes, and provide computationally efficient surrogate models for soft and continuum robots \citep{Bensch2024}.

Their main limitation is the counterpart of this flexibility: physical laws are enforced only as soft training constraints, so low loss residuals do not guarantee exact physical consistency at inference (Fig.~\ref{fig:lifecycle}). Performance depends on balancing the data and physics losses and on the fidelity of the underlying equations, since inaccurate priors introduce structural bias. Moreover, evaluating residuals requires governing equations in closed, differentiable form, limiting applicability to contact-rich and other nonsmooth systems while increasing training cost. Promising directions include adaptive loss weighting, uncertainty-aware formulations \citep{Bolderman2024,Sanyal2023RAMPNet}, and standardized reporting of physics-residual metrics.

\subsection{Physics-Informed Neural Operators} \label{sec:neural_operators_loss}

Physics-informed losses were used to guide neural operator learning when data alone were insufficient. In autonomous driving, \cite{Zhang2026_physics} introduced an acceleration-based physics loss (between predicted and measured accelerations) and combined it with online sliding-window updates, using the method for vehicle dynamics learning with a Koopman operator. In soft robotics, \cite{Ristich2025} decomposed the \gls{pde}-based dynamics of a soft arm into a passive physical term and unknown actuation forces. They learned two Koopman operators: one for the physical term, trained with a \gls{pde}-residual loss from phase-space samples (no trajectory data), and one for the unknown term, learned from trajectories. The two are then combined via Strang splitting. \cite{Kamath2026} trained a Koopman neural operator by combining a physics-informed loss that penalizes deviations from a nominal quadrotor dynamics, and a real-data loss from onboard flight experiments. Their model was then embedded in an \gls{mpc} for trajectory tracking. 

\cite{Tan2025_vehicle} proposed a physics-informed \gls{deeponet} \citep{Lu2021_DeepONet} for vehicle dynamics modeling. Their architecture was trained using a physics-informed loss built from known linear relations among vehicle states and analytical vehicle models. \cite{Sivakumar2026} adopted a \gls{fno} for quadrotor crash-area estimation under actuator failures and wind disturbances. They used a custom physics-informed loss function encoding mass conservation. 
\cite{Chen2024} adopted a \gls{gno} for autonomous vehicle navigation in traffic networks, and used a physics-informed loss to enforce consistency with the Fokker-Planck equation, capturing the dynamics evolution of the vehicles on the graph. \cite{lim2026_vega} used a \gls{pino} module to estimate the physical parameters of a vehicle (including drag, mass, rolling resistance) from measured speed and acceleration windows, and used a hybrid data-plus-physics residual loss.

\subsection{Other Types of Loss and Reward Functions}

To learn discontinuous contact dynamics in robot manipulation, \cite{Pfrommer2020} introduced a convex loss function that promotes physical consistency by activating contact forces only upon contact, preventing penetration, and maximizing frictional power dissipation. In learning stable motion primitives for manipulation planning, \cite{perez2023stable} proposed a contrastive \gls{il} loss combining imitation and stability terms. The contrastive component (pairwise, with a triplet-loss variant) enforces matching and separation of latent-state components, allowing the learned primitive to inherit global asymptotic stability from a stable reference latent dynamical system. \cite{donati_2025a} enforced algebraic admissibility conditions, including physical limits, symmetry, positive semi-definiteness, and the triangle inequality for inertial parameters, as penalties in a multi-step identification objective. Unlike \gls{pinn}-style losses, they constrain the identified model and parameters rather than governing-equation residuals, which are instead captured by the physics-based model component (Sec.~\ref{sec:hybrid_physics_nn}). The method was applied to  vehicle lateral dynamics learning.

\textit{Generative Models:}
\cite{bouvier2025ddat} projected the predictions of a diffusion policy onto dynamically admissible sets, with a loss term that penalizes deviations from under-approximated reachable sets of the dynamics during training and inference. Their approach improved the feasibility of the planned trajectories for quadrotors and quadrupeds. \cite{Zhang2025MINDV} introduced a physical-foresight coherence reward based on a pre-trained world-model referee, and used \gls{rl} post-training to favor dynamically coherent long-horizon manipulation videos, supporting planning and control-oriented \gls{vwm} rollouts. \cite{Serifi2024Robot} learned a differentiable surrogate reward that approximates a non-differentiable controller's behavior, and used it to fine-tune a diffusion policy for humanoid motion generation. This aligns the policy with the controller's tracking performance, improving the physical feasibility of text-conditioned motions.
\cite{zeng2025physicsinformed} proposed a physics-informed diffusion framework for system state estimation that combines a physics-conditioned embedding with a first-principles joint loss unifying denoising and physics residual terms. They validated the method on vehicle-related tasks, including dynamic wheel-load estimation and vehicle-state tracking under a vehicle-dynamics model, showing improved estimation accuracy and robustness under nonlinear dynamics and noisy observations.
\cite{shang2026roboscape} trained a \gls{vwm} with physics-informed regularization terms in their loss function to enforce geometric consistency in the temporal depth dimension across video frames, and dynamics consistency with adaptive keypoints, to learn the motion and deformation patterns of dynamic objects in the scene. Their \gls{vwm} was applied in dynamics simulation and policy evaluation for manipulation tasks. 

Collectively, these works show that physics-informed loss functions can be effectively integrated into generative models, including diffusion models and \glspl{vwm}, to enhance the physical plausibility and consistency of generated outputs. 
A summary of the generative model-based approaches is provided in Table \ref{tab:generative_models_summary}.


\section{Physics-Guided Inputs, Data, and Representations}
\label{sec:guided_inputs}
Physics-guided learning exploits physics priors to transform, enrich, curate, select, or correct input features, training data, and learned representations. Rather than modifying the model architecture or training objective, these methods embed physical knowledge into the inputs, data, or representations before training or as pre- or post-processing guidance at inference, while preserving the flexibility of standard \gls{ml} models. Compared with physics-encoded and physics-informed approaches, they are less mature (16\% of the surveyed papers, as shown in Fig.~\ref{fig:paper_distribution}) but have grown rapidly since 2022. 
They generally require less domain expertise, are easier to implement, and can be integrated into most learning algorithms. Notably, most generative and foundation-model  approaches adopt physics guidance to improve their physical consistency (Table~\ref{tab:generative_models_summary}), primarily for trajectory planning and prediction. 

Table \ref{tab:physics_guided_summary} summarizes the reviewed papers in this section, grouped by application category, indicating the robot type and how physics is incorporated. Papers on physics-guided neural operators are covered separately in Table~\ref{tab:neural_operator_summary}.

\begin{table*}[t]
\scriptsize
\centering
\setlength{\tabcolsep}{2pt}
\renewcommand{\arraystretch}{1.12}
\begin{threeparttable}
\caption{Summary of the reviewed papers on physics-guided inputs, data, and representations (Sec.~\ref{sec:guided_inputs}), grouped by application category. Papers using generative models (Sec.~\ref{sec:physics_guided_world_models}) are collected in Table~\ref{tab:generative_models_summary}, while the ones about physics-guided neural operators (Sec.~\ref{sec:neural_operators_guided}) are reported in Table~\ref{tab:neural_operator_summary}.}
\label{tab:physics_guided_summary}
\renewcommand{\tabularxcolumn}[1]{m{#1}}
\begin{tabularx}{\textwidth}{>{\centering\arraybackslash}m{2.6cm}|>{\centering\arraybackslash}m{3.0cm}|>{\centering\arraybackslash}m{2.3cm}|>{\centering\arraybackslash}X}
\toprule
\textbf{Paper} & \textbf{Category} & \textbf{Robot} & \textbf{How Physics is Incorporated in Inputs / Data / Representations} \\
\hline
\multicolumn{4}{c}{\rule{0pt}{2.5ex}\textbf{\textit{Dynamics Learning}}}\\
\hline
\cite{degroote2021neural} & \multirow{4}{*}{\shortstack{Physical models as\\structured \gls{nn} inputs}} & Slider-crank mechanism & Physics-based model computes displacement and velocity features as \gls{mlp} inputs \\
\cline{1-1}\cline{3-4}
\cite{delan_non_sym_friction} &  & Manipulator & Selection of physically meaningful inputs  and Fourier-series excitation trajectory design, matched to the robot's natural modes \\
\cline{1-1}\cline{3-4}
\cite{robinson2022} &  & Mechanical oscillator & Features computed from physics-based equations injected into \gls{mlp} hidden layers \\
\cline{1-1}\cline{3-4}
\cite{wu2025individualized} &  & Lower-limb prosthetics & \gls{dct} encoding of human kinematics frequency content \\
\hline
\multicolumn{4}{c}{\rule{0pt}{2.5ex}\textbf{\textit{Trajectory Planning}}}\\
\hline
\cite{chen2023} & \multirow{5}{*}{Geometric learning} & Mobile robot & Riemannian flow matching with geometrically consistent vector field on $SE(3)$ \\
\cline{1-1}\cline{3-4}
\cite{wang2022learning} &  & \multirow{4}{*}{Manipulator} & States projected onto $SO(3)$ manifold tangent space \\
\cline{1-1}\cline{4-4}
\cite{alhousani2023geometric} &  &  & States and actions represented on $SO(3)$ Riemannian manifold \\
\cline{1-1}\cline{4-4}
\cite{james2022bingham} &  &  & Bingham distribution over $SO(3)$ for rotation-uncertainty policy parametrization \\
\cline{1-1}\cline{4-4}
\cite{chisarilearning} &  &  & Riemannian flow matching from point cloud observations \\
\hline
\multicolumn{4}{c}{\rule{0pt}{2.5ex}\textbf{\textit{Control}}}\\
\hline
\cite{hong2023physics} & \multirow{2}{*}{\shortstack{Physical models as\\structured \gls{nn} inputs}} & Quadrotor & Residuals of physics-based model and physical state variables as \gls{rnn} inputs \\
\cline{1-1}\cline{3-4}
\cite{Bolderman2021} &  & Linear motor & Physics-guided input layer encodes friction dependence on position and signed velocity \\
\hline
\multicolumn{4}{c}{\rule{0pt}{2.5ex}\textbf{\textit{Estimation}}}\\
\hline
\cite{graber2018hybrid} & \multirow{5}{*}{\shortstack{Physical models as\\structured \gls{nn} inputs}} & \multirow{3}{*}{Vehicle} & Physics-based kinematic model's output fed as structured input to a \gls{gru} \\
\cline{1-1}\cline{4-4}
\cite{cortese2026} &  &  & \gls{ukf} with physics-based vehicle model provides preliminary estimate as \gls{mlp} input \\
\cline{1-1}\cline{3-4}
\cite{saleh2024physics} &  & Manipulator & Physics-based node features (forces, displacements) on graph; continuum-mechanics interaction priors in \gls{gnn} \\
\cline{1-1}\cline{2-4}
\cite{Elazab2026} & Physics-guided features & Quadruped & Biomechanics-informed feature engineering for fault detection/classification \\
\hline
\cite{shah2016encoding,falco2019representing} & Frequency-domain learning & Human-robot interaction & Fast Fourier transform of input data to capture kinematics patterns \\
\bottomrule
\end{tabularx}
\end{threeparttable}
\end{table*}
\subsection{Physical Models as Structured Inputs to Learning Algorithms}
This class of methods uses physics-based models to compute structured input features for downstream \gls{ml} algorithms.

To estimate a vehicle's sideslip angle, \cite{graber2018hybrid} used a physics-based kinematic model to provide structured inputs to a \gls{gru} network, which captures temporal dependencies and compensates for modeling inaccuracies. Similarly, \cite{degroote2021neural} used a physics-based model of a slider-crank mechanism to compute structured kinematic features for an \gls{mlp}, improving sample efficiency and generalization while constraining learning to a physically consistent representation.
\cite{cortese2026} used a physics-based vehicle model within a \gls{ukf} to generate a preliminary estimate of a vehicle's lateral velocity, which was refined by an \gls{mlp}. The resulting hybrid estimator outperformed both the standalone \gls{ukf} and an \gls{lstm}. In the context of linear motor control, \cite{Bolderman2021} used a physics-guided input layer to encode the dependency of mechanical friction on position and signed velocity, and fed the resulting features to an \gls{mlp} controller. \cite{hong2023physics} adopted an \gls{rnn} to process the residuals of a physics-based model, and used it for quadrotor dynamics learning in an \gls{mpc} trajectory tracking application. Input features include physically meaningful state variables (\eg{} positions, velocities, and control inputs), ensuring that the learned model operates in a space aligned with the governing equations of the system. 
\cite{yang2026physical} developed a driving \gls{vwm}, and injected physics through dedicated modules in the conditioning pathway: a coordinate-system aligner (relative/absolute motion consistency), an instance-flow guidance module (temporal consistency from 3D flow), and box-coordinate guidance (spatial/occlusion consistency). This supports prediction tasks, and improves downstream estimation and perception metrics. 

Across these works, physics-based models provide structured inputs or physically meaningful representations, yielding improved sample efficiency and generalization over purely data-driven approaches. On the other hand, they require domain knowledge to design the physics-based models, and may be limited by the accuracy of these models in capturing complex dynamics.

\subsection{Physics-Guided Features \& Training Data} \label{sec:guided_features}
This category of approaches exploits physics priors to design input features or training data for downstream learning tasks.

\cite{delan_non_sym_friction} used a \gls{delan} to learn non-symmetric Coulomb friction in robot manipulators. Physics priors are embedded through physically meaningful inputs (joint positions, velocities, and accelerations), reflecting the underlying mechanics and friction forces. Also, they designed informative training trajectories, based on finite Fourier series matched to the robot's natural modes.
\cite{robinson2022} injected partial knowledge of the system dynamics into the hidden layers of \glspl{mlp} via engineered input features computed from analytical equations. Applied to mechanical oscillators, this improved training speed and accuracy. \cite{Elazab2026} proposed a physics-guided framework for fault detection in quadruped robots. Biomechanics knowledge was embedded through physics-based features derived from raw sensor measurements, including gait symmetry, center of pressure, and an energy proxy. Their two-stage soft-voting architecture outperformed black-box deep learning in simulated fault scenarios.
\cite{saleh2024physics} proposed a physics-guided \gls{gnn} to estimate contact-induced deformations in robotic manipulation. Deformable objects are represented as graphs, where each node's physical features (\eg{} forces and displacements) are encoded by an \gls{mlp} and processed by the \gls{gnn} to capture global interactions. They also adopt a physics-encoded architecture enforcing locality and interaction patterns consistent with continuum mechanics, mimicking force propagation through the material.

Other authors incorporated physics primarily through data generation and augmentation. \cite{Jin2025DiffGen} used differentiable physics simulation and differentiable rendering, optimized with \gls{vlm}-based objectives, to synthesize physically plausible robot manipulation demonstrations from textual instructions, for downstream learning tasks. \cite{Xu2025PARC} proposed an iterative loop in which motions generated by a diffusion model are corrected by physics-based tracking controllers and recycled as augmented data. This improves agile terrain-traversal behavior for character human-like agents, supporting planning and control tasks.

Taken together, these methods provide one of the least invasive ways to embed physics: by acting on features and data, they are compatible with most learning algorithms. Data-centric approaches are particularly attractive, as they address robot data scarcity directly through physics-guided data generation and curation \citep{Jin2025DiffGen,Xu2025PARC}.
Their main limitation is that their physics priors do not constrain the learned model or its predictions, and data-level priors are inactive at inference time. Moreover, feature engineering and excitation design remain largely manual and application-specific, while synthetic data are limited by the fidelity of the underlying simulators. 

\subsection{Geometric Learning} \label{sec:geometric_learning}
Some approaches modified input and output features to preserve non-Euclidean geometric structures of the data. In the context of \gls{il} \citep{wang2022learning} and \gls{rl} \citep{alhousani2023geometric} with collaborative robot manipulators, input data were projected onto the tangent space of a target manifold (e.g., $SO(3)$ or symmetric and positive definite matrices), training was performed there, and outputs were mapped back to the manifold. \cite{james2022bingham} developed a \gls{rl} neural policy parametrization that leverages the Bingham distribution to model uncertainty over 3D rotations, allowing policy search to directly sample points in $SO(3)$ for robotic manipulation. \cite{chen2023} introduced a Riemannian flow matching framework for navigation, where the input is preprocessed to define a geometrically-consistent target vector field, and a Riemannian integration step is used to produce the output. Their loss function is formally equivalent to standard flow matching, although it uses a geometrically-consistent vector field as target. A Riemannian flow matching was also used by \cite{chisarilearning} to learn robotic manipulation policies from point cloud observations. 

Overall, the exploitation of geometric priors in input representations was shown to improve learning performance in tasks involving rotations and geometric constraints, by preserving the underlying manifold structure of the data.

\subsection{Frequency-Domain Learning} \label{sec:frequency_domain}
A series of works proposed to transform data into the frequency domain to improve learning performance. We regard these methods as physics-guided only when the retained frequency components, or the structure imposed on them, follow from known physical properties of the system.
\cite{falco2019representing} proposed a frequency-domain descriptor of human motion for human-robot interaction, explicitly exploiting the biomechanical property that human motion contains no significant frequency components beyond $10$~Hz. This bound sets the cut-off of the representation and yields both compression and invariance to the duration of the action. The same principle was employed by \cite{shah2016encoding}, where the low-pass bandwidth of human actions produces a compact descriptor of segmented actions, whose size is independent of the motion duration.
\cite{wu2025individualized} used a \gls{dct} for the individualized modeling of lower-limb joint kinematics in prosthetics. Here, the transform is applied to gait cycles normalized by the gait phase, so that the periodicity of the representation follows from the cyclic nature of human locomotion, while the number of retained coefficients is selected with a cumulative-energy criterion.

\subsection{Physically Consistent World Representations} \label{sec:physics_guided_world_models}
Some authors designed Video World Models (\glspl{vwm}) with explicit measures to enforce physically consistent representations in the generated videos (Table \ref{tab:generative_models_summary}). 

In autonomous driving, the \gls{vwm} of \cite{Yang2025GenieDrive} was conditioned on physical states, using 4D occupancy as an intermediate representation with explicit space-time structure, a tri-plane \gls{vae} for occupancy compression, and mutual-control attention to couple controls with occupancy evolution. Their generated videos were shown to improve motion prediction. \cite{Zhou2026TowardPC} added a \textit{physical condition generator} to transform infeasible trajectories from planners/simulators into physically consistent ones before video generation. This was coupled with a physics-enhanced multi-view video generator, trained on mixed real and simulated scenarios, targeting robust planning and prediction in autonomous driving. In manipulation, \cite{Mao2025RobotLF} coupled \gls{vwm}-based video generation with physical world reconstruction, and learned an object-centric residual policy in the reconstructed simulator. 
They queried an external \gls{vlm} to estimate mass and friction parameters from reconstructed images, and used these parameters to enhance physical simulation. Also, they used geometric transformations for gravity alignment, to ensure the reconstructed world is gravity-aligned.
\cite{wang2026ego} introduced an ego-dynamics-augmented driving \gls{vlm}, where a physics-aware ego-state encoder-decoder conditions latent world-model transitions and imagination rollouts. By explicitly separating ego-motion dynamics from scene evolution, they improved long-horizon prediction quality and enabled zero-shot cross-chassis adaptation.
\cite{pin_wm_2025} trained a \gls{vwm} for non-prehensile manipulation by identifying rigid-body dynamics parameters (inertia, friction, and restitution terms) from visual observations, using a differentiable physics simulator and Gaussian-Splatting rendering loss. The resulting \gls{vwm} embeds contact-rich physical priors into the learned dynamics, and uses physics-aware randomization to generate digital cousins for robust policy transfer in pushing and poking tasks.
\cite{wang2026contactgaussianwmlearningphysicsgroundedworld} pushed this integration further by unifying video rendering and collision geometry with the same Gaussian representation, and by differentiating through a contact dynamics engine to learn physical properties like mass, inertia, friction and stiffness from visual inputs, in an end-to-end fashion. This makes their learned \gls{vwm} physically grounded for contact-rich prediction and planning in manipulation. \cite{Yuan2023PhysDiff} improved the physical consistency of humanoid motions generated by a \gls{vwm}. Their method projects generated motions onto physically plausible trajectories produced by an imitation policy trained in a physics simulator. The projection module is embedded directly into the diffusion denoising network.
\cite{Romero2025Learning} imposed object-level kinematic masks to guide the training of \glspl{vwm} and improve kinematic consistency and moving object interactions, with potential applications in scene prediction for manipulation tasks. 

These methods show that physics-guided world representations can improve the physical plausibility of generated videos, benefiting downstream planning, prediction, and control. However, most rely on external physics models, such as differentiable simulators, imitation policies, or \glspl{vlm},  with only a few differentiating through contact dynamics end-to-end \citep{wang2026contactgaussianwmlearningphysicsgroundedworld}. Consequently, physical consistency depends on the fidelity of the external models and estimated parameters. Moreover, visual realism remains distinct from physical validity, and the scarcity of standardized dynamics-aware metrics makes physical consistency difficult to assess (Sec.~\ref{sec:open_questions}). Current approaches also focus mainly on rigid-body dynamics, with deformation and articulated contact receiving limited attention.

\subsection{Physics-Guided Diffusion-based Generation} \label{sec:physics_guided_diffusion}
The papers in this category used physics-based guidance to steer diffusion-based generative models (Table \ref{tab:generative_models_summary}).
\cite{Lv2026KinematicsAware} designed a kinematics-aware diffusion policy for whole-body robot arm control. They represented states and actions in a 3D node space on the robot arm, and constrained the diffusion process with analytical kinematic mappings between the node space and the robot's joint space, thus enforcing kinematic feasibility of the generated actions. 
\cite{Du2025DynaGuide} augmented the diffusion denoising process for manipulation tasks with a guidance gradient from an external dynamics model: predicted outcomes are compared with desired conditions, and their differentiable metric is backpropagated into action denoising, enabling post-hoc behavior steering without retraining the base policy. \cite{tucker2026pi} applied a similar principle to aerial manipulation: a foundation \gls{vla} policy is adapted with real-time chunking and payload-aware guidance terms. They injected gradient-based guidance corrections at inference to maintain dynamic feasibility under mass and underactuation changes. \cite{Xu2024Dynamics} used dynamics guidance in diffusion sampling for manipulator design generation: a learned differentiable dynamics model provides gradients that steer denoising toward task-effective interaction profiles.
\cite{Ma2024Hierarchical} used a low-level kinematics-aware diffusion controller for manipulation. They exploited differentiable kinematics to distill end-effector pose trajectories into joint-space diffusion trajectories, thus enforcing kinematic feasibility for multi-task manipulation.

Taken together, these methods show that physics guidance is an effective mechanism for steering diffusion models toward physically feasible solutions without modifying their architecture or retraining the base model. By injecting kinematic or dynamic knowledge during sampling, they improve task performance while preserving the flexibility and generative capabilities of diffusion-based policies. 

\subsection{Physics-Guided Neural Operators} \label{sec:neural_operators_guided}
Some works embedded physics into Koopman neural operators through guided transformations on inputs and observables. \cite{Mamakoukas2021} augmented observables with estimated time derivatives of the state, so the lifted representation explicitly includes kinematic and dynamic rates, rather than only positions. They validated the method for dynamics learning and control of an inverted pendulum and a tail-actuated robotic fish. In a differential-drive robot setting, \cite{Rosenfelder2025} incorporated first-principles nonholonomic geometry and dynamics by parametrizing the states and inputs assuming rolling-without-slipping, in both kinematic and second-order forms. They showed that purely data-driven Koopman models fail to recover these structures (“data does not replace geometry”). 
To learn quadrotor dynamics, \cite{Zinage2022} and \cite{Sriram2023} adopted $SE(3)$-consistent states (rigid-body pose with velocities) and physics-based observables derived from rotation matrices and nonlinear system's topology, aligning the Koopman representation with flight kinematics and dynamics. \cite{Sriram2023} further embedded their model in an \gls{mpc} for trajectory tracking.
Similarly, \cite{Zhang2025_Koopman} adopted a structured high-dimensional state from bicycle/needle kinematics, so their Koopman-based path tracking MPC evolves in variables that directly encode the curvature-constrained motion of a flexible robotic needle.

\cite{Tan2025_vehicle} used physics-based kinematic and dynamic vehicle models to preprocess the inputs of a \gls{deeponet} for vehicle dynamics learning, and showed improved performance over an \gls{lstm} and physical models.  

Physics-guided \glspl{no} provide the least intrusive route to operator learning, redefining only the states and observables while leaving the architecture and training loss unchanged. This also alleviates the dictionary-design challenge of physics-encoded operators (Sec.~\ref{sec:neural_operators_encoded}). \cite{Rosenfelder2025} showed that physics-guided structures must be supplied rather than learned, as purely data-driven Koopman models failed to recover the nonholonomic geometry of a differential-drive robot. However, these methods require known kinematic models, and rely on manually designed, platform-specific observables. By acting only on the inputs, they do not constrain the learned operator and cannot guarantee physically consistent predictions.


\section{Software Tools} \label{sec:software_tools}

\begin{table*}[t]
\scriptsize
\centering
\setlength{\tabcolsep}{2pt}
\renewcommand{\arraystretch}{1.1}
\begin{threeparttable}
\caption{Open-source software frameworks for embedding physics priors in robot learning. Checkmarks indicate whether a framework supports physics-encoded architectures (P.E.), physics-informed loss functions (P.I.), and the target tasks of simulation (Sim), dynamics learning/estimation (DL), and control (Ctrl). The Hardware column reports deployment options, while the last column links to the URL of the open-source software page.}
\label{tab:software_frameworks_corrected}
\renewcommand{\tabularxcolumn}[1]{m{#1}}
\begin{tabularx}{\textwidth}{
>{\centering\arraybackslash}m{2.8cm}|
>{\centering\arraybackslash}m{2.5cm}|
>{\centering\arraybackslash}m{1.2cm}|
>{\centering\arraybackslash}m{0.4cm}|
>{\centering\arraybackslash}m{0.4cm}|
>{\centering\arraybackslash}m{1.2cm}|
>{\centering\arraybackslash}m{0.45cm}|
>{\centering\arraybackslash}m{0.45cm}|
>{\centering\arraybackslash}m{0.45cm}|
>{\centering\arraybackslash}m{1.1cm}|
>{\centering\arraybackslash}X}
\toprule
\textbf{Framework} &
\begin{tabular}[c]{@{}c@{}}\textbf{Language}\\\textbf{(Backend)}\end{tabular} &
\textbf{Optimizers} &
\begin{tabular}[c]{@{}c@{}}\textbf{P.E.}\end{tabular} &
\textbf{P.I.} &
\textbf{License} &
\textbf{Sim} &
\textbf{DL} &
\textbf{Ctrl} &
\textbf{Hardware} &
\textbf{URL} \\
\hline
\multicolumn{11}{c}{\rule{0pt}{2.4ex}\textbf{\textit{Software for Model-Structured \& Hybrid Physics-Learning Architectures}}} \\
\hline
NeuroMANCER \citep{Drgona2023Domain,Drgona2026NeuroMANCER} & Python (PyTorch) & Wrapper & \checkmark & \checkmark & BSD 3-Clause &  & \checkmark & \checkmark & CPU/GPU & \url{https://github.com/pnnl/neuromancer} \\
\hline
nnodely \citep{Rosati2026Neural} & Python (PyTorch/JAX/TF/Keras) & Wrapper & \checkmark & \checkmark & MIT &  & \checkmark & \checkmark & CPU/GPU & \url{https://github.com/tonegas/nnodely} \\

\hline
\multicolumn{11}{c}{\rule{0pt}{2.4ex}\textbf{\textit{Software for Neural Ordinary Differential Equations}}} \\
\hline
torchdiffeq \citep{Chen2026Differentiable} & Python (PyTorch) & Wrapper & \checkmark &  & MIT & \checkmark & \checkmark &  & CPU/GPU & \url{https://github.com/rtqichen/torchdiffeq} \\
\hline
torchdyn \citep{Poli2020TorchDyn,DiffEqML2026TorchDyn} & Python (PyTorch) & Wrapper & \checkmark &  & Apache-2.0 & \checkmark & \checkmark & $\sim$ & CPU/GPU & \url{https://github.com/DiffEqML/torchdyn} \\
\hline
DiffEqFlux.jl \citep{Rackauckas2019DiffEqFlux,DiffEqFlux2026SciML} & Julia (SciML/Lux/Flux) & Wrapper & \checkmark & \checkmark & MIT & \checkmark & \checkmark &  & CPU/GPU & \url{https://github.com/SciML/DiffEqFlux.jl} \\

\hline
\multicolumn{11}{c}{\rule{0pt}{2.4ex}\textbf{\textit{Software for Equation Discovery \& System Identification}}} \\
\hline
PySINDy \citep{deSilva2020PySINDy,dynamicslab2026PySINDy} & Python & Custom & \checkmark &  & MIT & \checkmark & \checkmark &  & CPU & \url{https://github.com/dynamicslab/pysindy} \\
\hline
deepSI \citep{Beintema2021Nonlinear,Schoukens2026deepSI} & Python (PyTorch) & Wrapper & \checkmark &  & BSD 3-Clause & \checkmark & \checkmark &  & CPU/GPU & \url{https://github.com/MaartenSchoukens/deepSI} \\
\hline
SysIdentPy \citep{LacerdaJunior2020SysIdentPy,LacerdaJunior2026SysIdentPy} & Python (NumPy/PyTorch) & Custom & \checkmark &  & BSD-3-Clause & \checkmark & \checkmark &  & CPU/GPU & \url{https://github.com/wilsonrljr/sysidentpy} \\
\hline
deeptime \citep{Hoffmann2021Deeptime,DeeptimeML2026Deeptime} & Python \& C++ & Custom & \checkmark &  & LGPL-3.0 & \checkmark & \checkmark &  & CPU/GPU & \url{https://github.com/deeptime-ml/deeptime} \\

\hline
\multicolumn{11}{c}{\rule{0pt}{2.4ex}\textbf{\textit{Software for Physics-Informed Machine Learning}}} \\
\hline
DeepXDE \citep{Lu2021DeepXDE,Lu2026DeepXDE} & Python (TF/PT/JAX/Paddle) & Wrapper & $\sim$ & \checkmark & LGPL-2.1 & \checkmark & \checkmark &  & CPU/GPU & \url{https://github.com/lululxvi/deepxde} \\
\hline
NeuralPDE.jl \citep{Zubov2021neuralPDE,SciML2026neuralPDE} & Julia (SciML/Lux/Flux) & Wrapper &  & \checkmark & MIT & \checkmark & \checkmark &  & CPU/GPU & \url{https://github.com/SciML/NeuralPDE.jl} \\
\hline
PhysicsNeMo \citep{Hennigh2020NVIDIA,NVIDIA2026PhysicsNeMo} & Python/C++ (PyTorch) & Wrapper & \checkmark & \checkmark & Apache-2.0 & \checkmark & \checkmark &  & CPU/GPU multi-GPU & \url{https://github.com/NVIDIA/physicsnemo} \\
\hline
NeuroDiffEq \citep{chen2020neurodiffeq,NeuroDiffGym2026neurodiffeq} & Python (PyTorch) & Wrapper &  & \checkmark & MIT & \checkmark & \checkmark &  & CPU/GPU & \url{https://github.com/NeuroDiffGym/neurodiffeq} \\

\hline
\multicolumn{11}{c}{\rule{0pt}{2.4ex}\textbf{\textit{Software for Differentiable Simulation}}} \\
\hline
Brax \citep{Freeman2021Brax,Google2026Brax} & Python (JAX) & Wrapper & \checkmark &  & Apache-2.0 & \checkmark &  &  & CPU/GPU & \url{https://github.com/google/brax} \\
\hline
Nimble \citep{Werling2021Fast,Werling2026Nimble} & Python (PyTorch) & Wrapper & \checkmark &  & BSD 3-Clause & \checkmark &  &  & CPU/GPU & \url{https://github.com/keenon/nimblephysics} \\
\hline
TDS \citep{Heiden2021NeuralSim,Coumans2026Tiny} & C++/CUDA & Custom & \checkmark &  & Apache-2.0 & \checkmark &  &  & CPU/GPU & \url{https://github.com/erwincoumans/tiny-differentiable-simulator} \\
\hline
Warp \citep{Macklin2024Warp,NVIDIA2026Warp} & Python (PyTorch/JAX/Paddle) & Custom / Wrapper & \checkmark &  & Apache-2.0 & \checkmark &  &  & CPU/GPU & \url{https://github.com/NVIDIA/warp} \\
\hline
Dojo.jl \citep{Dojo2026} & Julia & Custom & \checkmark &  & MIT & \checkmark & \checkmark & \checkmark & CPU & \url{https://github.com/dojo-sim/Dojo.jl} \\
\hline
JAX-Fluids \citep{JAXFluids2026} & Python (JAX) & Wrapper & \checkmark &  & MIT & \checkmark &  &  & CPU/GPU TPU & \url{https://github.com/tumaer/JAXFLUIDS} \\

\bottomrule
\end{tabularx}
\begin{tablenotes}
\scriptsize
\item \checkmark\ denotes primary or native support, while $\sim$ denotes partial, indirect, or application-dependent support. Blank cells indicate functionalities that are not objectives of the framework. In the Optimizers column, Wrapper indicates that the framework relies on optimization algorithms provided by its underlying backend or external libraries, whereas Custom indicates dedicated optimization procedures or training algorithms. 
\end{tablenotes}
\end{threeparttable}
\end{table*}

A growing ecosystem of open-source software supports the integration of physics and model-based priors into robot learning. 
The reviews of \cite{Karniadakis2021} and \cite{Cuomo2022} discuss physics-informed and scientific machine learning libraries specific for differential-equation solving, while \cite{newbury2024review} focuses on differentiable simulators and analyzes them in terms of simulation-specific design choices. In Table~\ref{tab:software_frameworks_corrected}, we adopt a different organizing principle: we compare software tools according to how they support the embedding of physics priors in robot learning, distinguishing physics-encoded (P.E.) formulations and  physics-informed (P.I.) loss functions\footnote{We exclude physics-guided approaches, as the capability to embed physics priors in training data or model inputs is potentially supported by almost all frameworks.}. We additionally report the main robotics-oriented tasks supported by each framework, namely simulation (Sim), dynamics learning (DL), and control (Ctrl), together with programming languages, optimization methods, licenses, hardware support, and direct links to the open-source pages.


In the following sections, we discuss each software tool in detail, following the main categories highlighted in Table~\ref{tab:software_frameworks_corrected}. Specifically, we review software frameworks for model-structured and hybrid physics-learning frameworks, neural differential equation libraries, equation discovery and system identification toolkits, physics-informed machine learning, and differentiable simulators, discussing how each category supports the integration of physics priors within robot learning pipelines.


\subsection{Software for Model-Structured and 
\\Hybrid Physics-Learning Architectures} \label{sec:software_model_structured}

These tools explicitly embed physics priors within \gls{nn} architectures, and aim to support dynamics learning, control and estimation use cases.

\subsubsection{NeuroMANCER} \citep{Drgona2023Domain,Drgona2026NeuroMANCER} is a differentiable programming framework for system identification and control. It integrates \glspl{nn} with optimization layers to formulate constrained learning, physics-encoded identification, and model-based optimal control problems. It provides a unified interface for combining learning and differentiable constrained optimization.

\subsubsection{nnodely} \citep{Rosati2026Neural} is a framework for dynamics learning, control, and estimation of physical systems, supporting \glspl{msnn} and hybrid physics-neural architectures. It provides an end-to-end workflow for model definition, dataset construction, training, evaluation, and deployment, together with a library of physics-encoded building blocks for \gls{nn} architectures. It supports the composition of interconnected modules (\eg{} dynamics, estimators, and controllers), deployment to standard formats such as ONNX, and complementary approaches including \glspl{pinn} and \glspl{node}.

\subsection{Software for Neural Differential Equations}

These software frameworks embed physics priors at the architectural level, by modeling the system dynamics using \glspl{node} (Sec.~\ref{sec:node_vi}), and training via differentiable integration.

\subsubsection{torchdiffeq} \citep{chen2018neural,Chen2026Differentiable} provides differentiable \gls{ode} solvers that enable the implementation of \glspl{node}. It offers efficient adjoint-based training and \gls{gpu} support, but it is primarily a low-level building block used inside broader modeling pipelines, rather than a fully end-to-end learning framework. 

\subsubsection{torchdyn} \citep{Poli2020TorchDyn,DiffEqML2026TorchDyn} extends \gls{node}-based modeling with higher-level abstractions, including related continuous-depth and implicit models, aiming to make them accessible as plug-and-play primitives in PyTorch. It facilitates experimentation with continuous-time models and provides structured tutorials, but it is not specifically designed around domain-oriented architectural building blocks for physical systems modeling/control (in the sense of reusable engineering components). 

\subsubsection{DiffEqFlux.jl} \citep{Rackauckas2019DiffEqFlux, DiffEqFlux2026SciML} bridges differential equation solvers and \glspl{nn} in the Julia SciML ecosystem \citep{Rackauckas2020Universal}, enabling \glspl{node}, \glspl{sde}, delay equations, and related scientific machine learning workflows. Beyond basic \gls{node} functionality, it underpins the ``universal differential equations'' viewpoint for hybrid mechanistic--data-driven models, while remaining primarily focused on numerical modelling/training interfaces rather than control-oriented system design abstractions.

\subsection{Software for Equation Discovery \& System Identification}

These tools aim to learn dynamical models from data, often emphasising interpretability, classical identification structures, or data-driven state-space formalisms.

\subsubsection{PySINDy} \citep{deSilva2020PySINDy,dynamicslab2026PySINDy} implements \gls{sindy} \citep{Brunton2016} (Sec.~\ref{sec:sindy}), enabling the discovery of governing equations from data via sparse regression over function libraries. It produces interpretable dynamical models and is widely used for equation discovery pipelines in multiple domains.

\subsubsection{deepSI} \citep{Beintema2021Nonlinear,Beintema2023Deep,Schoukens2026deepSI} is a PyTorch framework for data-driven identification of dynamical systems. It provides end-to-end workflows for deep state-space identification and a platform for developing and benchmarking modern identification methods, with particular emphasis on subspace- and encoder-based approaches for nonlinear state-space models.

\subsubsection{SysIdentPy} \citep{LacerdaJunior2020SysIdentPy,LacerdaJunior2026SysIdentPy} provides classical system identification methods centred on polynomial NARMAX/NARX model structures, including tools for model structure selection and parameter estimation. It emphasises interpretability and established identification workflows, while also supporting modern estimation and integration patterns in Python. As such, it is primarily a classical identification toolkit rather than a machine learning framework. 

\subsubsection{deeptime} \citep{Hoffmann2021Deeptime,DeeptimeML2026Deeptime} offers tools for dynamical systems analysis, including Koopman models and Markov state models, with a scikit-learn-like API and deep model components for time-series-based dynamical modelling. It is widely used for time-series modelling and dynamical model estimation/analysis, but it is not specifically designed for embedding physics priors into \gls{ml} architectures for modelling and control.

\subsection{Software for Physics-Informed Machine Learning}

The following software tools primarily embed physics knowledge through loss functions, by enforcing differential equations and boundary conditions during training (Sec.~\ref{sec:physics_informed}).

\subsubsection{DeepXDE} \citep{Lu2026DeepXDE, Lu2021DeepXDE} is a library for \glspl{pinn}, supporting forward and inverse problems involving \glspl{ode} and \glspl{pde}. It provides automatic differentiation-based residual computation and supports multiphysics, with an actively maintained code and documentation. However, it is not well suited for control-oriented workflows and does not provide native export support for deployment on external runtimes or hardware targets.

\subsubsection{NeuralPDE.jl} \citep{Zubov2021neuralPDE,SciML2026neuralPDE} is part of the Julia SciML ecosystem \citep{Rackauckas2020Universal} and enables the automatic construction of physics-informed loss functions from symbolic descriptions of differential equations. It integrates tightly with numerical solver infrastructure and supports a broad range of differential equation types (including algorithmic tooling to construct and train \glspl{pinn}), but it is primarily oriented toward scientific computing applications rather than robotics workflows. 

 \subsubsection{NVIDIA PhysicsNeMo} \citep{Hennigh2020NVIDIA,NVIDIA2026PhysicsNeMo} (historically related to SimNet and now evolved into the PhysicsNeMo stack) provides a  framework for physics-informed learning and neural operators, with an emphasis on scalable training, geometry/constraint handling, and \gls{gpu}-accelerated pipelines. While highly flexible and production-oriented, it mainly targets \gls{pde}-based surrogate modelling and does not explicitly focus on robotic applications. 

\subsubsection{NeuroDiffEq} \citep{chen2020neurodiffeq,NeuroDiffGym2026neurodiffeq} is a Python library based on PyTorch for solving \glspl{ode} and \glspl{pde} using \glspl{nn}. It provides a flexible interface for defining differential problems with initial and boundary conditions, leveraging automatic differentiation to compute residuals. Compared to other \gls{pinn} frameworks, it emphasizes ease of use and flexibility, but remains primarily focused on differential equation solving rather than structured modeling and control applications.

\subsection{Differentiable Simulation Engines}

These tools embed physics directly into differentiable simulators, enabling gradient-based learning and optimization through simulation.

\subsubsection{Brax} \citep{Freeman2021Brax,Google2026Brax} is a JAX-implemented differentiable physics engine optimized for fast, massively parallel rigid-body simulation and \gls{rl} on hardware accelerators. By co-locating environment processing and training on the same device, it enables fast end-to-end learning. However, its model-based capabilities stem from explicit simulator fidelity rather than learned structured models.

\subsubsection{Nimble Physics} \citep{Werling2021Fast,Werling2026Nimble} is an analytically differentiable rigid-body physics engine (developed as a differentiable fork rooted in DART-style simulation), supporting articulated dynamics and contact with efficient gradient computation. It targets robotics-style requirements (hard contacts, complex geometries), while enabling gradient-based downstream tasks through the simulator.

\subsubsection{Tiny Differentiable Simulator} \citep{Heiden2021NeuralSim,Coumans2026Tiny} is a lightweight C++/CUDA differentiable physics engine supporting articulated rigid-body dynamics and contact models, designed to enable end-to-end differentiation through simulation. It is commonly discussed in the context of hybrid simulators augmented with \glspl{nn} (\eg{} NeuralSim), positioning it as a differentiable simulation substrate rather than a structured neural modeling framework. 

\subsubsection{Warp} \citep{Macklin2024Warp,NVIDIA2026Warp} is an NVIDIA-developed differentiable \gls{gpu} computing framework in Python, providing Just-In-Time (JIT)-compiled kernels for simulation, robotics, and geometry processing, with automatic differentiation integration into learning pipelines. It enables differentiable simulation and spatial computing at scale, and is increasingly used as a substrate for differentiable simulation and inverse problems.

\subsection{Choosing a Software Framework} \label{sec:software_selection}

Practitioners implementing the methods reviewed in this survey face a fragmented software ecosystem, with frameworks typically supporting only specific classes of physics-embedded learning.

For physics-encoded architectures (Sec.~\ref{sec:encoded_architecture}), nnodely targets \glspl{msnn} (Sec.~\ref{sec:msnn}) and hybrid physics-learning models (Sec.~\ref{sec:hybrid_physics_nn}) through cross-application architectural building blocks, while NeuroMANCER combines structured models with differentiable constrained optimization. \Glspl{node} and \glspl{vin} (Sec.~\ref{sec:node_vi}) are supported by torchdiffeq, torchdyn, and DiffEqFlux.jl, the latter also supporting universal differential equations for hybrid models. PySINDy implements \gls{sindy} (Sec.~\ref{sec:sindy}), deeptime provides Koopman operators (Sec.~\ref{sec:neural_operators_encoded}), and deepSI and SysIdentPy support deep state-space and NARMAX identification for hybrid and residual models. In contrast, no dedicated framework provides ready-to-use implementations of \gls{delan}, \glspl{lnn}, or \glspl{hnn} (Sec.~\ref{sec:lagrangian}--\ref{sec:hamiltonian}), which are typically built on general-purpose automatic differentiation libraries.

Physics-informed methods (Sec.~\ref{sec:physics_informed}) are better supported. DeepXDE, NeuralPDE.jl, and NeuroDiffEq automate the construction of physics-informed losses, while PhysicsNeMo extends these capabilities to large-scale neural operators. These frameworks, however, primarily target differential-equation solving rather than robotics, leaving embedded deployment and closed-loop integration to users.

For the physics-guided route (Sec.~\ref{sec:guided_inputs}), differentiable simulators are among the primary enabling tools, supporting physically consistent data generation and gradient-based optimization through simulation. Nimble provides differentiable rigid-body contacts, Brax and Warp emphasize massively parallel GPU simulation, TDS offers a lightweight C++/CUDA engine, and Dojo.jl targets control-oriented workflows. Conversely, inference-time physics guidance (Sec.~\ref{sec:physics_guided_diffusion}) remains largely unsupported by dedicated software and is typically implemented within task-specific diffusion pipelines.


\subsection{Discussion}

Overall, the reviewed software ecosystem comprises complementary frameworks that support different aspects of physics-embedded machine learning. Consequently, practitioners must combine multiple tools according to the requirements of the robotics application and the chosen physics-embedding paradigm.
A key limitation is that no existing framework supports physics-guided, physics-encoded, and physics-informed learning within a unified workflow spanning model definition, training, evaluation, deployment, and embedded execution. Developing interoperable software abstractions that seamlessly integrate these complementary paradigms, together with differentiable solvers, uncertainty quantification, and standardized benchmarking, remains an open research direction. Such ecosystems would simplify rapid prototyping, reproducible evaluation, and the deployment of physics-embedded learning algorithms on real-world robotic platforms.


\section{Open Research Questions and Challenges} \label{sec:open_questions}

Despite the rapid progress reviewed in this survey, embedding physics priors in robot learning methods remains an open research field. We organize the main open questions and challenges around the following themes.


\textbf{\textit{Challenge 1 -- Generality versus Physics-based Inductive Bias:}}
A fundamental trade-off exists between the flexibility of data-driven learning and the inductive bias introduced by physics priors. The success of foundation models suggests that large-scale pre-training can yield remarkable generalization. However, unlike language and vision, robotics remains fundamentally data-limited due to the cost of real-world experiments, hardware, safety constraints, and the variability of physical interactions. Moreover, it remains unclear whether large-scale pre-training alone can produce models that inherently respect physical laws.

Key questions therefore arise: Which physics priors should be embedded, and how strongly should they constrain learning? Incorrect or incomplete priors may introduce structural bias, while overly restrictive priors may limit model expressiveness.

The reviewed literature suggests that the appropriate amount and fidelity of embedded physics are problem-dependent. Strong architectural priors, such as energy-based parameterizations in \gls{delan} and \glspl{hnn} \citep{delan_lutter2019,hnn,Gupta2020}, were found to improve sample efficiency and physically plausible extrapolation when data are scarce. Hybrid approaches that retain an analytical backbone while learning residual dynamics \citep{Bolderman2024,cadelac_iros25,adaptive_control_se3_port_hnn,dikici2025learning} provide a practical compromise between physical structure and flexibility, while partially compensating for modeling errors. More broadly, smooth rigid-body dynamics benefit from stronger structural constraints, whereas contact-rich and highly nonlinear systems require greater flexibility \citep{Pfrommer2020,djeumou2023how}. Despite these insights, the field still lacks principled guidelines for selecting appropriate physics priors and determining how strongly they should be embedded for a given robotics application.

\vspace{0.2cm}
\textbf{\textit{Challenge 2 -- Reusable and Scalable Physics Priors:}}
A related challenge is the design of reusable and scalable physics priors. Most reviewed methods embed knowledge tailored to a specific robot, requiring substantial re-engineering when applied to different embodiments, tasks, or operating conditions.

Key questions are: Which physics priors are invariant across robots and tasks? How can reusable physical structure be separated from robot-specific components? Can physics-encoded architectures be composed of transferable building blocks rather than engineered individually for each platform?

Several works provide promising directions within restricted settings. Context-conditioned and adaptive formulations \citep{cadelac_iros25,adaptive_control_se3_port_hnn,djeumou2023how,mungiello2026_roboracer} improve robustness across operating conditions without retraining from scratch. Modular and equivariant architectures \citep{modlanet_se3_cartesian_map,Sharma2026} promote reusable physical structure, while the \glspl{msnn} of \cite{mungiello2026_roboracer,piccinini2025model,DaLio2020mental} reuse modular components, such as neural \glspl{fir} and local model decompositions, across vehicle dynamics learning tasks.

Nevertheless, current methods remain largely confined to closely related robots and shared morphologies. Transferring across substantially different embodiments, contact conditions, sensing modalities, and actuation mechanisms still requires considerable redesign. Developing compositional physics priors and reusable architectural building blocks that separate invariant physical structure from robot-specific components therefore remains a central challenge.

\textbf{\textit{Challenge 3 -- Physics Priors under Model Mismatch and Sim-to-Real Transfer:}}
Since real-robot data are expensive, physics-embedded models are often developed and trained in simulation \citep{Muratore2022}. A central question is therefore how inaccuracies in the assumed physics influence the behavior of physics-embedded models. While embedded priors can improve generalization and sim-to-real transfer, inaccurate assumptions may introduce structural bias that additional data alone cannot eliminate.

Key questions are: Which physics priors remain valid across the reality gap? Which components should be adapted online, and which can remain fixed? How can sim-to-real transfer failures be attributed to modeling assumptions?

The reviewed literature suggests that embedded structure facilitates transfer when the underlying physics remains approximately invariant. For example, in \gls{delan}, the Euler--Lagrange formulation transfers directly, requiring only $\inertiaMat(\genPos)$ and $\EPot(\genPos)$ to be learned or fine-tuned on real data. 
Physics-informed losses have also been shown to improve out-of-distribution generalization \citep{Saviolo2022}, allowing learning to rely on physical priors in data-scarce regions \citep{Bolderman2024}.

Conversely, embedded priors can become detrimental when the assumed model omits the dominant sources of error, such as friction, contacts, compliance, or actuator dynamics \citep{Muratore2022}. The resulting mismatch introduces systematic errors that learned residuals can only partially compensate, reducing physical parameter interpretability and weakening theoretical guarantees. At the same time, these failures are often informative, as they reveal which physical assumptions are no longer valid and therefore require refinement.

\vspace{0.2cm}
\textbf{\textit{Challenge 4 -- Interpretability, Explainability, and Trust:}}
One of the main motivations for embedding physics priors is to improve model interpretability, \ie{} the extent to which model parameters or functions admit a physical interpretation. However, combining analytical models with data-driven learning may weaken this correspondence, causing learned parameters to lose their physical meaning.
A related challenge is explainability, \ie{} understanding why a model produced a given prediction, which is particularly important in safety-critical robotics applications.

Key questions include: Are the learned physical components interpretable using physical principles? Can we identify which embedded prior drives a prediction? Can models provide human-understandable explanations? How can trust and effective debugging be supported in real-world deployments?

Some reviewed works partially address interpretability. For example, \glspl{msnn} explicitly associate learned sub-modules with physical effects, enabling interpretation of the individual trained parameters \citep{mungiello2026_roboracer,DaLio2020mental}. Equation-discovery and sparse-regression methods produce compact and interpretable symbolic representations \citep{Chu2020,Purnomo2023,Lathourakis2024,Villeda2023}, while other approaches preserve the physical meaning of learned parameters through architectural constraints or regularization \citep{chrosniak2024deep,Kolluri2025,fang2025fine}. However, these interpretability results are often qualitative, and the field still lacks standardized quantitative metrics for interpretability.

Compared with interpretability, explicit explainability remains largely unexplored. Most works infer explainability from structured architectures rather than evaluating it through attribution, counterfactual analyses, or user-centered studies. Similarly, trust is typically addressed indirectly through uncertainty estimation \citep{Kabzan2019,Kim2020,djeumou2023how} or physics-trust regularization \citep{Bolderman2022_feedforward,Bolderman2024}, while formal guarantees remain rare. Hence, the community still lacks standardized explainability protocols and quantitative trust metrics that combine physical consistency, uncertainty calibration, and actionable debugging.

\vspace{0.2cm}
\textbf{\textit{Challenge 5 -- Assessing Physical Consistency:}}
Assessing the physical consistency of learned models remains poorly standardized across the literature. Existing works evaluate different notions of consistency depending on the application. Energy-based methods quantify energy conservation, passivity, or power consistency \citep{hnn,delan_lutter2019,delan_2021,delan_4ec}. Equivariant models assess momentum conservation, symmetry preservation, or long-horizon rollout stability \citep{Sharma2026}. Contact-aware methods measure non-penetration, frictional consistency, or complementarity violations \citep{Pfrommer2020}, while physics-informed approaches typically evaluate residuals of governing Newton--Euler, \glspl{ode}, or \glspl{pde} equations \citep{Sanyal2023RAMPNet,Sivakumar2026}. Additional metrics include constraint violations, conservation-law residuals, and trajectory feasibility.

Despite their usefulness, these metrics remain fragmented, assessing different aspects of physical consistency over different prediction horizons, perturbation scenarios, and tasks, making direct comparisons difficult. This limitation is particularly evident for \glspl{vwm}, where visually realistic rollouts may still violate physical laws \citep{Li2026worldmodels}. Although recent \glspl{vwm} increasingly report dynamics-aware metrics \citep{wang2026ego,pin_wm_2025,shang2026roboscape}, a standardized evaluation protocol has yet to emerge.

A second limitation concerns benchmarks. Current evaluations remain largely tied to paper-specific datasets, simulators, and experimental setups, even when using common differentiable simulation frameworks \citep{Freeman2021Brax,Macklin2024Warp,Heiden2021NeuralSim}. Likewise, recent multi-platform datasets \citep{felan} improve reproducibility, but no benchmark currently provides a unified protocol for evaluating physical consistency across different robotic domains.

A key research direction is therefore the development of standardized metrics and benchmark suites that assess complementary aspects of physical consistency, including conservation laws, constraint satisfaction, long-horizon prediction, uncertainty calibration, and closed-loop behavior. Such protocols would enable reproducible and meaningful comparisons across physics-guided, physics-encoded, and physics-informed learning methods.

\vspace{0.2cm}
\textbf{\textit{Challenge 6 -- Bridging Physics and Safety:}}
A fundamental challenge is to combine physics priors with formal safety guarantees. In real-world robotics, physical consistency alone is insufficient: learned models must also satisfy safety constraints and, ideally, provide guarantees on closed-loop behavior. This is particularly important in safety-critical applications such as human-robot interaction, autonomous driving, and surgical robotics.

Key questions include: How can safety be guaranteed when physics priors are imperfect or the model is uncertain? How can physics-embedded learning be combined with formal verification methods?

Only few surveyed papers provide explicit theoretical guarantees\footnote{By an explicit theoretical guarantee we mean a proven property of the closed-loop or predictive behavior (\eg{} stability, passivity, constraint satisfaction, recursive feasibility, or safety), as opposed to structural properties satisfied by construction, such as the positive definiteness of the inertia matrix in physics-encoded architectures.}. \cite{Bolderman2024} established input-to-state stability for feed-forward motor control, while \cite{lnn_hnn_soft_robots} derived passivity-based stability guarantees for soft-robot control with learned energy models. Most reviewed approaches, however, are deterministic predictors without formal certification.

Explicit probabilistic treatment of model or prediction uncertainty also remains comparatively uncommon. For example, \cite{pinn_delan_contact_force} combined a \gls{delan} with Gaussian basis functions for contact-force uncertainty estimation, \cite{djeumou2023how} used a neural \gls{sde} to estimate uncertainty outside the training distribution, and \cite{Kabzan2019,Kim2020} integrated \gls{gpr} and \gls{lstm} models into uncertainty-aware \gls{mpc}.
More broadly, the integration of physics-embedded learning with formal verification techniques, such as Lyapunov analysis, control barrier functions, reachability analysis, and probabilistic safety certification, remains largely unexplored. Bridging this gap is essential for the deployment of learning-based methods in safety-critical robotics.

\section{Future Research Directions}
\label{sec:future_directions}

To address the open challenges outlined in Sec.~\ref{sec:open_questions}, we identify seven research directions that, in our view, represent the most promising paths toward complementing data-driven robot learning with physics priors.


\textbf{\textit{Direction 1 -- Combining Multiple Physics-Prior Routes:}}
A promising future direction is the systematic combination of physics-guided inputs and data, physics-encoded architectures, and physics-informed loss functions within unified learning frameworks. While each route has been investigated in isolation, only 4\% of the reviewed works combine two routes, and existing combinations remain largely \textit{ad hoc}. For example, \cite{delan_4ec} paired a \gls{delan} architecture with a power-consistency regularization loss, while \cite{Bolderman2024} combined a physics-based model, a learned \gls{mlp} residual, and a physics-regularized loss.

Future work should develop principled methodologies for selecting and combining complementary physics-prior routes according to the robotics task, available data, and prior system knowledge. Open questions include how to balance hard architectural constraints with soft loss regularization, whether physics-guided inputs can compensate for weaker architectural priors, and how to identify redundant or conflicting priors that unnecessarily limit model expressiveness.

\vspace{0.2cm}
\textbf{\textit{Direction 2 -- Physics-Aware Foundation Models:}}
A promising research direction is the development of physics-aware foundation models for robotics. While foundation models have transformed natural language processing and computer vision through large-scale pre-training, the methods surveyed in this paper are almost exclusively trained (or fine-tuned) for a single robot or application.
Moreover, recent evidence suggests that foundation models do not automatically acquire a general understanding of physics through data alone \citep{Buschoff2026}, motivating the explicit incorporation of physics priors.

Several ingredients have already emerged in the literature. Existing approaches primarily exploit physics-guided representations \citep{wang2026ego,Romero2025Learning,pin_wm_2025}, with fewer examples of physics-informed regularization \citep{stylevla2026,bouvier2025ddat,shang2026roboscape} and physics-encoded architectures \citep{cui2026physical,yang2024equibot}. However, these methods remain largely task- and robot-specific.

An important open question is whether physics-aware foundation models can generalize not only across tasks and environments, but also across robot embodiments with different morphology, mass, actuation, and contact dynamics. Achieving this goal will possibly require combining physics-guided, physics-encoded, and physics-informed approaches within large-scale pre-trained models, enabling efficient adaptation while preserving physical consistency across diverse robotic platforms.

\vspace{0.2cm}
\textbf{\textit{Direction 3 -- Differentiable Physics-Embedded Learning Pipelines:}} 
A promising research direction is the integration of physics-embedded learning into fully differentiable, end-to-end pipelines spanning perception, prediction, planning, control, and world-model simulation. End-to-end differentiability enables task-level gradients to propagate through the entire robotic pipeline, allowing physics-guided, physics-encoded, and physics-informed models to be jointly optimized. Recent work has also demonstrated differentiable learning through nonlinear \gls{mpc}, where solver sensitivities are propagated to a learned policy that adapts controller parameters online \citep{Jahncke2026}. Combined with differentiable simulators, these approaches could substantially improve sample efficiency by shifting training to simulation. 

However, end-to-end differentiability alone does not guarantee physically meaningful gradients. Gradient fidelity remains a major challenge, particularly in the presence of contact, discontinuities, optimization layers, and approximate simulators.

Another important direction is the online lifelong adaptation of physics priors during robot operation. Rather than learning fixed models offline, future methods should continuously refine their embedded physical knowledge as robots encounter new tasks, environments, and operating conditions, while preserving physical consistency and safety. Developing differentiable methods that jointly enable reliable gradient propagation, sim-to-real transfer, and safe online adaptation remains a key challenge for the next generation of physics-embedded robot learning.

\vspace{0.2cm}
\textbf{\textit{Direction 4 -- Physics-Embedded Learning for Contact-Rich Tasks:}}
Contact-rich dynamics remains one of the most important open areas for physics-embedded robot learning. While many existing methods focus on smooth rigid-body dynamics, real-world robotic tasks involve impacts, friction, compliance, deformation, and hybrid mode transitions, making them considerably more difficult to model and learn. Although several approaches have begun to incorporate contact physics through structured architectures, differentiable contact models, and physics-informed constraints \citep{Pfrommer2020,Lathourakis2024,Ristich2025,bouvier2025ddat,Yuan2023PhysDiff}, these remain largely task-specific.

Future work should develop unified physics-embedded models capable of representing contact, friction, compliance, and mode transitions across manipulation, locomotion, and multi-body robotic systems. Open challenges include learning robust contact representations from partial observations, preserving physical consistency during long-horizon interactions, and generalizing across different contact conditions, object properties, and robot morphologies.

\vspace{0.2cm}
\textbf{\textit{Direction 5 -- Automated Discovery of Physics-Encoded Architectures:}}
A promising research direction is the automated discovery of physics-encoded learning architectures. Today, designing physics-encoded models typically requires substantial engineering effort and domain expertise, and most architectures are tailored to a specific robot or application. As a result, transferring them to new systems often requires manual redesign.

Only a few works have explored the automatic discovery of physics-aware architectures. Topology-learning approaches \citep{Ledezma2017,Ledezma2018}, \glspl{eqln} \citep{Sahoo2018,Villeda2023}, and \gls{sindy}-based methods \citep{Purnomo2023,Lathourakis2024} demonstrate that model structures can be partially learned from data while retaining physical interpretability.

Future research should investigate the automatic discovery of both the topology and composition of physics-encoded architectures from libraries of physically admissible building blocks, reducing manual engineering while preserving interpretability and physical consistency. An important question is whether such approaches can match the performance of hand-crafted \glspl{msnn} on complex, high-\gls{dof} robotic systems. A particularly promising direction is hybrid methodologies, in which users specify candidate physical building blocks or constraints, while learning algorithms automatically discover their optimal composition and interconnection. This idea has been explored by \glspl{eqln} and \gls{sindy}-like methods, which combine predefined libraries of candidate functions with data-driven model discovery. Extending these methodologies to discover richer, modular physics-encoded architectures for high-dimensional robotic systems deserves further research.

\vspace{0.2cm}
\textbf{\textit{Direction 6 -- Scaling to High-\gls{dof} Systems:}}
Most surveyed methods are validated on low-dimensional systems. Manipulators and
vehicles account for 62\% of the reviewed papers, and are typically modeled with
fewer than ten \glspl{dof}, while 17 papers report results exclusively on canonical mechanical systems such as pendulums and cart-poles (Fig.~\ref{fig:paper_distribution}). In contrast,
legged robots, whose quadruped and humanoid morphologies typically exceed twenty
\glspl{dof}, appear in only 16 papers. The scalability of physics-embedded
learning to high-\gls{dof} systems therefore remains largely unexplored.
Existing work provides promising starting points, including scalable \gls{delan} for floating-base dynamics \citep{felan}, modular architectures \citep{modlanet_se3_cartesian_map}, Lie-group variational integrator networks \citep{duruisseaux2023lie}, and equivariant models for systems with hundreds of interacting bodies \citep{Sharma2026}.

A key direction is to develop physics-embedded \gls{ml} models that scale with system complexity while preserving computational efficiency and physical consistency. This requires architectures that exploit modularity, sparsity, equivariance, and reduced-order representations, while handling contact, compliance, underactuation, and unmodeled dynamics. Their evaluation should extend beyond prediction accuracy to include computational cost, long-horizon stability, constraint satisfaction, and transfer across robot morphologies and operating conditions.

\vspace{0.2cm}
\textbf{\textit{Direction 7 -- Open-Source Software Frameworks:}}
A key research direction is the development of interoperable software frameworks and benchmark suites for physics-embedded robot learning. Current tools (Sec.~\ref{sec:software_tools}) typically specialize in a single paradigm, such as physics-informed losses, continuous-time learning, differentiable simulation, system identification, or physics-encoded architectures, making it difficult to combine complementary approaches within a common workflow. Rather than a monolithic framework, future software should provide modular abstractions for physics and learning modules, differentiable solvers, uncertainty quantification, benchmarking, and deployment. Likewise, standardized benchmarks and evaluation protocols spanning multiple robot platforms, tasks, and operating conditions are needed to enable reproducible comparisons and accelerate progress.
\section{Conclusions} \label{sec:conclusions}

This survey reviewed the state of the art in physics-embedded robot learning through three complementary paradigms: physics-guided inputs, data, and representations; physics-encoded model architectures; and physics-informed training losses. Our analysis shows that physics-encoded architectures currently represent the
largest body of work (71\% of the reviewed methods), while physics-guided
(16\%) and physics-informed (13\%) approaches are rapidly gaining momentum\footnote{For papers assigned to two physics embedding routes, these percentages are computed using the primary embedding route.},  particularly since 2022. Overall, the field has grown rapidly, with more than half of the reviewed methods published between 2023 and 2025. By introducing a unified taxonomy, this survey provides a common conceptual foundation for a field that remains fragmented across terminology, methodologies, software tools, and application domains.

Our review suggests that physics priors should not be viewed as an alternative to data-driven learning, but as robotics-specific inductive biases that complement it. When embedded appropriately, physics priors have the potential to improve generalization, physical consistency, interpretability, and sample efficiency, while preserving the expressive power of \gls{ml} methods. At the same time, crucial challenges remain, including the selection of which physics priors to embed, where to inject them, and how strongly to enforce them according to prior fidelity, available data, and the required closed-loop properties. This explains why no ``universally optimal'' embedding strategy has emerged. Additional challenges include standardized evaluation protocols, safety guarantees for real-world deployment, and transferring physics priors across robots and tasks. Moreover, adoption remains concentrated on a narrow set of platforms, with 62\% of the reviewed methods targeting manipulators or vehicles, and only 7\% addressing legged robots. Finally, only 4\% of the reviewed methods combine multiple physics embedding routes, and these combinations remain largely \textit{ad hoc}, highlighting the need for more systematic integration of complementary approaches.

Looking forward, we envision a new generation of robotics algorithms that methodologically combine physics-guided, physics-encoded, and physics-informed components, rather than treating them in isolation. Together with physics-aware foundation models, scalable architectures, and interoperable software ecosystems, these developments have the potential to bridge classical model-based robotics and data-driven learning, enabling more reliable, trustworthy, and broadly applicable autonomous systems.

\begin{acks}
  \textbf{M. Piccinini} (lead): conceptualization, overall structure of the survey; writing across all sections; paper collection; figures; review and editing.  
  \textbf{L. Schulze}: writing of Sec.~\ref{sec:history}, \ref{sec:lagrangian}, \ref{sec:hamiltonian}, \ref{sec:node_vi}, and \ref{sec:symmetry_aware}; survey structure; review and editing.
  \textbf{A. Plebe}: writing of Sec.~\ref{sec:bio_inspired}; survey structure; open challenges and future directions; review and editing.
  \textbf{M. Saveriano}: writing of Sec.~\ref{sec:guided_features}, \ref{sec:geometric_learning}, and \ref{sec:frequency_domain}; survey structure; review and editing.
  \textbf{T. Beckers}: writing of Sec.~\ref{sec:physics_informed}; review and editing.
  \textbf{Y. Gao}: online repository; support with paper collection.
  \textbf{O. Arenz}: review and editing; support with Sec.~\ref{sec:lagrangian} and \ref{sec:hamiltonian}.
  \textbf{B. Zarrouki}: major review and editing across all sections.
  \textbf{D. Wang}: support with Sec.~\ref{sec:intro}; review and editing.
  \textbf{F. R. Schäfer}: support with research curation; review and editing.
  \textbf{J. Peters}: review and editing; open challenges and future directions.
  \textbf{J. Betz}: review and editing; open challenges and future directions; funding.
  \textbf{G. P. Rosati Papini}: support with survey structure and conceptualization; writing of Sec.~\ref{sec:software_tools}; online repository; review and editing.

  We also acknowledge the support of Prof. Mauro Da Lio on the conceptualization and classification of \glspl{msnn}.
\end{acks}

\begin{funding}
This work was supported by the Alexander von Humboldt Foundation though a Humboldt Postdoctoral Fellowship, and by the Italian FIS 2 Call, Grant Assignment Decree No. 1236 adopted on 01/08/2023 by the Italian Ministry of University and Research (MUR), for the project FIS-2023-03684 “Structured neural network framework for modeling and control of autonomous systems - Neu4mes”, CUP E53C24003800001.
\end{funding}

\section{List of Acronyms}
\label{sec:list_of_acronyms}

\renewcommand*{\glossarysection}[2][]{ }
\glsaddallunused
\printnoidxglossary[type=\acronymtype,style=list]

\bibliographystyle{aux/SageH}
\bibliography{
  biblio/softwares.bib,
  biblio/surveys.bib,
  biblio/physics_input_data.bib,
  biblio/physics_loss.bib,
  biblio/model_structured.bib,
  biblio/Lagrangian.bib,
  biblio/Hamiltonian.bib,
  biblio/hybrid_physics.bib,
  biblio/operators.bib,
  biblio/others_physics_architecture.bib,
  biblio/others.bib,
  biblio/topology_learning.bib,
  biblio/generative_models.bib,
  biblio/non_NN_methods.bib,
  biblio/neural_ode_vi.bib}

\end{document}